%% file: iclr2025_conference.tex
\documentclass{article} % For LaTeX2e

\usepackage[utf8]{inputenc} 
\usepackage{iclr2025_conference,times} 

\input{math_commands.tex}

\usepackage{url} 
\usepackage{minitoc} 
\usepackage{graphicx} 
\usepackage{url}
\usepackage{graphicx}
\usepackage{amsmath,amssymb} % define this before the line numbering.
\usepackage{rotating}
\usepackage{pdflscape}

\usepackage{ragged2e}
\usepackage{colortbl}
\usepackage{cite}
\usepackage{color}
\usepackage{wrapfig}
\usepackage{hhline}
\usepackage{epsfig}
\usepackage{graphicx}
\usepackage{amsmath}
\usepackage{amssymb}
\usepackage{multirow}
\usepackage{makecell}
\usepackage{booktabs}
\usepackage{algorithm}
\usepackage{algorithmic}
\usepackage{arydshln}
\usepackage{colortbl}
\usepackage{enumitem}
\usepackage{siunitx}
\usepackage{placeins}
\usepackage{bm}
\usepackage{makecell}
\usepackage{multirow} 
\usepackage{array}
\usepackage[table]{xcolor} 

\usepackage{array}
\usepackage{caption}
\usepackage{amssymb}
\usepackage{pifont}
\usepackage{subcaption}  % subtitle

\usepackage{listings}

\newcommand{\ie}{\textit{i.e.}}
\newcommand{\eg}{\textit{e.g.}}

\usepackage[dvipsnames,svgnames]{xcolor} 
\usepackage{hyperref}
\hypersetup{
  colorlinks=true,
  citecolor=SteelBlue,
}

\title{Reconstruction Alignment Improves Unified Multimodal Models}

\author{Ji Xie$^{1}$, Trevor Darrell$^{1}$, Luke Zettlemoyer$^{2}$, XuDong Wang$^{1,3}$\thanks{corresponding author} \\
$^{1}$UC Berkeley, $^{2}$University of Washington, $^{3}$Duke University \\
\textit{Project Page: \href{https://reconstruction-alignment.github.io/}{\small{https://reconstruction-alignment.github.io/}}}
}

\definecolor{ForestGreen}{rgb}{0.13, 0.55, 0.13}
\definecolor{Green}{rgb}{0.0, 0.5, 0.0}
\definecolor{green(munsell)}{rgb}{0.0, 0.66, 0.47}
\definecolor{green(ryb)}{rgb}{0.4, 0.69, 0.2}
\definecolor{green(pigment)}{rgb}{0.0, 0.65, 0.31}
\definecolor{bananayellow}{rgb}{1.0, 0.88, 0.21}
\definecolor{chromeyellow}{rgb}{1.0,0.8,0.156}
\definecolor{tiffany}{rgb}{0.6078,0.7961,0.7804}
\definecolor{customgreen}{HTML}{F9FFF9}

\usepackage{colortbl}
\newcommand{\cmark}{\text{\ding{51}}}%
\newcommand{\xmark}{\text{\ding{55}}}%
\newcommand{\cmarkColor}{\textcolor{Green!90}{\text{\ding{51}}}}%

\usepackage{xspace}
\newcommand{\ours}{\textsc{RecA}\xspace}
\newcommand{\oursfull}{Reconstruction Alignment\xspace}
\newcommand{\versus}{\textit{vs.}\xspace}

\begin{document}
\iclrfinalcopy

\maketitle

\vspace{-14pt}
\input{sections/01abstract}

\begin{figure}[htbp]
\vspace{-8pt}
\centering
\includegraphics[width=\linewidth]{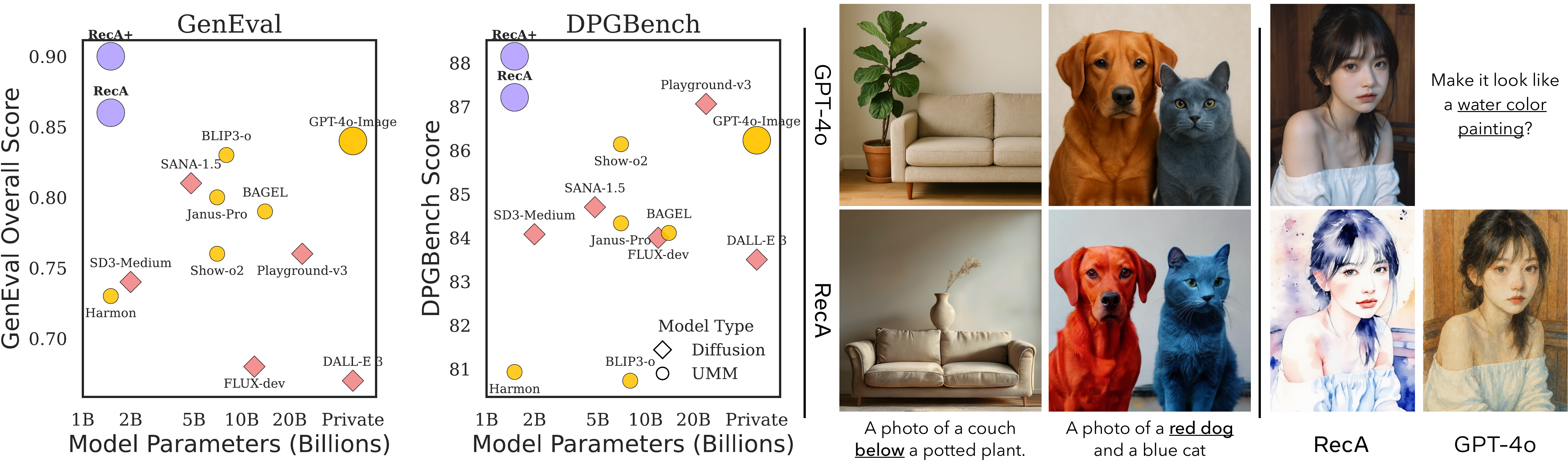}
\vspace{-16pt}
\caption{\textbf{Post-training UMMs with reconstruction alignment (\ie, \textbf{\ours}) substantially improve image generation and editing.} 
\textbf{\textit{Left}}: performance comparison on GenEval and DPGBench, where a 1.5B-parameter model post-trained with \ours surpasses much larger models across multiple benchmarks (Table~\ref{tab:main_table}: GenEval, DPGBench and WISE); \textbf{\textit{Middle}}: compared with GPT-4o, \ours follows generation instructions more faithfully, especially for \textit{color attributes} and \textit{spatial positions}; 
\textbf{\textit{Right}}: for editing, \ours better preserves \textit{instance identity}, \textit{overall layout}, and \textit{object shapes} of the original images, such as the girl's lips.}
\label{fig:teaser}
\end{figure}

% Table~\ref{tab:editing_results}: ImgEdit and GEdit-Bench-EN); 

\input{sections/02introduction}
\input{sections/04method}
\input{sections/05experiments}
\input{sections/03related_work}

\input{sections/06conclusion}

\bibliography{iclr2025_conference}
\bibliographystyle{iclr2025_conference}

\appendix

\input{sections/appendix}

\end{document}

%% file: math_commands.tex
\usepackage{amsmath,amsfonts,bm}
\usepackage{xcolor}

\definecolor{darkgreen}{RGB}{0,100,0}
\definecolor{darkred}{RGB}{160,0,0}
\definecolor{darkgrey}{RGB}{60,60,60}
\definecolor{brown}{RGB}{99,49,15}

\newcommand{\textgr}[1]{\text{\textcolor{ForestGreen}{#1}}}

\newcommand{\textre}[1]{\textbf{\textcolor{darkred}{#1}}}

\def\eqref#1{equation~\ref{#1}}
\def\1{\bm{1}}

\DeclareMathAlphabet{\mathsfit}{\encodingdefault}{\sfdefault}{m}{sl}
\SetMathAlphabet{\mathsfit}{bold}{\encodingdefault}{\sfdefault}{bx}{n}

%% file: sections/01abstract.tex
\begin{abstract}

% Version 4 
We discover a simple but effective way for visual understanding to help visual generation in a unified model.
Unified multimodal models (UMMs) unify visual understanding and generation within a single architecture. 
However, conventional training relies on image–text pairs (or sequences) whose captions are typically sparse and miss fine-grained visual details, even when {they use} hundreds of words to describe a simple image. 
We introduce \textbf{\oursfull} (\textbf{\ours}), a resource-efficient post-training method that leverages visual understanding encoder embeddings as dense ``text prompts,'' providing rich supervision without captions. 
Concretely, \ours conditions a UMM on its own visual understanding embeddings and optimizes it to reconstruct the input image with a self-supervised reconstruction loss, thereby using visual understanding to improve image generation.
Despite its simplicity, \ours is broadly applicable: across autoregressive, masked-autoregressive, and diffusion-based UMMs, it consistently improves generation and editing fidelity. 
With only 27 GPU hours, post-training with \ours substantially improves image generation performance on GenEval (0.73→0.90) and DPGBench (80.93→88.15), while also boosting editing benchmarks (ImgEdit 3.38→3.75, GEdit 6.94→7.27). 
Notably, \ours surpasses much larger open-source models and applies broadly across diverse UMM architectures, establishing it as an efficient and general post-training alignment strategy for UMMs. 
\end{abstract}

%% file: sections/02introduction.tex
\section{Introduction}

\begin{figure}[t]
\centering
\includegraphics[width=\linewidth]
{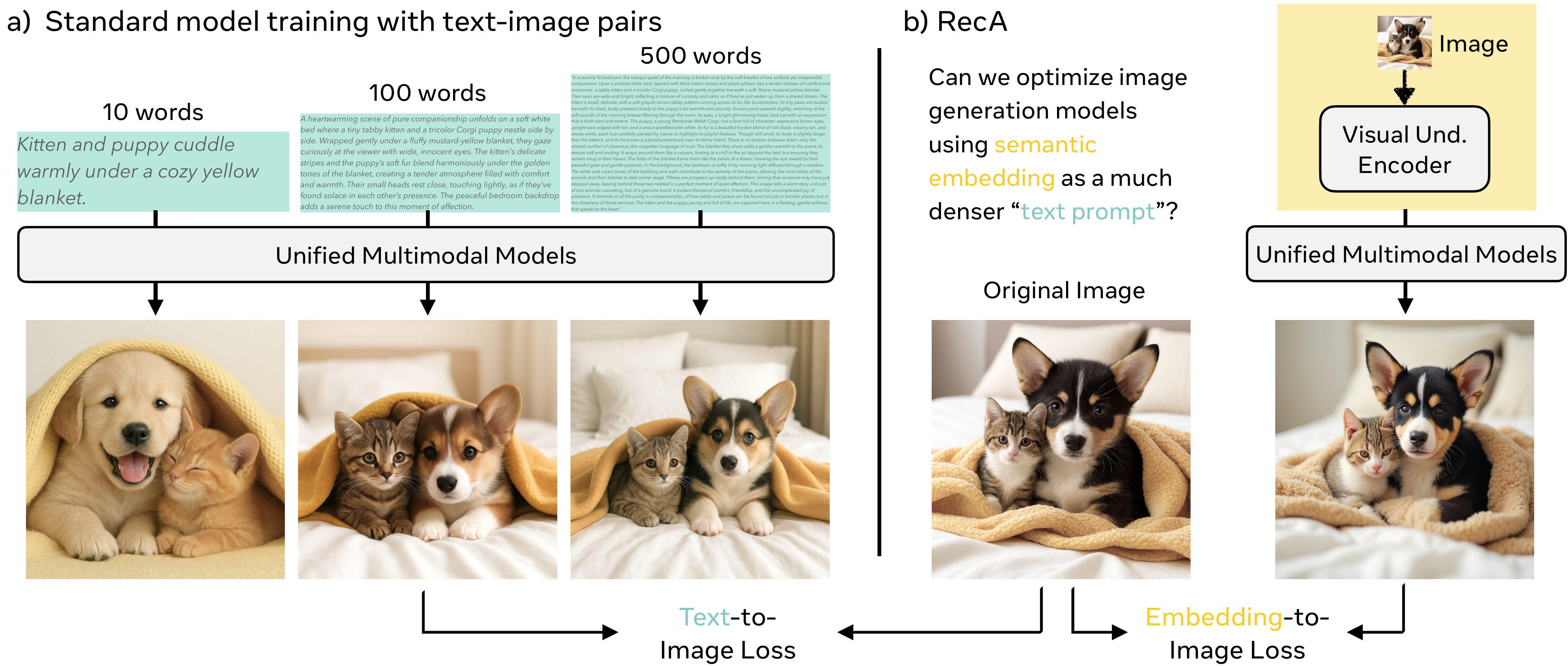}
\vspace{-16pt}
\caption{\textbf{Dense supervision from visual embeddings.} 
a) Typical image generation models are trained on \textcolor{tiffany}{image–caption} pairs {and/or sequences whose} text is a \textcolor{tiffany}{\emph{sparse}} representation of visual information. 
\textbf{\emph{An image is worth far more than a hundred words}} and contains rich details that text alone cannot capture.  
As shown in the left three examples, even lengthy captions (500 words) miss key aspects such as \textit{textures, styles, layouts, shapes, and attributes}, leading to imperfect generations relative to the original image.  
b) By contrast, embeddings from \emph{visual understanding encoders}, \eg, CLIP, preserve richer and more faithful semantics.  
Can these \textcolor{chromeyellow}{image–embedding} pairs provide the \textcolor{chromeyellow}{\emph{dense}} supervision needed to enhance image generation and editing?  
Surprisingly, the answer is \emph{\textbf{yes}}: we find that \textcolor{black}{image–embedding} pairs can improve T2I and image editing in a \textbf{zero-shot} manner.}
\label{fig:motivation}
\vspace{-8pt}

\end{figure}

Building on the success of large language models (LLMs)~\citep{gpt3, llama, qwen2.5}, researchers have developed \emph{Multimodal} Large Language Models (MLLMs)~\citep{llava,qwen_vl,Qwen2.5-VL,clip,siglip,internvl,zhu2025internvl3} with strong visual understanding performance.
Recently, unified multimodal models (UMMs), or Omni Models, have been proposed to both understand and generate across modalities—reading and writing visual content and text within a single architecture~\citep{tong2024metamorph, an2025unictokens, lmfusion, Nexus-Gen, seed-x, vila-u, pan2025generative, openai2024introducing4o, li2025onecatdecoderonlyautoregressivemodel, wu2025qwen, niu2025does}. 
The academic community envisions that a unified model can inherit the reasoning and world knowledge of LLMs while letting visual understanding help visual generation.

\setlength{\intextsep}{8pt} % or a small value like 2pt
\begin{wrapfigure}{l}{0.50\linewidth}
    \centering
    \vspace{-8pt}
    \includegraphics[width=\linewidth]{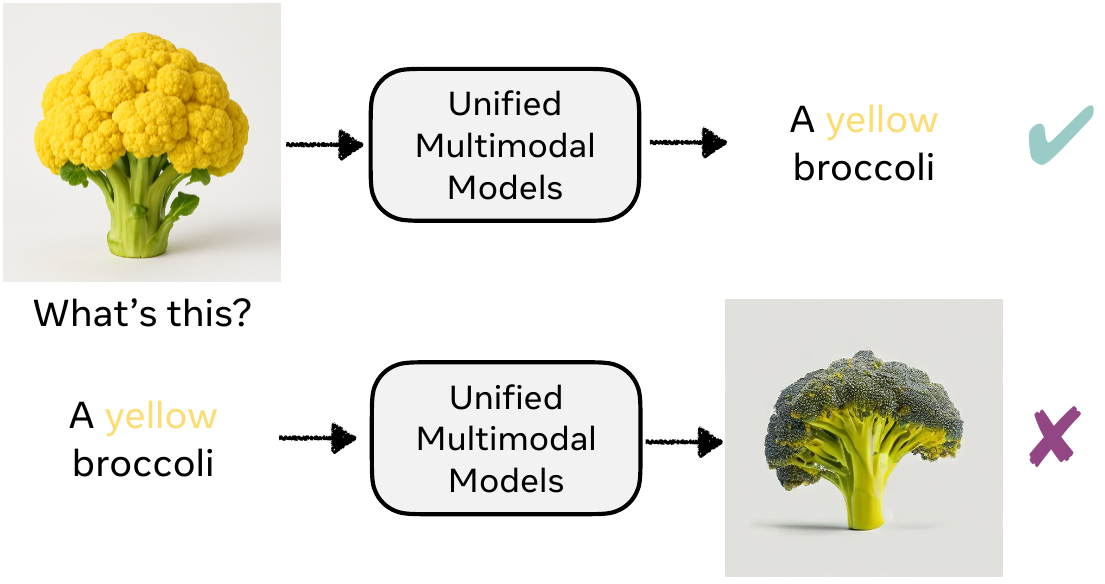}
    \vspace{-14pt}
    \caption{UMMs can often correctly recognize an uncommon concept (yellow broccoli) but fail to generate it, revealing misalignment between understanding and generation.}
    \label{fig:misaligned}
\end{wrapfigure}

However, UMMs face a fundamental limitation: conventional training relies on image–text pairs, where text captions provide supervision. Even captions spanning hundreds of words omit critical visual details such as spatial layout, geometry, or fine-grained attributes (Figure~\ref{fig:motivation}), introducing systematic biases into the learned representations. 
For instance, because captions rarely describe broccoli's color, models tend to overfit to the rule \emph{broccoli $\to$ green}, often collapsing to green outputs when given atypical prompts like ``a \emph{yellow} broccoli'' (Figure~\ref{fig:misaligned}).
This misalignment motivates us to explore alternative forms of supervision. Rather than relying on captions, we leverage embeddings from UMM's visual understanding encoders~\citep{clip,siglip,internvl,zhu2025internvl3}, which can map pixels into a language-aligned semantic space interpretable by themselves. Crucially, embeddings from \emph{understanding} encoders (\eg, CLIP, SigLIP) capture semantic structure far more effectively than those from \emph{generation} encoders (\eg, VAE, VQ-GAN). 
These semantic embeddings provide dense, semantically grounded supervision without paired captions, raising a central question: \emph{Can we improve the generation capabilities of UMMs by training them with semantic embeddings as maximally informative ``text prompts''?}

Building on these insights, we propose \ours, a resource-efficient post-training strategy. The core idea is simple: condition UMMs on their own visual understanding encoder embeddings—dense ``visual prompts'' that encode layout, color, and attributes beyond what captions capture—and train them to reconstruct the image. This semantic reconstruction lets visual understanding help visual generation with richer supervision and no additional labels. Despite its simplicity, \ours yields substantial improvements. With only \emph{27 A100 GPU hours}, a 1.5B-parameter UMM post-trained with \ours surpasses significantly larger open-source models, achieving state-of-the-art results on GenEval (0.86) and DPGBench (87.21). Importantly, these gains are obtained \textit{\textbf{without any GPT-4o-Image distillation data or reinforcement learning}}~\citep{openai2024introducing4o,chen2025blip3,chen2025sharegpt}, in contrast to prior work. Moreover, when post-trained with GPT-4o-Image data, \ours further improves to GenEval (0.90) and DPGBench (88.15). \ours also boosts image editing quality, raising ImgEdit from 3.38 to 3.75 and GEdit from 6.94 to 7.27. 
Moreover, \ours applies broadly across UMM families with different generation mechanisms, including Show-o~\citep{showo}  {(Discrete)}, Harmon~\citep{wu2025harmon} {(MAR)}, OpenUni~\citep{openuni} {(Continuous)}, and BAGEL~\citep{bagel} {(Continuous)}, highlighting its generality. 

The key contributions can be summarized as follows: 

\begin{itemize}[leftmargin=12pt, noitemsep, topsep=0pt]
\item \textbf{Method:} We introduce \ours, a semantic reconstruction–based post-training method that uses visual understanding as ``dense prompts'' to help visual generation, providing rich supervision without captions.

\item \textbf{Generality:} We show that \ours consistently improves diverse UMM architectures, spanning autoregressive models to hybrid frameworks that integrate autoregressive and diffusion models.

\item \textbf{Performance:} We demonstrate strong empirical gains, with a 1.5B-parameter model surpassing GPT-4o and larger open-source models using only 27 A100 GPU hours, significantly outperforming prior state-of-the-art methods \emph{without distillation} or \emph{reinforcement learning (RL)}.

\end{itemize}

%% file: sections/04method.tex
\begin{figure}[t]
\centering
\includegraphics[width=0.98\linewidth]
{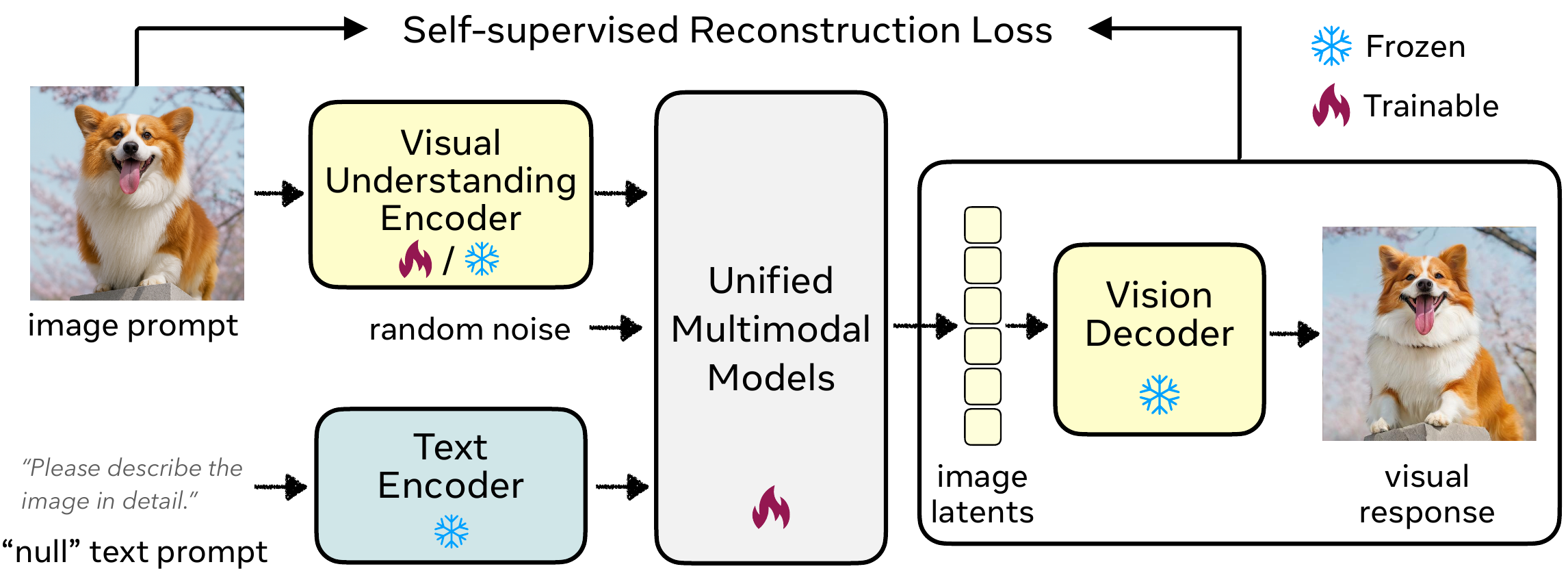}
\vspace{-2pt}
\caption{\textbf{Overview of the reconstruction alignment (\textbf{\ours}) pipeline}.  
A visual \emph{understanding} encoder (\eg, CLIP or SigLIP) extracts semantic features from the input image, which are fused with template text embeddings and passed to the Unified Multimodal Model (UMM) to regenerate the image. The UMM is optimized with a self-supervised loss (diffusion or cross-entropy) between the original and generated images or image latents. We \textbf{freeze the understanding encoder} except in cases where the UMM employs shared encoder for both understanding and generation (e.g., Harmon). At inference time, \ours \textbf{requires no additional inputs}, operating as a standard UMM.
}
\vspace{-8pt}

\label{fig:framework}
\end{figure}

\section{Reconstruction Alignment}

In this section, we present \textbf{\oursfull (\ours)} as a self-supervised image reconstruction objective. By training the model to reconstruct images from its visual understanding embeddings, \ours provides dense supervision that captures fine-grained visual details often omitted by text captions. We show the overall pipeline in Figure~\ref{fig:framework}.

\begin{figure}[t]
    \centering
    \includegraphics[width=\linewidth]{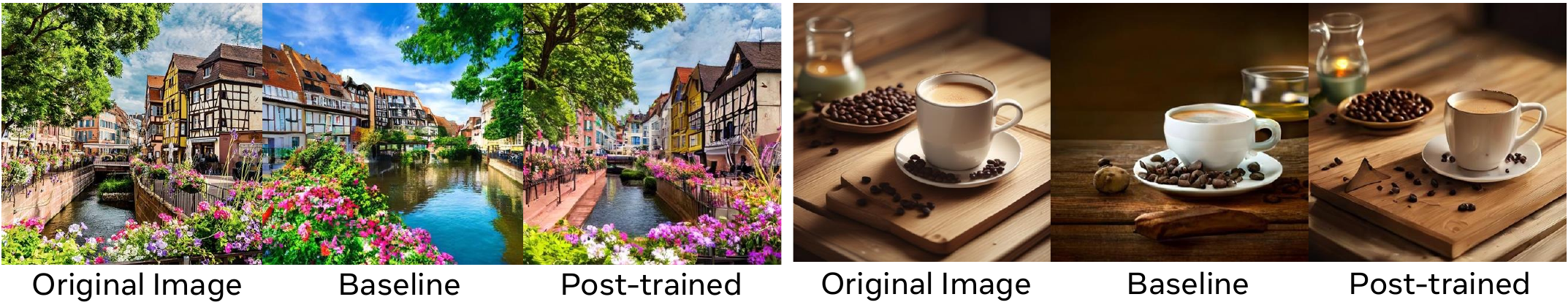}
    \vspace{-12pt}
    \caption{\textbf{Post-training with \ours restores visual details missed by the baseline models.} 
    For each query image (left), we feed its visual understanding embeddings back into the UMM with the instruction \emph{``Describe the image in detail.''}
    The baseline model (center)'s visual responses, \ie, images, preserve the main subject but distort layout, textures, and colors, while \ours markedly restores visual details like geometry, color, and overall fidelity.}
    \label{fig:motivation_reconstruction}
    \vspace{-8pt}
\end{figure}

% \begin{figure}[t]
%     \centering
%     \begin{subfigure}{0.45\textwidth}
%         \centering
%         \includegraphics[width=\linewidth]{figures/motivation_distance.pdf}
%     \end{subfigure}
%     \hfill
%     \begin{subfigure}{0.54\textwidth}
%         \centering
%         \includegraphics[width=\linewidth]{figures/motivation_reconstruction.pdf}
%     \vspace{-16pt}
%     \end{subfigure}
%     \vspace{-10pt}
%     % \caption{\textbf{Observation 1 (left):} After PCA, text embeddings (e.g., ``white'') are observed to lie close to their corresponding image embedding regions (e.g., the white dog) in the deeper layers of MLLMs. \textbf{Observation 2 (right):} UMMs fail to reconstruct images from their own visual understanding embeddings, while \ours significantly restores visual details and fidelity.}
%     % \caption{\textbf{Observation 1 (left):} After PCA, text embeddings (e.g., ``white'') are observed to lie close to their corresponding image embedding regions (e.g., the white dog) in the deeper layers of MLLMs. \textbf{Observation 2 (right):} UMMs fail to reconstruct images from their own visual understanding embeddings, while \ours significantly restores visual details and fidelity.}
%     \label{fig:motivation_method}
%     \vspace{-10pt}
% \end{figure}

\subsection{Motivation and Setup}

% Modern visual encoders project image features into a language-aligned semantic space, enabling MLLMs to interpret images as if they were text. To verify this, we apply PCA~\citep{abdi2010principal} to both text embeddings and image embeddings in the deeper layers of Harmon-1.5B~\citep{wu2025harmon}, and compute their pairwise distances. As shown in \textbf{Observation 1} (Figure~\ref{fig:motivation_method}), text embeddings (e.g., ``white'') are located close to the corresponding image regions (e.g., the white dog), which suggests that textual and visual semantics in LLM are already well aligned. Therefore, a truly \emph{unified} model with strong textual understanding should be able to reconstruct the original image from such embeddings.

Equipped with a visual understanding encoder, modern multimodal LLMs (MLLMs) effectively treat images as rich, dense context and successfully handle not only basic perceptual and question answering tasks, but also challenging multimodal reasoning problems such as symbolic, spatial, and mathematical reasoning~\citep{Qwen2.5-VL,openai2024introducing4o,huang2025vision,chen2024spatialvlm}.

One may expect that this dense, text-like context can be naturally extended to the image generation task. However, when we extract embeddings from the understanding encoder, insert them into a prompt template (e.g., \emph{``Describe the image in detail.''}), and ask the UMM to generate, the results in Figure~\ref{fig:motivation_reconstruction} reveal a gap: the main subject is preserved, but spatial layout and composition are scrambled. This raises the hypothesis that the image itself can be treated as a \emph{dense ``text prompt''} to train the model, enabling the generation branch of the UMM to better exploit the LLM features and improve its generation capability.

\subsection{\ours Training Paradigm}

\textbf{Training losses.} Traditional UMMs are trained with a combination of text-to-image (T2I) and image-to-text (I2T) objectives. Formally:

\begin{equation}
        \vspace{-2pt}
\mathcal{L}_{\text{t2i}} = \mathcal{L}(f_{\theta}(t_{\text{prompt}}), I_{\text{gt}}), \quad 
\mathcal{L}_{\text{i2t}} = \mathcal{L}(f_{\theta}(\text{concat}(t_{\text{question}}, \mathbf{h}_v)), t_{\text{answer}})
\end{equation}

where $\mathcal{L}(\cdot,\cdot)$ denotes the training loss (\eg, cross-entropy for autoregressive models~\citep{showo,janus}, diffusion loss for diffusion-based models~\citep{zhou2025transfusion,bagel}), $t_{\text{prompt}}, t_{\text{question}}, t_{\text{answer}}$ are text inputs/outputs, $\mathbf{h}_v$ are embeddings extracted from the understanding encoder, and $I_{\text{gt}}$ is the ground-truth image (we omit the explicit VAE decoder notation for simplicity). $f_{\theta}(\cdot)$ represents the UMM with parameters $\theta$. Preliminaries of different image generation paradigms are listed in Appendix~\ref{sec:preliminaries}.

Our key idea is to replace conventional T2I supervision with a semantic-level image reconstruction loss. Instead of using text captions that are inherently sparse in visual information, we condition the UMM on its own visual understanding embeddings. The reconstruction loss is:

\begin{equation}
    \vspace{-2pt}
\mathcal{L}_{\text{\ours}} = \mathcal{L}(f_{\theta}(\text{concat}(t_{\text{template}}, \mathbf{h}_v)), I_{\text{gt}})
\end{equation}

where $t_{\text{template}}$ is a prompt template triggering image reconstruction (\eg, \emph{Describe the image in detail}). {More details of the template collection and visual embedding integration are provided in Appendix~\ref{sec:templates}}. The overall training loss is:

\begin{equation}
     \vspace{-2pt}
\mathcal{L}_{\text{total}} = \lambda_{\text{\ours}} \mathcal{L}_{\text{\ours}} + \lambda_{\text{i2t}} \mathcal{L}_{\text{i2t}} + \lambda_{\text{t2i}} \mathcal{L}_{\text{t2i}}.
\end{equation}

\label{method:i2t}

We set $\lambda_{\text{\ours}}=1,\lambda_{\text{t2i}}=0$. For UMMs that share parameters between understanding and generation, we set $\lambda_{\text{i2t}}=1$ to preserve I2T capability; otherwise, we set $\lambda_{\text{i2t}}=0$ and \textbf{freeze the understanding component}. We freeze the understanding encoder in most cases, except when the UMM employs the same encoder for both tasks (\eg, Harmon~\citep{wu2025harmon}). 

{\textbf{Input resolution.} Previous works show that higher-resolution visual understanding embeddings retain substantially more pixel-level detail~\citep{allakhverdov2025image,lin2025uniworld}. To encourage the model to focus on semantic-level reconstruction, we resize input images to the minimum resolution accepted by the understanding encoder.}

\textbf{Model inference.} At inference time, our post-trained UMM operates identically to a standard one and requires no additional visual embeddings. For image generation, only a text prompt is needed; for image editing, the inputs remain the text prompt and the original image. This preserves the model's original usability while providing enhanced generation capabilities.

\subsection{Difference Between \ours and Previous Works}

Previous works integrate image reconstruction in different ways: 
(I) \textbf{diffusion-supervised enhancement} (\eg, ViLex)~\citep{vilex,ma2025genhancer,luo2024deem,wang2024diffusion} leverages pretrained diffusion models to \textit{regularize vision encoders} and improve \emph{visual understanding}; 
(II) \textbf{reconstruction from hidden states} (\eg, ROSS)~\citep{wang2024reconstructive,wang2025ross3d}) \textit{adds lightweight decoders} to reconstruct input images from \emph{intermediate embeddings}, thereby regularizing the model to preserve fine-grained details for \emph{visual understanding}; 
(III) \textbf{representation alignment} (\eg, REPA, VA-VAE)~\citep{yu2024representation,yao2025reconstruction} introduces \textit{additional alignment modules} that map hidden states in DiT~\citep{DiT,ma2024sit} or VAE~\citep{vae} to representations obtained from \emph{external, pretrained vision encoders}; 
and (IV) \textbf{reconstruction as a prior} (\eg, Lumos)~\citep{ma2025learning} adds \textit{additional DINO features~\citep{caron2021emerging} into the attention blocks of the diffusion model}, which is further trained on large-scale text-to-image data. 

\ours treats images as \emph{dense prompts} and adopts \emph{semantic-level reconstruction} as a native \textbf{post-training} objective for UMMs, requiring no auxiliary modules or text-to-image data. \ours is fundamentally different in terms of the methodology, architecture, motivation and task.

% \section{Difference Between \ours and Classifier-Free Guidance (CFG)}
% \label{sec:difference_cfg}
% \textbf{Classifier-free guidance} (CFG~\citep{ho2022classifier}) is typically used to improve image generation fidelity. 
% At each generation step, we compute a conditional prediction $\mathbf{o}_{\text{cond}}$ and an unconditional prediction $\mathbf{o}_{\text{uncond}}$, where $\mathbf{o}$ denotes either the autoregressive head's logits or the diffusion head's predicted noise, as we introduced in Appendix~\ref{sec:preliminaries}. The output is given by:

% \begin{equation}
%     \vspace{-2pt}
%     \mathbf{o} = (1+\omega)\,\mathbf{o}_{\text{cond}} - \omega\,\mathbf{o}_{\text{uncond}},
%     % \vspace{-2pt}
% \end{equation}

% with $\omega$ the guidance scale. \ours is conceptually orthogonal to CFG. Whereas CFG relies on contrast between a conditional and a null-text (or template) prompt, \ours leverages embeddings from the visual understanding encoder as dense prompts for reconstruction-based alignment. The two techniques are fully compatible and can be applied together. 

%% file: sections/05experiments.tex
\section{Experiments}
\label{sec:exp}

\begin{table*}[t!]
    \centering
    \caption{\textbf{Results on GenEval and DPGBench}. Scores marked with (*) are our reproduced results using 12 random seeds. 
    We use Harmon-1.5B as baseline and post-train it with \ours. The \textcolor{gray!80}{gray-colored} rows denote private models, and their results are cited from~\citep{yan2025gpt,geng2025xomnireinforcementlearningmakes}. Arrows ($\uparrow$) denote that higher is better. Detailed comparison is listed in Appendix~\ref{sec:quantitative_results}.
    }
    \vspace{-2.5mm}
    \label{tab:main_table}
    \small
    \setlength{\tabcolsep}{2.0pt}
    \renewcommand\arraystretch{0.9}
    % --------------------------------------------------------
    %  Column layout: | Method | Params | 7×GenEval | DPGBench |
    % --------------------------------------------------------
    \begin{tabular}{lcccccccccc}
        % \toprule
        % \bottomrule[1pt]\rowcolor[HTML]{FAFAFA}
         & & \multicolumn{7}{c}{GenEval $\uparrow$} & \multicolumn{1}{c}{\ \ \ \  DPG$\uparrow$ \ \ \ } \\
        \cmidrule(lr){3-9} \cmidrule(lr){10-10}
        \multicolumn{1}{c}{Model} & \multicolumn{1}{c}{Params} & \multicolumn{1}{c}{Single Obj.} & Two Obj. & Counting & Colors & Position & Color Attri. & Overall & Score \\
        \toprule[0.8pt]

        % ============================ Gen. Only ============================
        % \rowcolor[HTML]{F1F9FF}
        % \multicolumn{10}{l}{\textit{Generation Only Models}} \\
        % \toprule[0.4pt]
        % \textcolor{gray}{DALL-E~3} & \textcolor{gray}{--} & \textcolor{gray}{0.96} & \textcolor{gray}{0.87} & \textcolor{gray}{0.47} & \textcolor{gray}{0.83} & \textcolor{gray}{0.43} & \textcolor{gray}{0.45} & \textcolor{gray}{0.67} & \textcolor{gray}{83.50} \\
        % \toprule[0.8pt]
        % \rowcolor[HTML]{F9FFF9}
        % \multicolumn{10}{l}{\textit{Unified Multimodal Models}} \\
        % \toprule[0.4pt]            
        % \color{black} Show-o* & 1.3B & 0.98 & 0.84 & 0.67 & 0.82 & 0.30 & 0.52 & 0.69 & 83.63 \\
        Harmon* & 1.5B & 0.99 & 0.87 & 0.69 & 0.86 & 0.45 & 0.51 & 0.73 & 80.93 \\   
        SD3-Medium & 2B & 0.99 & 0.94 & 0.72 & 0.89 & 0.33 & 0.60 & 0.74 & 84.08 \\
        Janus-Pro & 7B & 0.99 & 0.89 & 0.59 & 0.90 & \textbf{0.79} & 0.66 & 0.80 & 84.33 \\        
        FLUX-dev & 12B & 0.99 & 0.85 & 0.74 & 0.79 & 0.21 & 0.48 & 0.68 & 84.00 \\
        BAGEL* & 14B & 0.99 & 0.93 & 0.80 & 0.86 & 0.51 & 0.63 & 0.79 & 84.03 \\
        Playground-v3 & 24B & 0.99 & 0.95 & 0.72 & 0.82 & 0.50 & 0.54 & 0.76 & 87.06 \\ \color{gray}
        \textcolor{gray}{GPT-4o-Image} & \textcolor{gray}{-} & \textcolor{gray}{0.99} & \textcolor{gray}{0.92} & \textcolor{gray}{\textbf{0.85}} & \textcolor{gray}{0.89} & \textcolor{gray}{0.74} & \textcolor{gray}{0.71} & \textcolor{gray}{0.84} & \textcolor{gray}{86.23} \\
        \rowcolor[HTML]{F9FFF9}
        \bf \ours & 1.5B & 1.00 & \textbf{0.98} & 0.71 & \textbf{0.93} & 0.76 & \textbf{0.77} & \textbf{0.86} & \textbf{87.21} \\
        \toprule[0.4pt]
        \multicolumn{10}{l}{\textit{Models Trained with GPT-4o Data}} \\
        \toprule[0.4pt]
        Ovis-U1 & 3.6B & \text{0.98} & \textbf{\text{0.98}} & \textbf{\text{0.90}} & \text{0.92} & \text{0.79} & \text{0.75} & \text{0.89} & \text{83.72} & \\
        OmniGen2 & 7B & \text{1.00} & \text{0.95} & \text{0.64} & \text{0.88} & \text{0.55} & \text{0.76} & \text{0.80} & \text{83.57} \\
        % \hline
        BLIP3-o* & 8B & \text{1.00} & \text{0.92} & \text{0.63} & \text{0.91} & \text{0.86} & \text{0.67} & \text{0.83} & \text{80.73} \\
        \rowcolor[HTML]{F9FFF9}
        \bf \ours & 1.5B & \text{1.00} & \text{0.97} & \text{0.76} & \textbf{\text{0.94}} & \textbf{\text{0.91}} & \textbf{\text{0.83}} & \textbf{\text{0.90}} & \textbf{\text{88.15}} \\
        
        % \bottomrule
    \end{tabular}
    \vspace{-3mm}
\end{table*}

We validate our \ours method across various unified multimodal models (UMMs), image datasets, and evaluation benchmarks. In particular, we investigate the following aspects:

\begin{itemize}[leftmargin=16pt, noitemsep, topsep=0pt]
    \item \textbf{SOTA Results:} \ours achieves state-of-the-art performance on both image generation and editing benchmarks. (Table~\ref{tab:main_table}, Table~\ref{tab:editing_results})  
    \item \textbf{Generality:} \ours delivers consistent performance gains across diverse UMM frameworks, demonstrating its generalizability. (Table~\ref{tab:model_improvement}, Figure~\ref{fig:baseline_vs_finetuned}, Figure~\ref{fig:imgedit})  
    \item \textbf{Robustness:} \ours consistently improves generation capabilities across diverse datasets and benchmarks, indicating that \textit{the gains are not from training-data memorization}. (Table~\ref{tab:sft_realign_comparison})
    \item \textbf{Training Paradigm:} \ours serves as a post-training method applied after UMM pre-training, and is most effective when used at the final stage of model training. (Table~\ref{tab:sft_realign_comparison}, Table~\ref{tab:training_order})
\end{itemize}

\subsection{Experiment Setup}
\label{subsec:setup}

\textbf{Model architectures.} We evaluate \ours across four open-source UMM architectures: 
\begin{itemize}[leftmargin=16pt, noitemsep, topsep=0pt]
    \item \textbf{Show-o {(Discrete)}}~\citep{showo} employs {discrete token generation with MaskGIT paradigm~\citep{chang2022maskgit}}, using CLIP~\citep{clip} / VQGAN~\citep{vqgan} as the understanding / generation encoder, evaluated at 256$\times$256 and 512$\times$512 resolutions. {Our main results adopt the CLIP-understanding variant, while Appendix~\ref{sec:showo_tokenizer} explores and analyzes the VQGAN-understanding variant in detail.}
    \item \textbf{Harmon {(MAR)}}~\citep{wu2025harmon} adopts masked autoregressive generation, with MAE~\citep{li2024autoregressive} / MAE \& VAE~\citep{vae} as the understanding / generation encoder, evaluated at 0.5B and 1.5B parameter scales. 
    \item \textbf{OpenUni {(Continuous)}}~\citep{openuni} generates in continuous latent space via diffusion, with InternVL3~\citep{zhu2025internvl3} / VAE~\citep{vae} as the understanding / generation encoder. It serves as an open-source counterpart of MetaQueries~\citep{metaqueries}; we evaluate both 1.6B and 3.6B variants at 512$\times$512 resolution without GPT-4o-Image distillation. 
    \item \textbf{BAGEL {(Continuous)}}~\citep{bagel} also employs continuous diffusion generation, with SigLIP~\citep{siglip} / VAE~\citep{vae} as the understanding / generation encoder. We report the results of BAGEL at 1024$\times$1024 resolution.
\end{itemize}

Together, these models cover the main families of UMM generation mechanisms: {discrete token prediction, masked autoregressive (MAR), and continuous diffusion}.

\textbf{Evaluation details.} 
We evaluate text-to-image generation on GenEval~\citep{geneval} and DPGBench~\citep{dpgbench}, and image editing on ImgEdit~\citep{ye2025imgedit} and GEdit-Bench-EN~\citep{liu2025step1x}. 
Our baselines include both generation-only models (e.g., SDXL~\citep{sdxl}, DALL-E 3~\citep{dalle3}) and unified multimodal models (e.g., Show-o~\citep{showo}, Harmon~\citep{wu2025harmon}, BAGEL~\citep{bagel}, GPT-4o-Image~\citep{openai2024introducing4o}). 

\textbf{Training data.} 
We post-train UMMs with high-quality open-source data, including MidjourneyV6~\citep{midjourneyv6}, LLaVA Mix-665K~\citep{llava}, and 10,000 FLUX-generated samples~\citep{texttoimage2M}. Further implementation, evaluation and dataset details are provided in Appendix~\ref{sec:setup}. To ensure fair comparison, our main experiments \textbf{exclude GPT-4o-Image distillation data like BLIP3o-60k}, which can inflate benchmark scores due to \emph{GenEval Template Leakage}. See Appendix~\ref{sec:geneval_leakage} for further discussion.

\begin{table*}[t!]
    \centering
    \caption{\textbf{\ours consistently improves image generation performance across unified multimodal models}. We report results for the largest model/resolution variant of each architecture; results for smaller models and detailed WISE scores are provided in Appendix~\ref{sec:quantitative_results}.}
    \vspace{-2mm}
    \footnotesize
    \setlength{\tabcolsep}{1.pt}
    \renewcommand\arraystretch{0.9}
    \label{tab:model_improvement}
    
    \begin{tabular}{lclllllllll}
        % \toprule
        \multirow{2}{*}{Model} & \multirow{2}{*}{\ours} & \multicolumn{7}{c}{GenEval} & \multirow{2}{*}{DPG} & \multirow{2}{*}{WISE}\\
        \cmidrule(lr){3-9}
        & & \multicolumn{1}{c}{Single} & \multicolumn{1}{c}{Two} & \multicolumn{1}{c}{Count} & \multicolumn{1}{c}{Color} & \multicolumn{1}{c}{Position} & \multicolumn{1}{c}{Attri.} & \multicolumn{1}{c}{Overall} & & \\

        \toprule[0.8pt]
        \multirow{2}{*}{Show-o} & \xmark & 97.2 & 80.3 & 61.9 & 78.2 & 27.3 & 52.3 & 66.2 & 82.21 & 0.40 \\
        & \cellcolor{customgreen}\! \cmark & \cellcolor{customgreen}\bf 98.2 \textgr{\tiny\!(+1.0)} & \cellcolor{customgreen}\bf 90.6 \textgr{\tiny\!(+10.3)} & \cellcolor{customgreen}\bf 66.8 \textgr{\tiny\!(+4.9)} & \cellcolor{customgreen}\bf 84.1 \textgr{\tiny\!(+5.9)} & \cellcolor{customgreen}\bf 37.4 \textgr{\tiny\!(+10.1)} & \cellcolor{customgreen}\bf 56.8 \textgr{\tiny\!(+4.5)} & \cellcolor{customgreen}\bf 72.3 \textgr{\tiny\!(+6.1)} & \cellcolor{customgreen}\bf 84.94 \textgr{\tiny\!(+2.73)} & \cellcolor{customgreen}\bf 0.40 \textbf{\tiny\!(0.00)} \\

        \toprule[0.4pt]
        \multirow{2}{*}{OpenUni} & \xmark & \bf 99.1 & 71.8 & 51.9 & 83.9 & 23.3 & 41.6 & 61.9 & 79.02 & 0.43 \\
        & \cellcolor{customgreen}\! \cmark & \cellcolor{customgreen}\bf 99.1 \tiny\!(0.0) & \cellcolor{customgreen}\bf 92.7 \textgr{\tiny\!(+20.9)} & \cellcolor{customgreen}\bf 52.3 \textgr{\tiny\!(+0.4)} & \cellcolor{customgreen}\bf 87.1 \textgr{\tiny\!(+3.2)} & \cellcolor{customgreen}\bf\! 43.8 \textgr{\tiny\!(+20.5)} & \cellcolor{customgreen}\bf 70.3 \textgr{\tiny\!(+28.7)} & \cellcolor{customgreen}\bf 74.1 \textgr{\tiny\!(+12.2)} & \cellcolor{customgreen}\bf 82.75 \textgr{\tiny\!(+3.73)} & \cellcolor{customgreen}\bf 0.54 \textgr{\tiny\!(+0.11)} \\

        \toprule[0.4pt]
        \multirow{2}{*}{Harmon} & \xmark & 99.4 & 87.3 & 68.7 & 86.4 & 44.9 & 51.1 & 72.9 & 80.93 & 0.41 \\
        & \cellcolor{customgreen}\! \cmark & \cellcolor{customgreen}\bf 99.9 \textgr{\tiny\!(+0.5)} & \cellcolor{customgreen}\bf 97.7 \textgr{\tiny\!(+10.4)} & \cellcolor{customgreen}\bf 71.4 \textgr{\tiny\!(+2.7)} & \cellcolor{customgreen}\bf 92.6 \textgr{\tiny\!(+6.2)} & \cellcolor{customgreen}\bf {75.7 \textgr{\tiny\!(+30.8)}} & \cellcolor{customgreen}\bf {76.6 \textgr{\tiny\!(+25.5)}} & \cellcolor{customgreen}\bf 85.7 \textgr{\tiny\!(+12.8)} & \cellcolor{customgreen}\bf 87.21 \textgr{\tiny\!(+6.28)} & \cellcolor{customgreen}\bf 0.50 \textgr{\tiny\!(+0.09)} \\

        \toprule[0.4pt]
        \multirow{2}{*}{BAGEL} & \xmark & 99.1 & 93.0 & 79.9 & 86.0 & 51.3 & 63.4 & 78.8 & 84.03 & 0.50 \\
        & \cellcolor{customgreen}\! \cmark & \cellcolor{customgreen}\bf 99.3 \textgr{\tiny\!(+0.2)} & \cellcolor{customgreen}\bf 93.9 \textgr{\tiny\!(+0.9)} & \cellcolor{customgreen}\bf 80.3 \textgr{\tiny\!(+0.4)} & \cellcolor{customgreen}\bf 87.6 \textgr{\tiny\!(+1.6)} & \cellcolor{customgreen}\bf 60.8 \textgr{\tiny\!(+9.5)} & \cellcolor{customgreen}\bf 72.6 \textgr{\tiny\!(+9.2)} & \cellcolor{customgreen}\bf 82.4 \textgr{\tiny\!(+3.6)} & \cellcolor{customgreen}\bf 85.29 \textgr{\tiny\!(+1.26)} & \cellcolor{customgreen}\bf 0.52 \textgr{\tiny\!(+0.02)} \\
        
        % \bottomrule
    \end{tabular}
    \vspace{-1mm}
\end{table*}

\begin{table*}[t]
    \centering
    \caption{\textbf{Results on image editing benchmarks}. We compare \ours post-trained BAGEL with previous methods. All scores are local reproduced. The \textcolor{gray}{gray-colored} rows denote private models, and their results are cited from~\citep{bagel,ye2025imgedit}.}
    \vspace{-2mm}
    \footnotesize
    \setlength{\tabcolsep}{3.4pt}
    \renewcommand\arraystretch{0.9}
    \label{tab:editing_results}
    
    \begin{tabular}{lccccccccccccc}
        % \toprule
        \multirow{2}{*}{Method} & \multicolumn{10}{c}{ImgEdit} & \multicolumn{3}{c}{GEdit-Bench-EN} \\
        \cmidrule(lr){2-11} \cmidrule(lr){12-14}
        & Bg. & Style & Adj. & Ext. & Rem. & Rep. & Add & Comp. & Act. & Ovr. & SC & PQ & Overall \\
        \toprule[0.8pt]
        \textcolor{gray}{Gemini 2.0} & \textcolor{gray}{-} & \textcolor{gray}{-} & \textcolor{gray}{-} & \textcolor{gray}{-} & \textcolor{gray}{-} & \textcolor{gray}{-} & \textcolor{gray}{-} & \textcolor{gray}{-} & \textcolor{gray}{-} & \textcolor{gray}{-} & \textcolor{gray}{5.43} & \textcolor{gray}{6.78} & \textcolor{gray}{5.36}  \\ 
        \textcolor{gray}{GPT-4o-Image} & \textcolor{gray}{4.57} & \textcolor{gray}{4.93} & \textcolor{gray}{4.33} & \textcolor{gray}{2.90} & \textcolor{gray}{3.66} & \textcolor{gray}{4.35} & \textcolor{gray}{4.61} & \textcolor{gray}{3.96} & \textcolor{gray}{4.89} & \textcolor{gray}{4.20} & \textcolor{gray}{7.85} & \textcolor{gray}{7.62} & \textcolor{gray}{7.53} \\
        \toprule[0.4pt]
        \color{black} BAGEL-NHR & 3.56 & 4.43 & 3.62 & 1.57 & 3.17 & 3.98 & 4.12 & 2.88 & 3.85 & 3.48 & 8.04 & 6.87 & 7.08 \\
        FLUX-Kontext & \textbf{3.89} & 4.62 & 3.69 & \textbf{1.81} & 2.97 & 4.20 & 3.80 & \textbf{3.00} & 4.18 & 3.60 & 6.95 & \textbf{7.30} & 6.27 \\
        BAGEL & 3.38 & 4.53 & 3.58 & 1.49 & 3.15 & 3.82 & 3.71 & 2.64 & 4.21 & 3.38 & 7.96 & 6.64 & 6.94 \\
        % \midrule
        \toprule[0.4pt]
        \rowcolor[HTML]{F9FFF9}
        \bf BAGEL-\ours & 3.85 & \textbf{4.73} & \textbf{3.86} & 1.68 & \textbf{3.75} & \textbf{4.28} & \textbf{4.20} & 2.94 & \textbf{4.56} & \textbf{3.75} & \textbf{8.24} & 6.87 & \textbf{7.27} \\
        \rowcolor[HTML]{F9FFF9}
        \textit{\versus baseline} & \bf \textgr{\scriptsize +0.47} & \bf \textgr{\scriptsize +0.20} & \bf \textgr{\scriptsize +0.28} & \bf \textgr{\scriptsize +0.19} & \bf \textgr{\scriptsize +0.60} & \bf \textgr{\scriptsize +0.46} & \bf \textgr{\scriptsize +0.49} & \bf \textgr{\scriptsize +0.30} & \bf \textgr{\scriptsize +0.35} & \bf \textgr{\scriptsize +0.37} & \bf \textgr{\scriptsize +0.28} & \bf \textgr{\scriptsize +0.23} & \bf \textgr{\scriptsize +0.33} \\
        % \bottomrule
    \end{tabular}
    \vspace{-2mm}
\end{table*}

\subsection{\ours Achieves SOTA Performance on Image Generation and Editing}

\textbf{SOTA performance without GPT-4o-Image distillation.} 
As shown in Table~\ref{tab:main_table}, post-training the Harmon-1.5B model with \ours achieves SOTA overall scores on GenEval and DPGBench. Despite using significantly fewer parameters, \ours surpasses recent methods, \eg, Janus-Pro~\citep{chen2025janus} and BAGEL~\citep{bagel}. % The \emph{Counting} subtask remains our main shortfall and Appendix~\ref{sec:limitations} analyzes this gap in detail.

\textbf{Image editing results are also substantially improved}. As shown in Table~\ref{tab:editing_results}, \ours consistently outperforms existing baselines across all subtasks and benchmarks. 
On ImgEdit, \ours achieves an overall score of 3.75, surpassing FLUX-Kontext (3.60) and the baseline (3.38). On GEdit-Bench-EN, \ours reaches 7.27, which closes the gap between open-source models and GPT-4o-Image. In particular, with only 10,000 unlabeled images and 27 GPU hours, \ours beats the concurrent work BAGEL-NHR~\citep{kuprashevich2025nohumansrequiredautonomoushighqualityimage} (3.48), which employs supervised fine-tuning (SFT) on 300,000 high-quality image editing datat.

\subsection{\ours is Effective Across Different UMM Frameworks}

\begin{figure}[t]
    \vspace{-2pt}
    \centering
    \includegraphics[width=\textwidth]{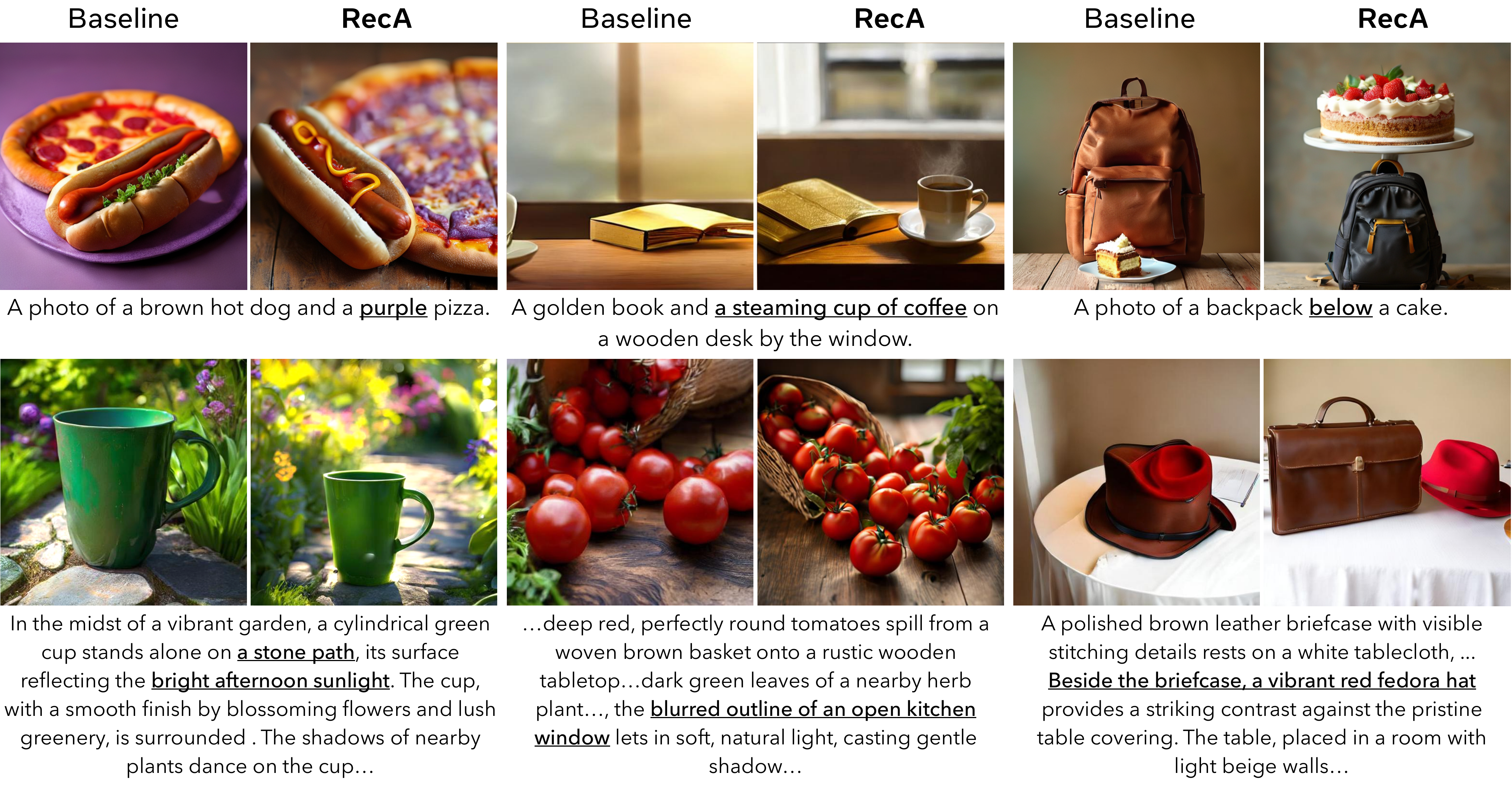}
    \vspace{-18pt}
    \caption{\textbf{Image generation results \versus baselines.} We use Harmon-1.5B as baseline. The post-trained model better handles multiple objects, complex attributes and spatial layouts, preserving fine details missed by the baseline. 
    % \tjd{consider moving one example from figure 6 and one from figure 7 to right side of figure 1 if you are able to merge the current charts in figure 1}.  In each pair, the left image is generated by the baseline model and the right by the \ours post-trained model. 
    }
    \label{fig:baseline_vs_finetuned}   
    \vspace{-5pt}

\end{figure}

\begin{figure}[t]
    \centering
    \includegraphics[width=\textwidth]{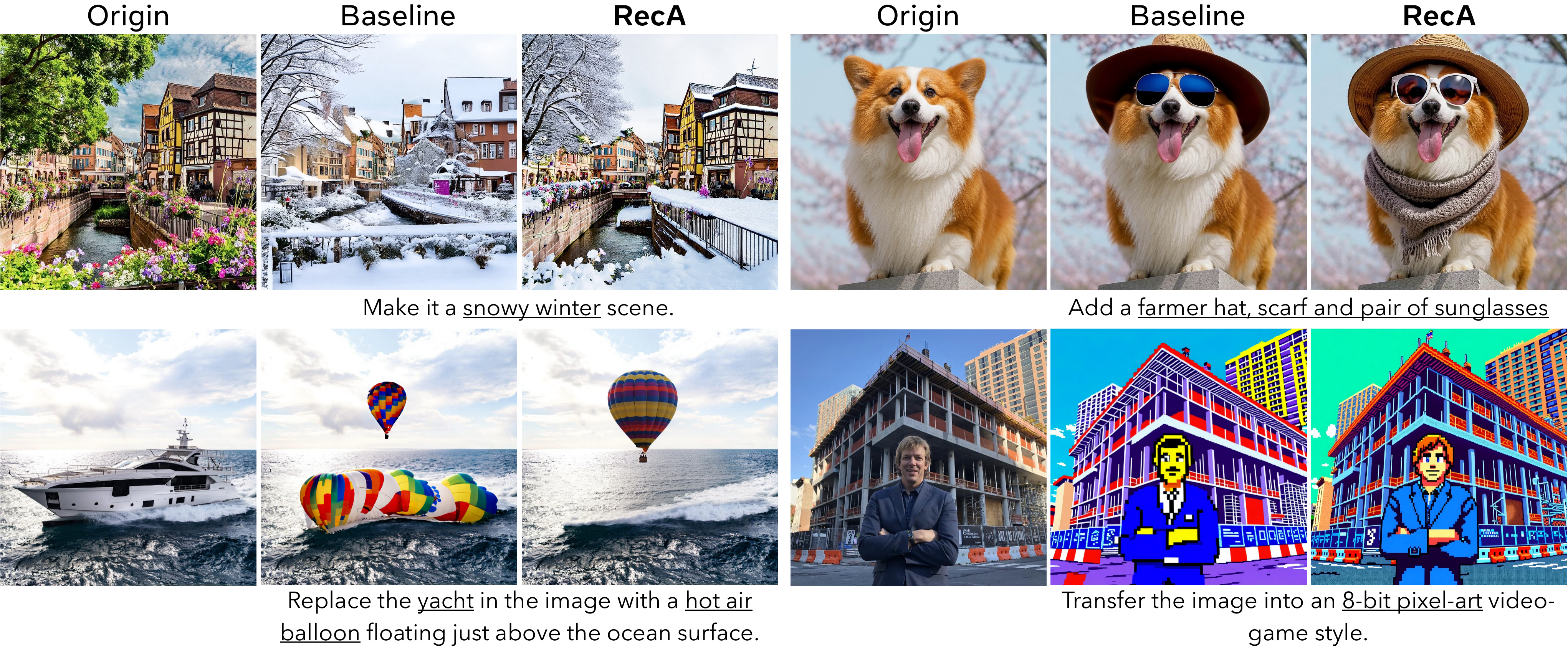}
    \vspace{-18pt}
    \caption{\textbf{Image editing results \versus baselines}. We use BAGEL as baseline. Our model consistently improves performance on object addition, replacement, style transfer, and scene modification.} % In each pair, the left image is generated by the baseline model and the right by the \ours post-trained model. 
    \label{fig:imgedit}
    \vspace{-6pt}
\end{figure}

\textbf{Consistent performance gains across UMM architectures.} As shown in Table~\ref{tab:model_improvement}, \ours demonstrates consistent and significant improvements across all evaluated UMM frameworks. We report results for the largest model variant of each architecture, with smaller model results provided in Appendix~\ref{sec:quantitative_results}. On GenEval and DPGBench, the most substantial improvement is achieved by Harmon-1.5B, which attains scores of 85.7 (+12.8) and 87.21 (+6.28), respectively.

\textbf{Performance on WISE benchmark.} WISE~\citep{wisebench} evaluates text-to-image reasoning with 1,000 knowledge-puzzle prompts (\eg, Einstein's favorite instrument). As shown in Table~\ref{tab:model_improvement}, \ours improves Harmon and OpenUni, while BAGEL and Show-o show limited or no gains. We conclude that \ours primarily improves semantic-level alignment, while reasoning ability remains an open problem for future work. Detailed WISE results are provided in Appendix~\ref{sec:quantitative_results}.

\subsection{More Results}

% \textbf{Qualitative results.} Figure~\ref{fig:baseline_vs_finetuned} and Figure~\ref{fig:imgedit} demonstrate consistent improvements after \ours. The baseline model fails on cases of \emph{multiple objects}, \emph{complex attributions}, and explicit \emph{spatial layouts}, while \ours post-trained model consistently succeeds. For dense prompts, the post-trained model preserves fine-grained details (e.g., kitchen window, sunlight) that the baseline model blurs or omits. Comprehensive qualitative results for both image generation and editing tasks are provided in Appendix~\ref{sec:more_qualitative_results}. For image editing, we observe consistent performance improvements across diverse tasks including object addition, replacement, style transfer, and color modification. Additional qualitative editing results are presented in Appendix~\ref{sec:more_image_editing}.

\textbf{Qualitative results.} For image generation, the baseline often fails on \emph{multiple objects}, \emph{complex attributes}, and \emph{spatial layouts}, and in Figure~\ref{fig:baseline_vs_finetuned}, the post-trained model preserves these fine-grained details well (e.g., kitchen window, sunlight). For image editing, Figure~\ref{fig:imgedit} shows consistent improvements across tasks such as object addition, replacement, style transfer, and color modification. Additional examples are provided in Appendix~\ref{sec:more_image_editing} and~\ref{sec:more_qualitative_results}.

\setlength{\intextsep}{2pt} % or a small value like 2pt
\setlength{\columnsep}{8pt} % default is ~10pt
\begin{wraptable}[4]{r}{0.58\textwidth}
    \centering
    \vspace{-1.3mm}
    \caption{Visual understanding performance.}
    \vspace{-3mm}
    \footnotesize
    \setlength{\tabcolsep}{2pt}
    \renewcommand\arraystretch{0.9}
    \label{tab:visual_understanding}
    \begin{tabular}{lcccccc}
        % \toprule
        Model & MME & POPE \tiny{(Acc)} & POPE \tiny{(F1)} & GQA & MMMU & SEED \\
        \toprule[0.8pt]
        Harmon & 1195 & 83.8 & \textbf{83.9} & \textbf{58.8} & 34.7 & 65.2 \\
        \rowcolor{customgreen}
        \ours & \textbf{1223} & \textbf{83.9} & 83.2 & 58.4 & \textbf{35.7} & \textbf{65.3} \\
        % \bottomrule
    \end{tabular}
    % \vspace{-12mm}
\end{wraptable}

\begin{table}[t]
\centering
\begin{minipage}{0.65\textwidth}
    \footnotesize
    \vspace{-9pt}
    \setlength{\tabcolsep}{6pt}
    \renewcommand\arraystretch{0.9}
    \begin{tabular}{lcccc}
        Dataset & SFT & \ours & GenEval & DPG \\
        \toprule[0.8pt]
        \rowcolor[HTML]{F1F9FF}
        MidjourneyV6 & \cmarkColor & \xmark & 74.76 & 80.89 \\
        % \rowcolor{gray!12}
        MidjourneyV6 (dense) & \cmarkColor & \xmark & 74.05 & 80.67 \\
        % \rowcolor{gray!12}
        MidjourneyV6 & \xmark & \cmarkColor & \textbf{85.69} & \textbf{87.21} \\
        % \rowcolor{gray!12}
        MidjourneyV6 & \cmarkColor & \cmarkColor & 81.22 & 86.54 \\
        \toprule[0.8pt]
        \rowcolor[HTML]{F9FFF9}
        BLIP3o-60k & \cmarkColor & \xmark & 84.95 & 85.19 \\
        % \rowcolor{gray!12}
        BLIP3o-60k & \xmark & \cmarkColor & 85.21 & \textbf{86.50} \\
        % \rowcolor{gray!12}
        BLIP3o-60k & \cmarkColor & \cmarkColor & \textbf{86.14} & 85.96 \\
        \toprule[0.8pt]
        \rowcolor[HTML]{FFFDF5}
        BLIP3o-60k (w/o GenEval) & \cmarkColor & \xmark & 80.88 & 85.24 \\
        % \rowcolor{gray!12}
        BLIP3o-60k (w/o GenEval) & \xmark & \cmarkColor & \textbf{84.76} & \textbf{86.37} \\
    \end{tabular}
\end{minipage}%
\hspace{2pt}%
\begin{minipage}{0.34\textwidth}
    \centering
    \vspace{2pt}
    \includegraphics[width=\linewidth]{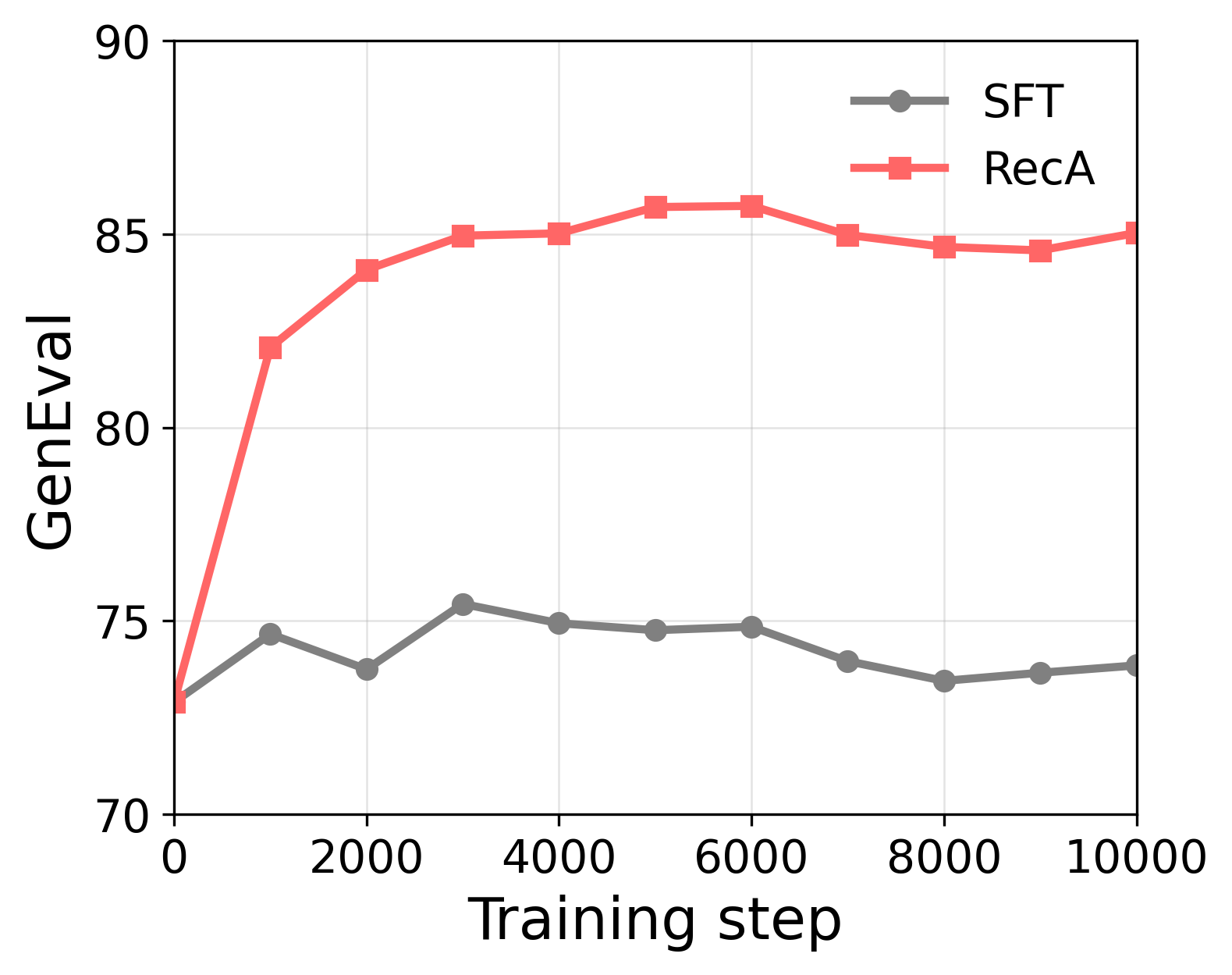}
\end{minipage}
    \vspace{-10pt}
    \caption{\textbf{Comparison of SFT and \ours across post-training datasets.} \ours consistently outperforms SFT, especially when \emph{template leakage data} is excluded (\textbf{yellow row}), demonstrating that the improvements are genuine and robust. All results are reported at 5k steps. {\textbf{Right:} GenEval overall scores of Harmon-1.5B with SFT \versus \ours across training steps.}}
    \vspace{-10pt}
    \label{tab:sft_realign_comparison}
\end{table}

% \textbf{Visual understanding results.} To verify that \ours preserves the visual understanding capabilities of UMMs, we evaluate our post-trained Harmon-1.5B on benchmarks including POPE~\citep{pope}, MME~\citep{mme}, GQA~\citep{gqa}, MMMU~\citep{mmmu}, and SEED~\citep{seedbench}. As shown in Table~\ref{tab:visual_understanding}, performance remains stable, with only minor variations within the typical range of fine-tuning. These results indicate that \ours improves generation fidelity with visual understanding maintained. For architectures with frozen understanding components (e.g., OpenUni, BAGEL), performance is unchanged by design, while share-parameter models also show no degradation.

\textbf{Visual understanding results.} We investigate the effect of \ours on UMMs that share parameters between understanding and generation (e.g., Harmon-1.5B). In Table~\ref{tab:visual_understanding}, results on POPE~\citep{pope}, MME~\citep{mme}, GQA~\citep{gqa}, MMMU~\citep{mmmu}, and SEED~\citep{seedbench} show that \ours can improve UMMs' generation fidelity without degrading visual understanding. For architectures with decoupled understanding components (OpenUni, BAGEL), their capability remains unchanged.

\setlength{\intextsep}{12pt}   % default vertical space around in-text floats
\setlength{\columnsep}{10pt}   % default horizontal gap between float and text

\subsection{Empirical Studies of \ours as a Post-Training Method}

\textbf{SFT vs. \ours as post-training methods.}  
We directly compare supervised fine-tuning (SFT) with \ours as alternative post-training strategies for UMMs. Unlike SFT, which depends on text–image pairs, \ours is self-supervised and requires only unlabeled images. Our key questions are: \emph{(i) which method is more effective, and (ii) when and how should \ours be applied relative to SFT?}

\begin{wraptable}[7]{r}{0.47\textwidth}
    \centering
    \vspace{-3.6mm}
    \caption{\textbf{Training recipe.} SFT \versus \ours as sequential post-training stages.}
    \vspace{-2.5mm}
    \footnotesize
    \setlength{\tabcolsep}{4pt}
    \renewcommand\arraystretch{0.9}
    \label{tab:training_order}
    \begin{tabular}{lccc}
        \ours Dataset & Order & GenEval & DPG \\
        \midrule
        BLIP3o-60k & \ours $\rightarrow$ SFT & 85.91 & 85.67 \\
        BLIP3o-60k & SFT $\rightarrow$ \ours & 89.00 & 87.50 \\
        \rowcolor{customgreen}
        MidjourneyV6 & SFT $\rightarrow$ \ours & \textbf{90.15} & \textbf{88.15} \\
    \end{tabular}
    % \vspace{-14mm}
\end{wraptable}

\textit{\ours demonstrates genuine effectiveness compared with SFT}. 
As shown in Table~\ref{tab:sft_realign_comparison}, on MidjourneyV6, SFT achieves 74.76 on GenEval. Adding denser captions via Midjourney-LLaVA~\citep{midjourneyv6llava} does not help; GenEval remains nearly unchanged and DPGBench even drops a little. In contrast, replacing SFT with \ours dramatically boosts performance under the same setup (85.69/87.21). On the complete BLIP3o-60k dataset, \ours still outperforms SFT on both benchmarks. Notably, removing \emph{template leakage data} leads to a performance drop for SFT (-4.07 on GenEval), whereas \ours maintains strong performance. {The training dynamics in the right panel of Table~\ref{tab:sft_realign_comparison} further demonstrate that \ours consistently outperforms SFT. While SFT performance stagnates, \ours rapidly ascends to a superior level and maintains a stable advantage throughout the training process; the evolution of GenEval subtasks is provided in Appendix~\ref{sec:geneval_subtasks}.}

% This demonstrates that self-supervised semantic reconstruction is a far more effective post-training strategy than supervised fine-tuning. Further analysis of SFT on BLIP3o-60k is provided in Appendix~\ref{sec:more_results_analysis}.

\textit{When to apply \ours?}  
We also study training order to understand when to apply \ours relative to SFT. As shown in Table~\ref{tab:sft_realign_comparison}, mixed training (half SFT, half \ours per batch) is unstable: it slightly helps on BLIP3o-60k but fails on MidjourneyV6. If we do SFT on BLIP3o-60k, sequential training is more decisive: as shown in Table~\ref{tab:training_order}, applying \ours after SFT (SFT$\rightarrow$\ours) consistently delivers the best results (90.15/88.15 on MidjourneyV6). Reversing the order (\ours$\rightarrow$SFT) degrades performance (-3.09/-1.83). This asymmetry underscores that SFT provides broad text–image alignment, while \ours serves best as a refinement stage enhancing semantic grounding and visual faithfulness.

\textbf{Visual understanding \versus visual generation encoder.}  
Most UMMs employ two encoders: one for understanding (semantic features) and one for generation (pixel-level features). BAGEL follows this design, making it a natural testbed for \ours.  
As shown in Table~\ref{tab:bagel_vision_encoder}, conditioning on the generation encoder (VAE) yields only marginal or degraded gains, while embeddings from the understanding encoder (ViT) produce consistently better results across GenEval, DPGBench, ImgEdit, and GEdit. This indicates that \ours benefits most from semantic embeddings that capture high-level conceptual information, rather than raw visual details.

\begin{wraptable}[9]{r}{0.57\textwidth} 
\vspace{-0.4cm}
\centering
% \caption{Using semantic embeddings from visual understanding encoders consistently outperforms generation encoder latents across benchmarks.}
\caption{\textbf{Vision encoder type and input resolution comparison}. Embeddings from visual understanding encoder with the lowest resolution are the best for \ours.}
\label{tab:bagel_vision_encoder}
\vspace{-0.3cm}
\footnotesize
\setlength{\tabcolsep}{2pt}
\begin{tabular}{lccccc}
% \toprule
Vision Encoder & Resolution & GenEval & DPG & ImgEdit & GEdit \\
\toprule[0.8pt]
Baseline & - & 78.8 &  84.03 & 3.38 & 6.94 \\
\toprule[0.4pt]
VAE (gen.) & $256\times 256$ & 78.5 & 83.92 & 3.63 & 7.08 \\ 
\rowcolor{customgreen}
ViT (und.) & $224\times 224$ & \textbf{82.4} & \textbf{85.29} & \textbf{3.75} & \textbf{7.27} \\
ViT (und.) & $512\times 512$ & 79.2 & 84.61 & 3.68 & 7.18 \\
% \bottomrule
\end{tabular}
\end{wraptable}

\textbf{Different resolutions for the understanding encoder.} For Show-o, Harmon and OpenUni, the input resolutions are fixed at $336\times 336$, $512\times512$ and $448\times 448$, respectively. BAGEL uses NaViT~\citep{dehghani2023patch} and can flexibly accommodate arbitrary input resolutions. We perform experiments on $224\times224$ and $512\times512$ input resolutions for \ours, and the results in Table~\ref{tab:bagel_vision_encoder} show that the model post-trained on $224\times224$ consistently outperforms its $512\times512$ counterpart across all benchmarks.{Prior work shows that higher-resolution embeddings retain substantially more pixel-level detail~\citep{allakhverdov2025image,lin2025uniworld}, which in turn drives the model to lean on low-level cues rather than the semantic features that \ours is designed to emphasize.}

\textbf{Conclusion.} Across data and strategies, \ours consistently proves to be a stronger post-training method than SFT. 
\textbf{The best post-training recipe is a two-stage pipeline}: first SFT on high-quality paired data for coarse alignment, followed by \ours for self-supervised fine-grained refinement.

%% file: sections/03related_work.tex
\section{Related Work}

\subsection{Unified Multimodal Models (UMMs)}

Unified multimodal model (UMM) is a backbone capable of multimodal understanding and generation within a single model or system~\citep{zhang2025unifiedmultimodalunderstandinggeneration}. 
Recent studies have explored different types of UMMs: {\textbf{(I) Discrete UMM}}. Models like Chameleon tokenize images and predict them autoregressively in a next-token prediction way~\citep{team2024chameleon,qu2024tokenflow,janus,liquid}, while Show-o~\citep{showo} introduces a \emph{discrete-diffusion} schedule to improve the token prediction process. {\textbf{(II) Continuous UMM}}. Models like Transfusion attach a diffusion (or flow-matching)~\citep{ddpm,flow} head to the shared transformer~\citep{zhou2025transfusion,ma2024janusflow}, while some UMMs keep pretrained MLLMs frozen and route its intermediate features with learnable queries to an external image generator~\citep{metaqueries,lin2025uniworld}. {\textbf{(III) MAR UMM}}. Masked-autoregressive (MAR) ~\citep{li2024autoregressive} is a novel autoregressive image generation paradigm, which has been adopted by models like Harmon~\citep{wu2025harmon,fan2025unified,MMAR,wang2025skyworkunipicunifiedautoregressive}.

\subsection{Post-training Strategies for UMMs}

Recent efforts explore different post-training techniques for enhancing the generation capability of UMMs. 
\textbf{(I) Chain-of-Thought~\citep{wei2022chain} or test-time verification} do reasoning before the generation or verify generated image step-by-step, but they depend on external models and do not improve the UMM's native generation capability. ~\citep{guo2025can,wang2025mint,fang2025got,tian2025unigen}. \textbf{(II) Reinforcement learning.} Methods such as DPO~\citep{rafailov2023direct} and GRPO~\citep{shao2024deepseekmath} optimize generation policies with human or automatic preference signals, but require curated paired data and carefully tuned advantage functions. ~\citep{wei2025skywork,han2025self,yan2025unified,mao2025unirl,tian2025unigen,jiang2025co}. \textbf{(III) High-quality synthetic data.} Recent works~\citep{chen2025blip3,wang2025gptimageedit15mmillionscalegptgeneratedimage,ye2025echo} construct large-scale synthetic image–text pairs for supervised fine-tuning, which boost benchmarks but incur high data-generation cost and potential distribution shift.

%% file: sections/06conclusion.tex
\begin{figure}[t]
    \centering
    % \vspace{-2mm}
    \includegraphics[width=\textwidth]{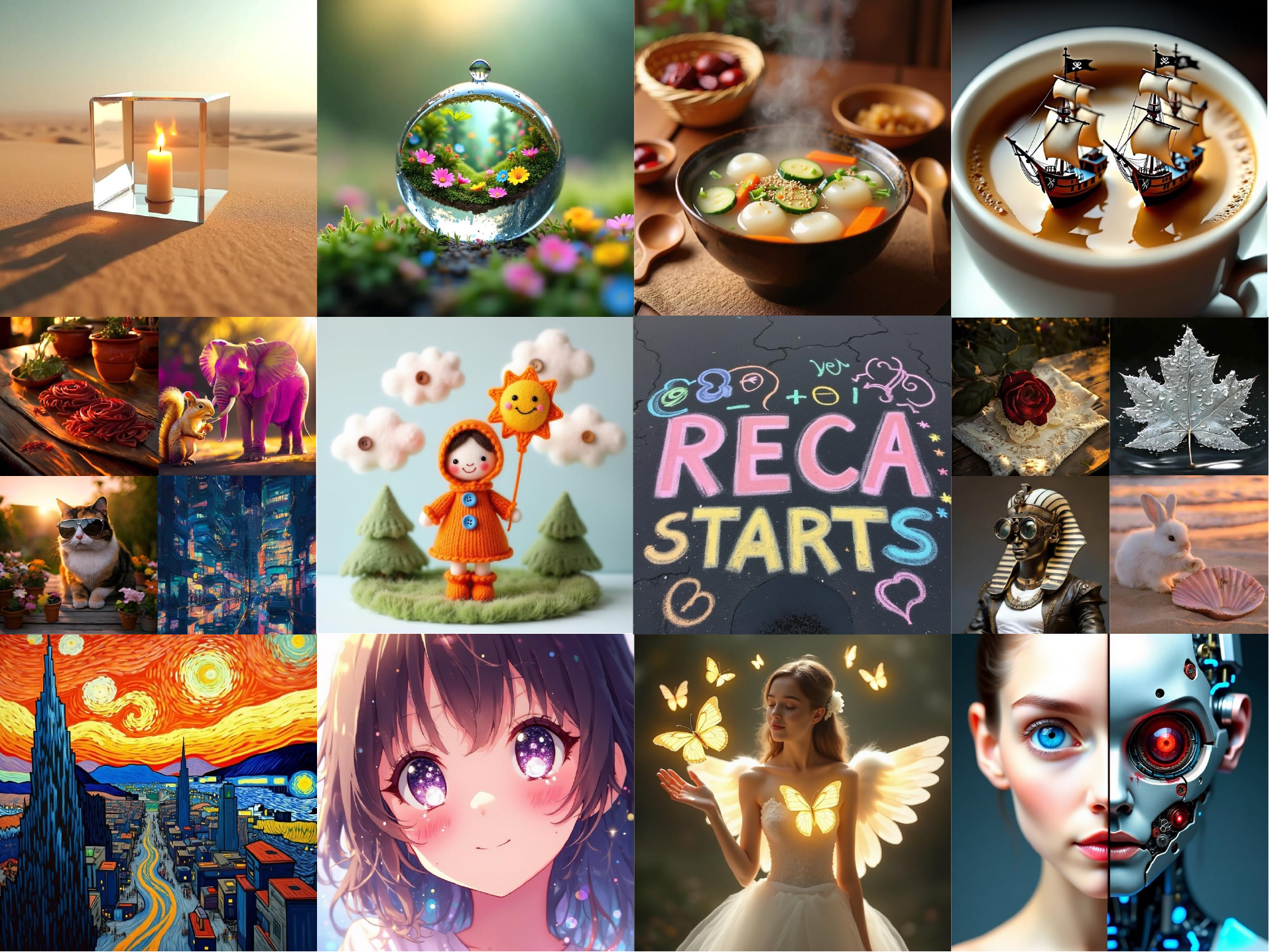}
    \vspace{-5mm}
    \caption{\textbf{Qualitative T2I results}. The large images ($1024 \times 1024$) are generated by the post-trained BAGEL, while the small images ($512 \times 512$) are generated by the post-trained Harmon. Detailed captions are listed in Appendix~\ref{sec:generated_captions}.
    }
    \label{fig:demo_t2i}
\end{figure}

\begin{figure}[t]
    % \vspace{-0.5em}
    \centering
    \includegraphics[width=\textwidth]{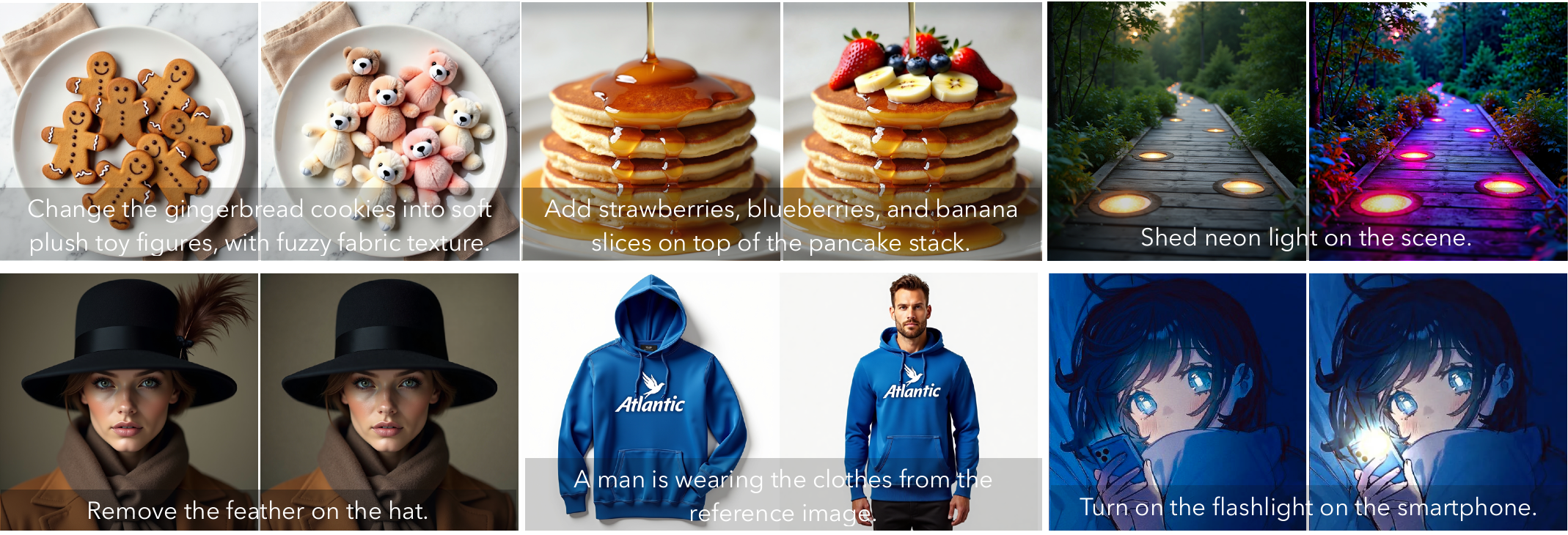}
    \vspace{-5mm}
    \caption{\textbf{Qualitative image editing results.} 
    Edited outputs are generated by BAGEL post-trained with \ours. 
    Left: original images; Right: edited images.}
    \label{fig:demo_edit}
    \vspace{-4mm}
\end{figure}

\section{Conclusion}

In this work, we propose \textbf{\ours}, a lightweight post-training method that replaces sparse text-to-image supervision with dense features from the model's own visual understanding encoder. 
\ours requires no extra caption data yet significantly improves image generation and editing across architectures. 
We discuss the limitations and future directions in Appendix~\ref{sec:limitations}. 

% We sincerely thank Ishan Misra, Xingyi Zhou, Haoqi Fan, Mannat Singh, Dewei Zhou, Haotian Tang, Yulun Wu, Baifeng Shi, Jiaxin Ge, Yuwei Niu, Xun Wang, Weiyang Jin, Xichen Pan, Jiaqi Liao, Jiuhai Chen for their insightful discussions and valuable feedback on our paper. We especially thank Size Wu and Yang Ye for their assistance in reproducing the baseline results.

\subsubsection*{Reproducibility Statement}
% We facilitate reproducibility by providing all training and evaluation hyper-parameters and implementation details in the Appendix~\ref{sec:setup}. These materials are sufficient to reproduce our experiments without additional assumptions. Detailed prompt templates are attached in Supplementary Material.
% We are committed to open-source research. 
We ensure reproducibility by providing all training and evaluation hyperparameters, along with full implementation details, in Appendix~\ref{sec:setup}. 
These materials are sufficient to replicate our experiments. 
Detailed prompt templates are included in the Supplementary Material. 
In line with our commitment to open-source research, we will make our codes and model weights publicly available.

%% file: sections/appendix.tex
\newpage

\appendix

% \newpage

We provide additional information in the supplementary material, as outlined below:

\begin{itemize}[leftmargin=16pt]
    \item \textbf{Sec.}~\ref{sec:preliminaries}: Brief introduction to different image generation paradigms.
    \item \textbf{Sec.}~\ref{sec:difference_cfg}: Difference Between \ours and Classifier-Free Guidance (CFG).
    \item \textbf{Sec.}~\ref{sec:templates}: Prompt templates for \ours.
    \item {\textbf{Sec.}~\ref{sec:appendix_mechanism}: Motivation and Mechanism Analysis of \ours.}
    \item \textbf{Sec.}~\ref{sec:setup}: Further implementation details for \ours implementation and evaluation.
    \item \textbf{Sec.}~\ref{sec:more_results_analysis}: {Additional experimental results and analysis}.
    \item \textbf{Sec.}~\ref{sec:limitations}: Limitations and future work for \ours.
    \item \textbf{Sec.}~\ref{sec:generated_captions}: Generated captions for figures in the DEMO.
    \item \textbf{Sec.}~\ref{sec:more_image_editing}: Additional qualitative results on image editing tasks.
    \item \textbf{Sec.}~\ref{sec:more_qualitative_results}: More uncurated qualitative examples for text-to-image generation.
    \item \textbf{Sec.}~\ref{sec:llm_usage}: The Use of Large Language Models (LLMs).
\end{itemize}

\section{Preliminaries of Image Generation Paradigms}

\label{sec:preliminaries}

\subsection{{Discrete} Paradigm}

{Discrete paradigms} use image tokenizer~\citep{vqvae,vqgan,chang2022maskgit} and treat images as a sequence of discrete tokens, predicting each token in turn. The joint likelihood factorizes as:

\begin{equation}
\begin{gathered}
    p(x) = \prod_{i=1}^N p_\theta(x_i \mid \mathbf{x}_{<i}),\\
    \mathcal{L}_{\mathrm{AR}} = -\mathbb{E}_{x\sim p_{\mathrm{data}}} \sum_{i=1}^N \log p_\theta(x_i \mid \mathbf{x}_{<i}),
\end{gathered}
\end{equation}

% \begin{figure}[t]
%     \centering
%     % \vspace{-2mm}
%     \includegraphics[width=\textwidth]{figures/DEMO.pdf}
%     \vspace{-5mm}
%     \caption{\textbf{Qualitative T2I results}. The large images ($1024 \times 1024$) are generated by the post-trained BAGEL, while the small images ($512 \times 512$) are generated by the post-trained Harmon. Detailed captions are listed in Appendix~\ref{sec:generated_captions}.
%     }
%     \label{fig:demo_t2i}
% \end{figure}

% \begin{figure}[t]
%     % \vspace{-0.5em}
%     \centering
%     \includegraphics[width=\textwidth]{figures/DEMO_edit.pdf}
%     \vspace{-5mm}
%     \caption{\textbf{Qualitative image editing results.} 
%     Edited outputs are generated by BAGEL post-trained with \ours. 
%     Left: original images; Right: edited images.}
%     \label{fig:demo_edit}
% \end{figure}

where the model is trained by minimizing the cross-entropy of each next-token prediction. At sampling time, tokens are drawn sequentially ($x_1\sim p_\theta(x_1)$, $x_2\sim p_\theta(x_2\mid x_1)$, etc.).  Models such as Chameleon~\citep{team2024chameleon} and Janus~\citep{janus} append it onto LLM's LM head.  

MaskGIT~\citep{chang2022maskgit} extends the {discrete paradigm} by parallel prediction with random masks.
Given a tokenized image sequence $\mathbf{y}=(y_1,\dots,y_N)$, where $N$ is the number of discrete tokens obtained from a image tokenizer's codebook of size $K$, and a binary mask vector $\mathbf{m}=(m_1,\dots,m_N)$ with $m_i=1$ indicating that position $i$ is masked, the model is trained with the masked prediction loss:

\begin{align}
    \mathcal{L}_{\rm mask} = -\mathbb{E}_{\mathbf{y},\mathbf{m}} \sum_{i=1}^N m_i \log p_\theta(y_i \mid \mathbf{y}^M) \,
\end{align}

where $\mathbf{y}^M$ denotes the input sequence with masked tokens replaced by a special $\texttt{[MASK]}$ symbol. During inference, the model starts from a fully masked input and iteratively fills it with predicted image tokens. At step $t$, the model predicts distributions $p_i^{t}=p_\theta(y_i \mid \mathbf{y}^{M,t})$ for all currently masked positions, where $p_i^{(t)}$ is a categorical distribution over the $K$ image tokens. The confidence score is defined as $c_i^{t}=\max_j p_{ij}^{t}$, i.e., the maximum probability across categories. A scheduling function $\gamma(t/T)$, where $T$ is the total number of refinement steps, controls the number of tokens to remain masked:

\begin{equation}
\begin{gathered}
    n_t = \lfloor N \cdot \gamma(t/T)\rfloor, \\
    m_i^{(t+1)} = \mathbb{1}\!\left[c_i^{(t)} \le \text{the $n_t$-th smallest confidence}\right].
\end{gathered}
\end{equation}

This process progressively decreases the mask ratio until the full image token sequence is generated. In contrast to autoregressive next-token prediction, MaskGIT decodes multiple tokens in parallel with bidirectional context, achieving faster sampling and higher generation fidelity. UMMs such as Show-o~\citep{showo} use this paradigm.

\subsection{{Continuous} Paradigm}

{The most representative method in the continuous paradigm} is diffusion~\citep{sohl2015deep,ddpm}, which learns a gradual noising process from Gaussian distribution. The forward (diffusion) process progressively corrupts a clean image $\mathbf x_0 \sim p_{\text{data}}$ into pure Gaussian noise $\mathbf x_T$:

\begin{equation}
\begin{gathered}
    q(\mathbf x_t \mid \mathbf x_{t-1}) = \mathcal{N}\!\bigl(\mathbf x_t; \sqrt{1-\beta_t}\,\mathbf x_{t-1},\,\beta_t \mathbf I\bigr), \\
    q(\mathbf x_t \mid \mathbf x_0) = \mathcal{N}\!\bigl(\mathbf x_t; \sqrt{\bar\alpha_t}\,\mathbf x_0,\,(1-\bar\alpha_t)\mathbf I\bigr),
\end{gathered}
\end{equation}

where $\beta_t\in (0,1)$ are pre-defined hyperparamters, $\alpha_t = 1-\beta_t$, and $\bar\alpha_t = \prod_{s=1}^t \alpha_s$. The model $\theta$ learns the reverse denoising distribution:

\begin{equation}
    p_\theta(\mathbf x_{t-1} \mid \mathbf x_t) = \mathcal{N}\!\bigl(\mathbf x_{t-1}; \mu_\theta(\mathbf x_t,t,z),\,\sigma_t^2 \mathbf I\bigr),
\end{equation}

and is trained with the noise prediction loss:

\begin{equation}
\begin{gathered}
    \mathcal{L}_{\mathrm{diff}} = \mathbb{E}_{\mathbf x_0,\epsilon,t,z}\Bigl[\|\epsilon - \epsilon_\theta(\mathbf x_t,t,z)\|^2\Bigr], \\
    \mathbf x_t = \sqrt{\bar\alpha_t}\,\mathbf x_0 + \sqrt{1-\bar\alpha_t}\,\epsilon,\quad \epsilon \sim \mathcal{N}(0,\mathbf I).
\end{gathered}
\end{equation}

Here $\epsilon_\theta(\mathbf x_t,t,z)$ is the predicted noise given the noisy input $\mathbf x_t$, timestep $t$ and condition signal $z$. Generation starts from $\mathbf x_T \sim \mathcal{N}(0, \mathbf I)$ and applies the learned reverse process until $\mathbf x_0$ is recovered.  

Instead of discrete noising steps, flow-matching~\citep{flow} defines a continuous trajectory $\{\mathbf x_t\}_{t\in[0,1]}$ that transports a prior distribution $p_0$ (e.g.\ Gaussian noise at $t=0$) into the data distribution $p_1\ (t=1)$. In ordinary differential equation (probability flow ODE):

\begin{align}
    \frac{\mathrm d}{\mathrm dt}\mathbf x(t) &= u_\theta(\mathbf x(t),t,z),
\end{align}

where $u_\theta$ is a neural velocity (vector) field. The ground-truth target field $u^*(x,t,z)$ can be derived from an optimal coupling between $p_0$ and $p_1$, and training minimizes:

\begin{align}
    \mathcal{L}_{\mathrm{flow}} = \mathbb{E}_{\mathbf x,t,z}\bigl[\|u_\theta(\mathbf x,t,z) - u^*(\mathbf x,t,z)\|^2\bigr].
\end{align}

Rectified flow~\citep{liu2022flow} simplifies this by defining a linear interpolation:

\begin{align}
    \mathbf x(t) &= (1-t)\mathbf x(0) + t \epsilon,\quad \epsilon \sim \mathcal{N}(0, \mathbf I),
\end{align}
so that the target velocity field is available in closed form:
\begin{align}
    u^*(\mathbf x(t),t,z) = \frac{\mathrm d}{\mathrm dt}\mathbf x(t) = \epsilon - \mathbf x(0).
\end{align}

Diffusion adds Gaussian noise at each step with variance $\beta_t$, whereas rectified flow directly defines a continuous interpolation between data and noise. UMMs such as BAGEL~\citep{zhou2025transfusion}, Janusflow~\citep{ma2024janusflow} and Metaqueries~\citep{metaqueries,openuni} append these paradigms above onto LLM.  

\subsection{Masked Autoregressive (MAR)}

MAR~\citep{li2024autoregressive} is a mask-based generation paradigm similar to MaskGIT, but it generalizes AR without vector quantization. Specifically, MAR employs a diffusion decoder to generate images with continuous tokens in the transformer. 

Inference also differs from MaskGIT: \textbf{(I)} MAR predicts continuous features instead of logits over discrete codes, and \textbf{(II)} while MaskGIT fills the most confident logits, MAR stochastically samples denoised patches at each step. UMMs such as Fluid~\citep{unifluid} and Harmon~\citep{wu2025harmon} follow this design.

% \paragraph{\ours is Paradigm-Agnostic.} 
% Across AR, AR+Diffusion, and MAR, generation can be seen as reconstructing high-dimensional visual features from partial or noisy observations. \ours introduces a unified semantic reconstruction objective that aligns all these paradigms without modifying their inherent training losses.

\section{Difference Between \ours and Classifier-Free Guidance (CFG)}
\label{sec:difference_cfg}
\textbf{Classifier-free guidance} (CFG~\citep{ho2022classifier}) is typically used to improve image generation fidelity. 
At each generation step, we compute a conditional prediction $\mathbf{o}_{\text{cond}}$ and an unconditional prediction $\mathbf{o}_{\text{uncond}}$, where $\mathbf{o}$ denotes either the autoregressive head's logits or the diffusion head's predicted noise, as we introduced in Sec.~\ref{sec:preliminaries}. The output is given by:

\begin{equation}
    \vspace{-2pt}
    \mathbf{o} = (1+\omega)\,\mathbf{o}_{\text{cond}} - \omega\,\mathbf{o}_{\text{uncond}},
    % \vspace{-2pt}
\end{equation}

with $\omega$ the guidance scale. \ours is conceptually orthogonal to CFG. Whereas CFG relies on contrast between a conditional and a null-text (or template) prompt, \ours leverages embeddings from the visual understanding encoder as dense prompts for reconstruction-based alignment. The two techniques are fully compatible and can be applied together. 

\section{Prompt Templates for \ours}
\label{sec:templates}

\begin{figure}[t]
    \centering
    \includegraphics[width=\linewidth]{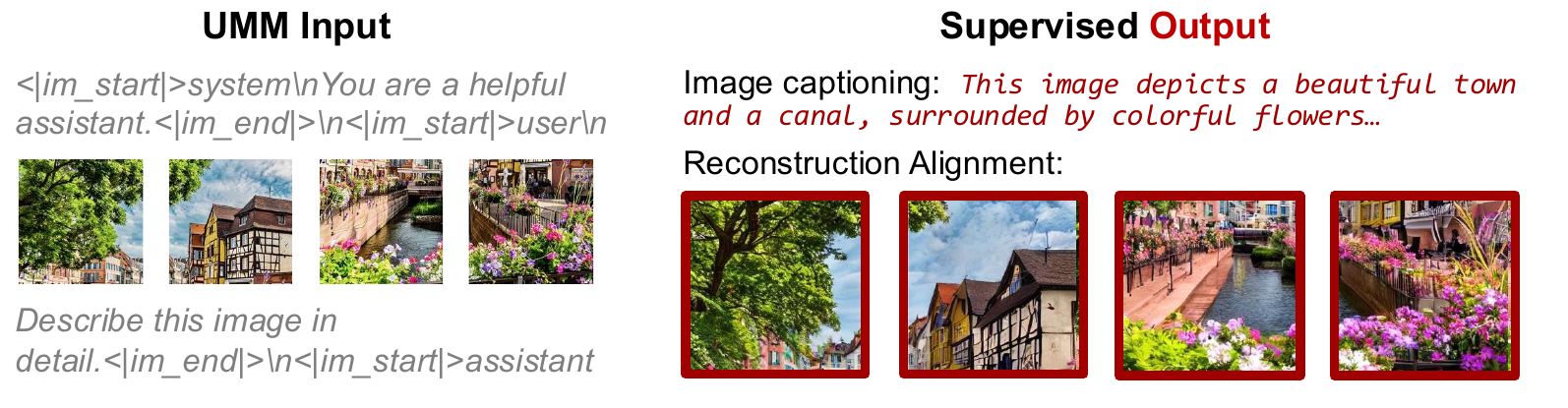}
    \caption{{\textbf{Visual embedding integration in \ours.} The left side shows the UMM input format, where visual understanding embeddings (represented as image patches) are concatenated with text prompts following standard image captioning format. The right side contrasts two supervision paradigms: image captioning uses sparse text descriptions as supervision, while \ours uses the original images themselves as dense reconstruction targets.}}
    \label{fig:how_integrated}
\end{figure}

\subsection{{Visual Embedding Integration}}

{The integration of visual understanding embeddings in \ours follows the standard multimodal input paradigm used in modern MLLMs for image captioning tasks. As illustrated in Figure~\ref{fig:how_integrated}, the input sequence consists of:}

{\begin{enumerate}[leftmargin=12pt, noitemsep, topsep=0pt]
    \item Standard instruction tuning template (e.g., \texttt{<|im\_start|>system\textbackslash nYou are a helpful assistant.<|im\_end|>}).
    \item Tokenized image representations from the visual understanding encoder (e.g., CLIP, SigLIP).
    \item A template triggering image description or reconstruction.
    \item Trigger (e.g., \texttt{<|im\_start|>assistant}).
\end{enumerate}}

{During \ours training, the model is conditioned on the concatenation of the text template and visual embeddings, and trained to reconstruct the original image. This process is identical to how MLLMs process images for captioning tasks, except the supervision signal comes from image reconstruction loss rather than text generation loss. }

\begin{figure}[t]
    \centering
    \includegraphics[width=0.95\textwidth,page=2]{figures/obs1.pdf} 
    \caption{{\textbf{Deep Feature Alignment.} PCA~\citep{abdi2010principal} visualization of text embeddings (red mark, the word ``black'') and image embeddings (a photo of a black dog) across model layers. In deep transformer layers, text and visual features share the same semantic manifold. This indicates that visual embeddings effectively function as dense, language-aligned text prompts.}}
    \label{fig:obs1_pca}
    \vspace{-3pt}
\end{figure}

\subsection{Prompt Template Diversification}

For template diversification in \ours post-training, we utilize GPT-o3 to generate 360 distinct prompt templates based on the seed template ``Describe the image in detail.'' This diversification prevents the model from overfitting to a specific prompt format and ensures robust performance across various instruction phrasings. Below are representative examples from our template collection:

\begin{itemize}
\item ``Provide a detailed description of the image.''
\item ``What can you observe in this image? Please describe it comprehensively.''
\item ``Analyze and describe the visual elements present in this image.''
\item ``Give a thorough description of what you see in the image.''
\item ``Examine the image carefully and provide a detailed account.''
\item ``Describe the contents of this image in detail.''
\item ``What does this image show? Please provide a comprehensive description.''
\item ``Offer a detailed visual analysis of the image.''
\item ``Describe all the visual elements you can identify in this image.''
\item ``Provide an in-depth description of what is depicted in the image.''
\end{itemize}

The complete collection of 360 templates varies in linguistic structure, complexity, and prompting style while maintaining the core objective of triggering detailed image descriptions.

\section{{Mechanistic Analysis of \ours}}
\label{sec:appendix_mechanism}

{In this section, we provide a deeper analysis of the mechanism behind semantic reconstruction alignment (\ours). We investigate how the model processes visual information as dense prompts and the correlation between reconstruction and generation capabilities.}

\label{subsec:isomorphism}

{\textbf{Visual-Textual Feature Isomorphism.} A fundamental premise of \ours is that visual embeddings from the understanding encoder can serve as dense supervision for the generator. As shown in Figure~\ref{fig:obs1_pca}, in the deeper layers of the transformer (Layers $3\sim 20$), text features (e.g., the word ``black'') and image features (e.g., a photo of a black dog) aligns. This suggests that visual and textual information is effectively isomorphic in the semantic space, allowing us to use images as dense prompts to supervise the generation backbone.}

\begin{figure}[t]
    \vspace{-3pt}
    \centering
    % Left: Semantic Projection (Dog / Banana)
    \begin{subfigure}[t]{0.53\textwidth}
        \centering
        \includegraphics[width=\textwidth]{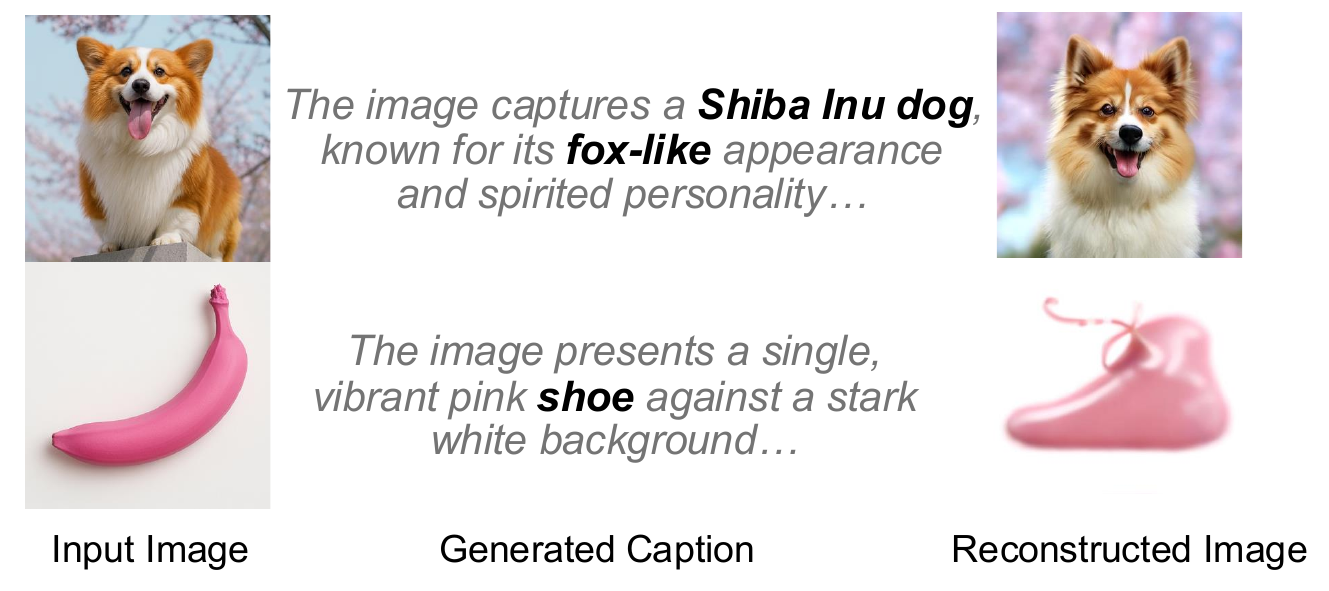} 
        \caption{{\textbf{Reconstruction reflects internal semantic projection.} 
        \textbf{Top:} The UMM extracts semantic features (e.g., ``fox-like'') from the input Corgi and resamples a visually consistent image.
        \textbf{Bottom:} For an anomalous concept (pink banana), the UMM maps the visual input to the semantic concept of a ``pink shoe'' and then faithfully produces a shoe. This confirms that generation is driven by the model's internal semantic interpretation.}}
        \label{fig:obs2_semantic_mirror}
    \end{subfigure}
    \hfill
    % Right: Task Correlation (Truck / Microwave)
    \begin{subfigure}[t]{0.45\textwidth}
        \centering
        \includegraphics[width=\textwidth]{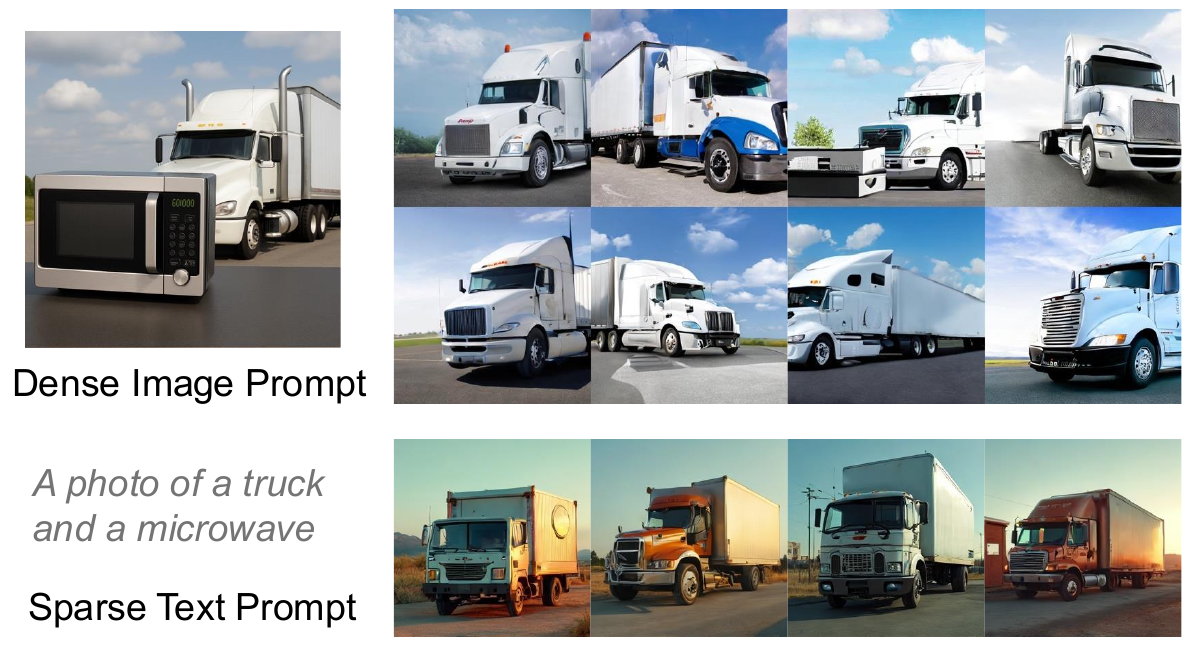} 
        \caption{{\textbf{UMM's Reconstruction correlates with generation capability.} \textbf{Top:} Provided with the dense image prompt, the UMM consistently fails to reconstruct the microwave across multiple seeds.
        \textbf{Bottom:} Consistent with the reconstruction failure, the text prompt also fails to generate the microwave. This suggests a strong correlation between the two tasks.}}
        \label{fig:obs3_correlation}
    \end{subfigure}
    \vspace{-3pt}

\end{figure}

{\textbf{Internal Semantic Projection.} As shown in Figure~\ref{fig:obs2_semantic_mirror}, We observe that the UMM treats input images as semantic prompts rather than raw pixels. For standard inputs (e.g., a dog), it performs semantic resampling. Notably, for rare inputs like a \textit{pink banana}, the UMM projects the visual features onto its nearest learned semantic concept, in this case, a ``pink shoe''. Crucially, it faithfully produces a shoe based on this internal mapping. This confirms that \textbf{images are interpreted as text-like semantic concepts}, and the UMM acts upon this interpretation. }

\label{subsec:correlation}

{\textbf{Correlation Between Reconstruction and Generation.} Furthermore, we find a strong empirical correlation between reconstruction and generation. As shown in Figure~\ref{fig:obs3_correlation}, we test a challenging composition (``microwave and truck''). (I) When the UMM is conditioned on the \textbf{dense image prompt}, it persistently fails to generate the microwave across multiple random seeds. (II) Correspondingly, when conditioned on the \textbf{sparse text prompt} (T2I), it also fails to generate.}

{This suggests that if the model cannot resolve a concept given dense visual cues, it is unlikely to succeed with sparse textual cues. Therefore, enhancing the model's ability to reconstruct complex details serves as a \emph{direct pathway} to improving its generation capabilities. \ours works by refining this mapping process.}

\section{Experimental Setup}

\label{sec:setup}

\begin{table}[t]
  \centering
  \small
  \caption{Fine-tuning hyperparameter setup.}    
  \setlength{\tabcolsep}{1pt}
  \label{tab:finetune-hp}
  \begin{tabular}{lccccccc}
    \toprule
    & \multicolumn{2}{c}{\textbf{Show-o}} & \multicolumn{2}{c}{\textbf{Harmon}} & \multicolumn{2}{c}{\textbf{OpenUni}} & \textbf{BAGEL} \\
    \cmidrule(lr){2-8}
    & 256$^{2}$ & 512$^{2}$ & 0.5B & 1.5B & 1.6B & 3.6B & 14B \\
    \midrule
    \multicolumn{8}{l}{\textbf{Optimization}} \\
    Optimizer              & AdamW & AdamW &  AdamW & AdamW & AdamW & AdamW & AdamW \\
    Learning rate          & 5e-7 & 5e-7 & 1e-5 & 1e-5 & 1e-5 & 1e-5 & 4e-5 \\
    $\beta$                & $(0.9, 0.999)$ & $(0.9, 0.999)$ & $(0.9, 0.95)$ & $(0.9, 0.95)$ & $(0.9, 0.95)$ & $(0.9, 0.95)$ & $(0.9, 0.95)$ \\
    weight decay           & 0.01 & 0.01 & 0.02 & 0.02 & 0.05 & 0.05 & 0 \\
    warmup steps           & 1000 & 1000 & 500 & 500 & 50 & 50 & 1000 \\
    Training steps         & 5K & 5K  & 3K  & 5K  & 5K & 5K  & 1K \\
    Grad.\ accumulation    & 5 & 5  & 2 & 2 & 16 & 16 & 1 \\    
    Per-GPU batch size     & 4 & 2 & 96 & 48 & 42 & 20 & 8 (6 GPUs) \\
    {Training time (hours)}  & 4 & 9 & 7 & 12.5 & 2 & 3 & 4.5 \\
    \midrule
    \multicolumn{8}{l}{\textbf{Trainable modules}} \\
    Frozen parts           & \multicolumn{2}{c}{CLIP} & \multicolumn{2}{c}{---} & \multicolumn{2}{c}{MLLM} & Und. Expert \\
    \midrule
    \multicolumn{8}{l}{\textbf{Loss weights}} \\
    $\lambda_{\text{i2t}}$ & \multicolumn{2}{c}{1.0} & \multicolumn{2}{c}{1.0}& \multicolumn{2}{c}{0.0} & 0.0 \\
    $\lambda_{\text{\ours}}$ & \multicolumn{2}{c}{1.0} & \multicolumn{2}{c}{1.0} & \multicolumn{2}{c}{1.0} & 1.0 \\
    \bottomrule
    \vspace{-25pt}
  \end{tabular}
\end{table}

\textbf{Implementation Details.} We post-train all models on a single NVIDIA A100 (80~GB) GPU. Table~\ref{tab:finetune-hp} presents the detailed hyper-parameter configurations for all models, including Show-o ($256^2$/$512^2$), Harmon (0.5B/1.5B), OpenUni (1.6B/3.6B), and BAGEL (14B). The table includes optimization settings, trainable modules, and loss weights used during \ours post-training. Hyper-parameters not listed in Table~\ref{tab:finetune-hp} follow the original open-source baselines. All $\lambda_{\text{t2i}}$ is set to 0.

For experiments that report averaged metrics (rows marked with $^*$ in Table~\ref{tab:main_table}), we run 12 repetitions using random seeds $\{0,1,\ldots,11\}$; other experiments default to seed $0$ unless otherwise stated. At inference time, we execute Show-o, Harmon, and OpenUni on an NVIDIA RTX 4090 (24~GB), while BAGEL inference remains on the A100.

\textbf{Evaluation Details.} Following previous work, we evaluate text-to-image generation capabilities using GenEval~\citep{geneval}, DPGBench~\citep{dpgbench} and WISE~\citep{wisebench}. Our baselines include: (I) \emph{Generation-only models}: SD3-Medium~\citep{sd3}, SDXL~\citep{sdxl}, SANA-1.5~\citep{xie2025sana}, Emu3-Gen~\citep{wang2024emu3}, FLUX-dev~\citep{flux}, Playground-v3~\citep{liu2024playgroundv3improvingtexttoimage}, and DALL-E 3~\citep{dalle3}. (II) \emph{Unified multimodal models}: Show-o~\citep{showo}, Harmon~\citep{wu2025harmon}, Janus-pro~\citep{chen2025janus}, OmniGen2~\citep{wu2025omnigen2explorationadvancedmultimodal}, BLIP3-o~\citep{chen2025blip3}, BAGEL~\citep{bagel}, Show-o2~\citep{xie2025show}, UniWorld-V1~\citep{lin2025uniworld}, Ovis-U1~\citep{wang2025ovisu1technicalreport}, and GPT-4o-Image~\citep{openai2024introducing4o}. For enhanced statistical reliability, we reproduced some models using 12 random seeds (tripling the standard 4 seeds).

For image editing evaluation, we employ ImgEdit~\citep{ye2025imgedit} and GEdit-Bench-EN~\citep{liu2025step1x} benchmarks. We exclude GPT-4o-Image and its distillation-trained models due to poor identity preservation. We select BAGEL~\citep{bagel}, and FLUX-Kontext~\citep{labs2025flux} as our primary baselines. Notably, we compare with the concurrent work BAGEL-NHR~\citep{kuprashevich2025nohumansrequiredautonomoushighqualityimage}, which employs supervised fine-tuning (SFT) using 300,000 high-quality image editing data. Due to changes in the GPT API version and benchmark maintenance issues, the leaderboard scores exhibit significant variance and are difficult to reproduce. Therefore, we follow BAGEL~\citep{bagel} and report our own local evaluations for consistency. Following the original benchmark, we use {gpt-4.1-2025-04-14} for ImgEdit and GEdit-Bench-EN evaluation. 
% \begin{wrapfigure}{r}{0.6\textwidth}
%     \vspace{-10pt}
%     \centering
%     \includegraphics[width=0.6\textwidth]{figures/why_flux.pdf}
%     \vspace{-18pt}
%     \caption{\textbf{Distribution mismatch affects BAGEL image quality}. Zoom in for detailed view.}
%     \label{fig:why_flux}
%     \vspace{-10pt}
% \end{wrapfigure}

\textbf{Training Data.} Due to limited availability of UMM training data, we use the high-quality open-source MidjourneyV6 dataset~\citep{midjourneyv6} {(MIT License)} for Show-o, Harmon, and OpenUni. For BAGEL, initial experiments using MidjourneyV6 data led to significantly higher initial losses and degraded image quality, which we attribute to distribution mismatch because MidjourneyV6 images are outside BAGEL's original training distribution. Therefore, we use 10,000 $1024\times 1024$ FLUX-generated~\citep{flux} images from Text-to-Image-2M~\citep{texttoimage2M} {(MIT License)}, as BAGEL was originally trained on FLUX-generated images~\citep{bagel}, ensuring better distribution alignment and training stability.

{As we discussed in Sec.~\ref{method:i2t}, to preserve visual understanding capabilities of UMMs with shared parameters for both understanding and generation (Show-o and Harmon), we also incorporate LLaVA Mix-665K~\citep{llava} (CC BY 4.0); for BAGEL and OpenUni, we set $\lambda_{\text{i2t}}=0$. }

GPT-4o-Image distillation data (e.g., BLIP3o-60k) can boost GenEval and DPGBench scores by 5$\sim$7\% on average~\citep{chen2025blip3}. However, these datasets contain GenEval prompt templates, creating evaluation bias which we call \textbf{GenEval Template Leakage}. In our primary experiments, we deliberately avoid BLIP3o-60k to ensure fair comparison. We provide a detailed analysis of this leakage issue in Sec~\ref{sec:geneval_leakage}.

\section{{More Experimental Results and Analysis}}
\label{sec:more_results_analysis}

We present additional qualitative text-to-image generation and image editing results in Figure~\ref{fig:demo_t2i} and Figure~\ref{fig:demo_edit}. The detailed captions of T2I results are listed in Appendix~\ref{sec:generated_captions}.

\subsection{{GenEval Template Leakage}}
\label{sec:geneval_leakage}

{GenEval~\citep{geneval} is a text-to-image evaluation benchmark which includes six sub-tasks: \emph{Single Object}, \emph{Two Objects}, \emph{Color}, \emph{Color Attributes}, \emph{Counting}, and \emph{Position}. The templates for these six sub-tasks are as follows:}

{\begin{itemize}[leftmargin=16pt]
    \item Single Object: \texttt{a photo of a \{object\}.}
    \item Two Objects: \texttt{a photo of a \{object1\} and a \{object2\}.}
    \item Color: \texttt{a photo of a \{color\} \{object\}.}
    \item Color Attri.: \texttt{a photo of a \{color\} \{object1\} and a \{color\} \{object2\}.}
    \item Counting: \texttt{a photo of \{count\} \{object\}.}
    \item Position: \texttt{a photo of a \{object\} \{spatial relationship\} \{object\}.}
\end{itemize}}

\begin{figure}[t]
    \centering
    \includegraphics[width=\textwidth]{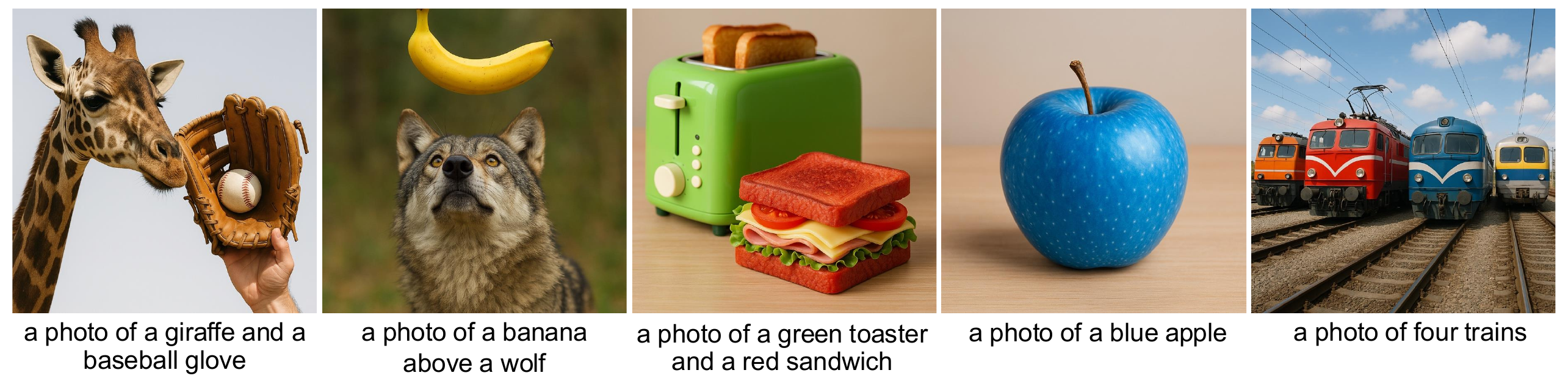}
    % \vspace{-3mm}
    \caption{{\textbf{GenEval template leakage in BLIP3o-60k.} Examples from BLIP3o-60k that mirror GenEval templates, creating evaluation bias for models trained on this dataset.}}
    \label{fig:geneval_leakage}
    % \vspace{-3mm}
\end{figure}

{However, we find that the BLIP3o-60k dataset~\citep{chen2025blip3}, distilled from GPT-4o-Image, contains 7,000 text–image pairs that nearly duplicate these templates. As shown in Figure~\ref{fig:geneval_leakage}, many training examples align closely with GenEval prompts, raising a serious risk of evaluation contamination. Models trained on BLIP3o-60k may gain unfair advantages by having effectively seen test-like data during training.}

{To ensure fair comparison, we evaluate both the full BLIP3o-60k dataset and a cleaned version with these 7,000 pairs removed. With the full set, SFT attains an inflated GenEval score of 84.95, but after removal drops to 80.88 (-4.07). In contrast, DPGBench remains unchanged (+0.05), confirming that these pairs only boost GenEval-specific performance rather than general generation capability. Thus, BLIP3o-60k reflects benchmark-targeted instruction tuning rather than a universal solution.}

{\textbf{\ours achieves comparable performance and superior robustness.} Using the full BLIP3o-60k dataset, \ours reaches 85.21 on GenEval, matching the benefit of SFT while requiring only unlabeled data and self-supervision. Crucially, on DPGBench \ours surpasses SFT (86.50 vs. 85.19), despite BLIP3o-60k containing dense-prompt data favorable to SFT. When 7,000 data are removed, \ours declines by only 0.45 and retains its DPGBench score. This robustness underscores that \ours provides a generalizable alignment signal, avoiding benchmark leakage and overfitting.}

\subsection{{\ours on Show-o with Image Tokenization}}

\label{sec:showo_tokenizer}

\begin{figure}[t]
    \centering
    \begin{minipage}{0.28\textwidth}
        \centering
        \includegraphics[width=\linewidth]{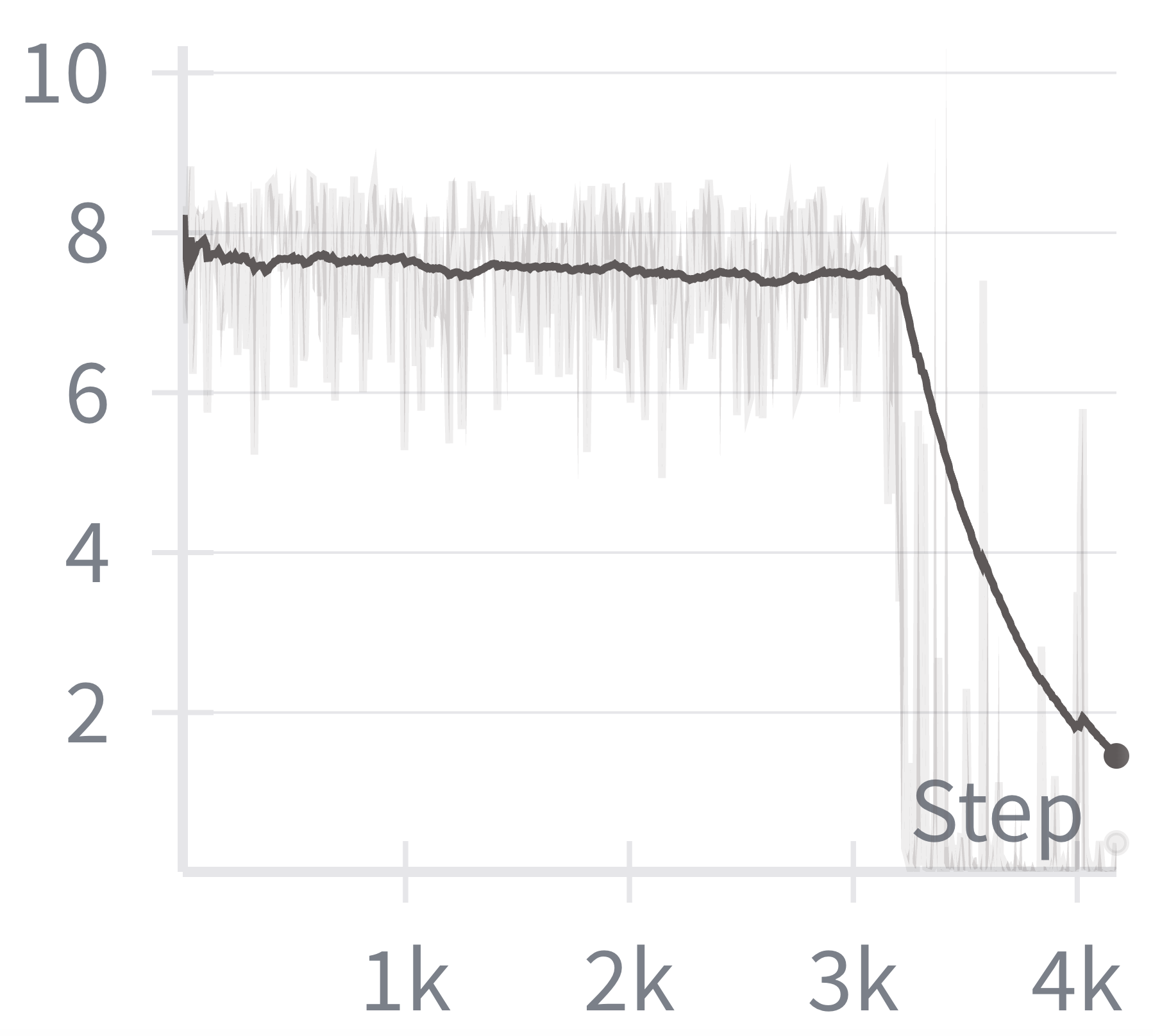}
    \end{minipage}\hfill
    \begin{minipage}{0.68\textwidth}
        \centering
        \small
        \setlength{\tabcolsep}{10pt}
        \renewcommand\arraystretch{1.00}
        \begin{tabular}{lccc}
            % \toprule
            Setting & GenEval $\uparrow$ & DPG $\uparrow$ & WISE $\uparrow$ \\
            \midrule[0.8pt]
            Baseline (no \ours) & 68.70 & 83.63 & 0.40 \\
            \midrule
            \textcolor{gray}{512$\times$512} & \textcolor{gray}{66.59} & \textcolor{gray}{83.55} & \textcolor{gray}{0.30} \\
            % \midrule
            \cellcolor[HTML]{F9FFF9}256$\times$256 & \cellcolor[HTML]{F9FFF9}70.61 & \cellcolor[HTML]{F9FFF9}84.77 & \cellcolor[HTML]{F9FFF9}\textbf{0.40} \\
            128$\times$128 & 67.47 & 83.22 & 0.39 \\
            Blur & \textbf{72.20} & \textbf{85.12} & 0.38 \\
            % \bottomrule
        \end{tabular}
    \end{minipage}
    \vspace{4pt}
    \caption{{\textbf{Token-copying collapse and mitigation for Show-o's VQGAN-understanding variant.} \emph{Left:} Cross-entropy loss plunges after 3,000 steps when training at $512\times512$. \emph{Right:} Quantitative results of Show-o under different reconstruction inputs.}}
    \label{fig:tokenizer_overfit}
\vspace{-10pt}
\end{figure}

{\textbf{Naive 512 $\times$ 512 reconstruction collapses.} When we do \ours on the VQGAN variant of Show-o at $512\times512$ resolution, which uses a VQGAN~\citep{vqgan} image tokenizer as the visual understanding encoder, the cross-entropy loss of \ours rapidly drops to nearly zero and the model's performance degrades, as shown in Figure~\ref{fig:tokenizer_overfit}. We observe that the model simply copies tokens from the input sequence instead of learning a meaningful conditioning. To mitigate this token-copying behavior, we explore three strategies:}

{\begin{itemize}[leftmargin=16pt]
    \item Resizing input images to $256\times256$, which aligns with the configuration in Sec.~\ref{method:i2t} and feeds the visual encoder its minimum supported resolution.
    \item Resizing input images to $128\times128$.
    \item Blurring input images by downsampling it by $8\times$ and then upsampling back.
\end{itemize}}

{The blurred reconstruction method yields the highest GenEval and DPGBench scores but slightly deteriorates WISE reasoning accuracy. Resizing inputs to $256\times256$ offers the best balance across all metrics. In contrast, $128\times128$ inputs fall outside Show-o's pretraining distribution and significantly reduce all of these scores.}

{\textbf{Why the gain of VQGAN variant is smaller than that of CLIP variant?} Notably, the tokenizer variant starts from higher performance (68.7/83.63 \versus 66.2/82.21), yet after \ours the 256$\times$256 reconstruction reaches only 70.61/84.77 compared with 72.3/84.94 for the Show-o CLIP variant. This gap is consistent with the weaker semantic representations provided by the VQGAN tokenizer and the lower visual understanding capacity of the Show-o tokenizer variant, leaving \ours with less high-level signal to leverage.}

\subsection{Comprehensive Quantitative Results}

\label{sec:quantitative_results}

We provide comprehensive quantitative analysis of \ours across all evaluated unified multimodal models in this section. Table~\ref{tab:main_table_full} shows the full comparison between \ours post-trained Harmon-1.5B and previous models. Table~\ref{tab:model_improvement_appendix} shows \ours's consistent performance improvements across all evaluated architectures, with particularly notable gains in positional understanding and color attribution tasks. The detailed WISE benchmark results in Table~\ref{tab:wise_results_appendix} reveal that \ours primarily enhances semantic alignment capabilities while showing modest improvements in reasoning-intensive tasks.

\begin{table*}[t!]
    \centering
    \caption{\textbf{Results on GenEval and DPGBench}. Scores marked with (*) are our reproduced results using 12 random seeds. 
    We use Harmon-1.5B as our base model and post-train it with \ours. The \textcolor{gray!80}{gray-colored} rows denote private models, and their results are cited from~\citep{yan2025gpt,geng2025xomnireinforcementlearningmakes}. Arrows ($\uparrow$) denote that higher is better. 
    }
    \vspace{-2.5mm}
    \label{tab:main_table_full}
    \small
    \setlength{\tabcolsep}{2.0pt}
    \renewcommand\arraystretch{0.9}
    % --------------------------------------------------------
    %  Column layout: | Method | Params | 7×GenEval | DPGBench |
    % --------------------------------------------------------
    \begin{tabular}{lcccccccccc}
        % \toprule
        % \bottomrule[1pt]\rowcolor[HTML]{FAFAFA}
         & & \multicolumn{7}{c}{GenEval $\uparrow$} & \multicolumn{1}{c}{\ \ \ \  DPG$\uparrow$ \ \ \ } \\
        \cmidrule(lr){3-9} \cmidrule(lr){10-10}
        \multicolumn{1}{c}{Model} & \multicolumn{1}{c}{Params} & \multicolumn{1}{c}{Single Obj.} & Two Obj. & Counting & Colors & Position & Color Attri. & Overall & Score \\
        \toprule[0.8pt]

        % ============================ Gen. Only ============================
        % \rowcolor[HTML]{F1F9FF}
        \multicolumn{10}{l}{\textit{Generation Only Models}} \\
        \toprule[0.4pt]
        SD3-Medium & 2B & 0.99 & 0.94 & 0.72 & 0.89 & 0.33 & 0.60 & 0.74 & 84.08 \\
        SDXL & 2.6B & 0.98 & 0.74 & 0.39 & 0.85 & 0.15 & 0.23 & 0.55 & 74.65 \\
        SANA-1.5 & 4.8B & 0.99 & 0.93 & 0.86 & 0.84 & 0.59 & 0.65 & 0.81 & 84.70 \\
        Emu3-Gen & 8B & 0.98 & 0.71 & 0.34 & 0.81 & 0.17 & 0.21 & 0.54 & 80.60 \\
        FLUX-dev & 12B & 0.99 & 0.85 & 0.74 & 0.79 & 0.21 & 0.48 & 0.68 & 84.00 \\
        Playground-v3 & 24B & 0.99 & 0.95 & 0.72 & 0.82 & 0.50 & 0.54 & 0.76 & 87.06 \\
        \textcolor{gray}{DALL-E~3} & \textcolor{gray}{--} & \textcolor{gray}{0.96} & \textcolor{gray}{0.87} & \textcolor{gray}{0.47} & \textcolor{gray}{0.83} & \textcolor{gray}{0.43} & \textcolor{gray}{0.45} & \textcolor{gray}{0.67} & \textcolor{gray}{83.50} \\
        \toprule[0.8pt]
        % \rowcolor[HTML]{F9FFF9}
        \multicolumn{10}{l}{\textit{Unified Multimodal Models}} \\
        \toprule[0.4pt]        \color{gray}
        \textcolor{gray}{GPT-4o-Image} & \textcolor{gray}{-} & \textcolor{gray}{0.99} & \textcolor{gray}{0.92} & \textcolor{gray}{\textbf{0.85}} & \textcolor{gray}{0.89} & \textcolor{gray}{0.74} & \textcolor{gray}{0.71} & \textcolor{gray}{0.84} & \textcolor{gray}{86.23} \\
        \color{black} Show-o* & 1.3B & 0.98 & 0.84 & 0.67 & 0.82 & 0.30 & 0.52 & 0.69 & 83.63 \\
        Harmon* & 1.5B & 0.99 & 0.87 & 0.69 & 0.86 & 0.45 & 0.51 & 0.73 & 80.93 \\
        Show-o2 & 7B & 1.00 & 0.87 & 0.58 & 0.92 & 0.52 & 0.62 & 0.76 & 86.14 \\
        Janus-Pro & 7B & 0.99 & 0.89 & 0.59 & 0.90 & \textbf{0.79} & 0.66 & 0.80 & 84.33 \\
        BAGEL* & 14B & 0.99 & 0.93 & 0.80 & 0.86 & 0.51 & 0.63 & 0.79 & 84.03 \\
        \rowcolor[HTML]{F9FFF9}
        \bf \ours & 1.5B & 1.00 & \textbf{0.98} & 0.71 & \textbf{0.93} & 0.76 & \textbf{0.77} & \textbf{0.86} & \textbf{87.21} \\
        \toprule[0.8pt]
        \multicolumn{10}{l}{\textit{Unified Multimodal Models Trained with GPT-4o Data}} \\
        \toprule[0.4pt]
        Ovis-U1 & 3.6B & \text{0.98} & \textbf{\text{0.98}} & \textbf{\text{0.90}} & \text{0.92} & \text{0.79} & \text{0.75} & \text{0.89} & \text{83.72} & \\
        OmniGen2 & 7B & \text{1.00} & \text{0.95} & \text{0.64} & \text{0.88} & \text{0.55} & \text{0.76} & \text{0.80} & \text{83.57} \\
        BLIP3-o* & 8B & \text{1.00} & \text{0.92} & \text{0.63} & \text{0.91} & \text{0.86} & \text{0.67} & \text{0.83} & \text{80.73} \\
        UniWorld-V1 & 20B & \text{0.99} & \text{0.93} & \text{0.79} & \text{0.89} & \text{0.49} & \text{0.70} & \text{0.80} & -- & \\
        % \hline
        \rowcolor[HTML]{F9FFF9}
        \bf \ours & 1.5B & \text{1.00} & \text{0.97} & \text{0.76} & \textbf{\text{0.94}} & \textbf{\text{0.91}} & \textbf{\text{0.83}} & \textbf{\text{0.90}} & \textbf{\text{88.15}} \\
        
        % \bottomrule
    \end{tabular}
    \vspace{-2mm}
\end{table*}

\begin{table*}[t!]
    \centering
    \caption{\textbf{\ours brings consistent performance gains to various UMM frameworks.} 
    We show the performance gains after applying \ours to different unified multimodal models. All models are evaluated on GenEval, DPGBench, and WISE benchmarks.}
    \vspace{-2mm}
    \footnotesize
    \setlength{\tabcolsep}{0.70pt}
    \renewcommand\arraystretch{1.0}

    \label{tab:model_improvement_appendix}
    
    \begin{tabular}{lcllllllll}
        % \toprule
        \multirow{2}{*}{Model} & \multirow{2}{*}{\ours} & \multicolumn{7}{c}{GenEval} & \multirow{2}{*}{DPG}\\
        \cmidrule(lr){3-9}
        & & \multicolumn{1}{c}{Single} & \multicolumn{1}{c}{Two} & \multicolumn{1}{c}{Count} & \multicolumn{1}{c}{Color} & \multicolumn{1}{c}{Position} & \multicolumn{1}{c}{Attri.} & \multicolumn{1}{c}{Overall} & \\

        \toprule[0.7pt]
        \multirow{2}{*}{Show-o-256} & \xmark & \bf 97.4 & 63.3 & 52.1 & 82.3 & 14.2 & 30.3 & 56.6 & 70.65 \\
         & \cellcolor[HTML]{F9FFF9}\cmark & \cellcolor[HTML]{F9FFF9}\bf 97.4 \tiny (0.0) & \cellcolor[HTML]{F9FFF9}\bf 73.6 \textgr{\tiny (+10.3)} & \cellcolor[HTML]{F9FFF9}\bf 56.0 \textgr{\tiny (+3.9)} & \cellcolor[HTML]{F9FFF9}\bf 83.8 \textgr{\tiny (+1.5)} & \cellcolor[HTML]{F9FFF9}\bf 20.3 \textgr{\tiny (+6.1)} & \cellcolor[HTML]{F9FFF9}\bf 40.2 \textgr{\tiny (+9.9)} & \cellcolor[HTML]{F9FFF9}\textbf{61.9 \textgr{\tiny (+5.3)}} & \cellcolor[HTML]{F9FFF9}\textbf{75.70 \textgr{\tiny (+5.05)}} \\

        \toprule[0.7pt]
        \multirow{2}{*}{Show-o-512} & \xmark & 97.2 & 80.3 & 61.9 & 78.2 & 27.3 & 52.3 & 66.2 & 82.21 \\
        & \cellcolor[HTML]{F9FFF9}\cmark & \cellcolor[HTML]{F9FFF9}\bf 98.2 \textgr{\tiny (+1.0)} & \cellcolor[HTML]{F9FFF9}\bf 90.6 \textgr{\tiny (+10.3)} & \cellcolor[HTML]{F9FFF9}\bf 66.8 \textgr{\tiny (+4.9)} & \cellcolor[HTML]{F9FFF9}\bf 84.1 \textgr{\tiny (+5.9)} & \cellcolor[HTML]{F9FFF9}\bf 37.4 \textgr{\tiny (+10.1)} & \cellcolor[HTML]{F9FFF9}\bf 56.8 \textgr{\tiny (+4.5)} & \cellcolor[HTML]{F9FFF9}\textbf{72.3 \textgr{\tiny (+6.1)}} & \cellcolor[HTML]{F9FFF9}\textbf{84.94 \textgr{\tiny (+2.73)}} \\

        \toprule[0.7pt]
        \multirow{2}{*}{OpenUni-1.6B} & \xmark & 96.8 & 63.3 & 46.4 & 80.1 & 18.5 & 30.8 & 56.0 & 76.29 \\
        & \cellcolor[HTML]{F9FFF9}\cmark & \cellcolor[HTML]{F9FFF9}\textbf{97.1 \textgr{\tiny (+0.3)}} & \cellcolor[HTML]{F9FFF9}\textbf{84.3 \textgr{\tiny (+21.0)}} & \cellcolor[HTML]{F9FFF9}\textbf{57.4 \textgr{\tiny (+11.0)}} & \cellcolor[HTML]{F9FFF9}\textbf{83.5 \textgr{\tiny (+3.4)}} & \cellcolor[HTML]{F9FFF9}\textbf{44.3 \textgr{\tiny (+25.8)}} & \cellcolor[HTML]{F9FFF9}\textbf{56.0 \textgr{\tiny (+25.2)}} & \cellcolor[HTML]{F9FFF9}\textbf{70.4 \textgr{\tiny (+14.4)}} & \cellcolor[HTML]{F9FFF9}\textbf{80.45 \textgr{\tiny (+4.16)}} \\

        \toprule[0.7pt]
        \multirow{2}{*}{OpenUni-3.6B} & \xmark & \bf 99.1 & 71.8 & 51.9 & 83.9 & 23.3 & 41.6 & 61.9 & 79.02 \\
        & \cellcolor[HTML]{F9FFF9}\cmark & \cellcolor[HTML]{F9FFF9}\bf 99.1 \tiny (0.0) & \cellcolor[HTML]{F9FFF9}\bf 92.7 \textgr{\tiny (+20.9)} & \cellcolor[HTML]{F9FFF9}\bf 52.3 \textgr{\tiny (+0.4)} & \cellcolor[HTML]{F9FFF9}\bf 87.1 \textgr{\tiny (+3.2)} & \cellcolor[HTML]{F9FFF9}\bf 43.8 \textgr{\tiny (+20.5)} & \cellcolor[HTML]{F9FFF9}\bf 70.3 \textgr{\tiny (+28.7)} & \cellcolor[HTML]{F9FFF9}\textbf{74.1 \textgr{\tiny (+12.2)}} & \cellcolor[HTML]{F9FFF9}\textbf{82.75 \textgr{\tiny (+3.73)}} \\

        \toprule[0.7pt]
        \multirow{2}{*}{Harmon-0.5B} & \xmark & 99.7 & 80.5 & 55.8 & 86.7 & 32.2 & 49.7 & 67.6 & 80.12 \\
        & \cellcolor[HTML]{F9FFF9}\cmark & \cellcolor[HTML]{F9FFF9}\bf 99.9 \textgr{\tiny (+0.2)} & \cellcolor[HTML]{F9FFF9}\bf 92.3 \textgr{\tiny (+11.8)} & \cellcolor[HTML]{F9FFF9}\bf 59.4 \textgr{\tiny (+3.6)} & \cellcolor[HTML]{F9FFF9}\bf 91.7 \textgr{\tiny (+5.0)} & \cellcolor[HTML]{F9FFF9}\bf 58.5 \textgr{\tiny (+26.3)} & \cellcolor[HTML]{F9FFF9}\bf 70.7 \textgr{\tiny (+21.0)} & \cellcolor[HTML]{F9FFF9}\textbf{78.7 \textgr{\tiny (+11.1)}} & \cellcolor[HTML]{F9FFF9}\textbf{84.67 \textgr{\tiny (+4.55)}} \\

        \toprule[0.7pt]
        \multirow{2}{*}{Harmon-1.5B} & \xmark & 99.4 & 87.3 & 68.7 & 86.4 & 44.9 & 51.1 & 72.9 & 80.93 \\
        & \cellcolor[HTML]{F9FFF9}\cmark & \cellcolor[HTML]{F9FFF9}\textbf{99.9 \textgr{\tiny (+0.5)}} & \cellcolor[HTML]{F9FFF9}\textbf{97.7 \textgr{\tiny (+10.4)}} & \cellcolor[HTML]{F9FFF9}\textbf{71.4 \textgr{\tiny (+2.7)}} & \cellcolor[HTML]{F9FFF9}\textbf{92.6 \textgr{\tiny (+6.2)}} & \cellcolor[HTML]{F9FFF9}\textbf{75.7 \textgr{\tiny (+30.8)}} & \cellcolor[HTML]{F9FFF9}\textbf{76.6 \textgr{\tiny (+25.5)}} & \cellcolor[HTML]{F9FFF9}\textbf{85.7 \textgr{\tiny (+12.8)}} & \cellcolor[HTML]{F9FFF9}\textbf{87.21 \textgr{\tiny (+6.28)}} \\

        \toprule[0.7pt]
        \multirow{2}{*}{BAGEL} & \xmark & 99.1 & 93.0 & 79.9 & 86.0 & 51.3 & 63.4 & 78.8 & 84.03 \\
        & \cellcolor[HTML]{F9FFF9}\cmark & \cellcolor[HTML]{F9FFF9}\bf 99.3 \textgr{\tiny (+0.2)} & \cellcolor[HTML]{F9FFF9}\bf 93.9 \textgr{\tiny (+0.9)} & \cellcolor[HTML]{F9FFF9}\bf 80.3 \textgr{\tiny (+0.4)} & \cellcolor[HTML]{F9FFF9}\bf 87.6 \textgr{\tiny (+1.6)} & \cellcolor[HTML]{F9FFF9}\bf 60.8 \textgr{\tiny (+9.5)} & \cellcolor[HTML]{F9FFF9}\bf 72.6 \textgr{\tiny (+9.2)} & \cellcolor[HTML]{F9FFF9}\textbf{82.4 \textgr{\tiny (+3.6)}} & \cellcolor[HTML]{F9FFF9}\textbf{85.29 \textgr{\tiny (+1.26)}} \\
        
        % \bottomrule
    \end{tabular}
    \vspace{-3mm}
\end{table*}

\begin{table*}[t]
  \centering
  \caption{\textbf{WISE benchmark results}. Performance on reasoning-based text-to-image generation. Harmon and OpenUni exhibit zero-shot improvement but Show-o and BAGEL do not.}
  \vspace{2mm}
  \footnotesize
  \setlength{\tabcolsep}{3pt}
  \renewcommand\arraystretch{0.80}
  \label{tab:wise_results_appendix}

  \begin{tabular}{lclllllll}
    % \toprule[1pt]
    Model & \ours & Cultural & Time & Space & Biology & Physics & Chemistry & Overall \\
    \midrule
    \multirow{2}{*}{Show-o-256} & \xmark & 0.27 & 0.38 & 0.45 & 0.26 & 0.37 & 0.24 & 0.33 \\
     & \cellcolor[HTML]{F9FFF9}\cmark & \cellcolor[HTML]{F9FFF9}\textbf{0.28 \textgr{\tiny (+0.01)}} & \cellcolor[HTML]{F9FFF9}\textbf{0.40 \textgr{\tiny (+0.02)}} & \cellcolor[HTML]{F9FFF9}\textbf{0.46 \textgr{\tiny (+0.01)}} & \cellcolor[HTML]{F9FFF9}\textbf{0.27 \textgr{\tiny (+0.01)}} & \cellcolor[HTML]{F9FFF9}\textbf{0.38 \textgr{\tiny (+0.01)}} & \cellcolor[HTML]{F9FFF9}0.24 \tiny (0.00) & \cellcolor[HTML]{F9FFF9}\textbf{0.34 \textgr{\tiny (+0.01)}} \\
    \midrule
    \multirow{2}{*}{Show-o-512} & \xmark & 0.31 & 0.42 & 0.55 & 0.31 & 0.48 & \textbf{0.32} & 0.40 \\
     & \cellcolor[HTML]{F9FFF9}\cmark & \cellcolor[HTML]{F9FFF9}0.31 \tiny (0.00) & \cellcolor[HTML]{F9FFF9}0.42 \tiny (0.00) & \cellcolor[HTML]{F9FFF9}\textbf{0.56 \textgr{\tiny (+0.01)}} & \cellcolor[HTML]{F9FFF9}\textbf{0.32 \textgr{\tiny (+0.01)}} & \cellcolor[HTML]{F9FFF9}\textbf{0.50 \textgr{\tiny (+0.02)}} & \cellcolor[HTML]{F9FFF9}0.30 \textre{\tiny (-0.02)} & \cellcolor[HTML]{F9FFF9}0.40 \tiny (0.00) \\
    \midrule
    \multirow{2}{*}{OpenUni-1.6B} & \xmark & 0.40 & 0.41 & 0.54 & 0.35 & 0.52 & 0.36 & 0.43 \\
     & \cellcolor[HTML]{F9FFF9}\cmark & \cellcolor[HTML]{F9FFF9}0.40 \tiny (0.00) & \cellcolor[HTML]{F9FFF9}\textbf{0.46 \textgr{\tiny (+0.05)}} & \cellcolor[HTML]{F9FFF9}\textbf{0.56 \textgr{\tiny (+0.02)}} & \cellcolor[HTML]{F9FFF9}\textbf{0.42 \textgr{\tiny (+0.07)}} & \cellcolor[HTML]{F9FFF9}\textbf{0.54 \textgr{\tiny (+0.02)}} & \cellcolor[HTML]{F9FFF9}0.34 \textre{\tiny (-0.02)} & \cellcolor[HTML]{F9FFF9}\textbf{0.45 \textgr{\tiny (+0.02)}} \\
    \midrule
    \multirow{2}{*}{OpenUni-3.6B} & \xmark & 0.36 & 0.38 & 0.59 & 0.43 & 0.50 & 0.34 & 0.43 \\
     & \cellcolor[HTML]{F9FFF9}\cmark & \cellcolor[HTML]{F9FFF9}\textbf{0.52 \textgr{\tiny (+0.16)}} & \cellcolor[HTML]{F9FFF9}\textbf{0.51 \textgr{\tiny (+0.13)}} & \cellcolor[HTML]{F9FFF9}\textbf{0.69 \textgr{\tiny (+0.10)}} & \cellcolor[HTML]{F9FFF9}\textbf{0.50 \textgr{\tiny (+0.07)}} & \cellcolor[HTML]{F9FFF9}\textbf{0.64 \textgr{\tiny (+0.14)}} & \cellcolor[HTML]{F9FFF9}\textbf{0.40 \textgr{\tiny (+0.06)}} & \cellcolor[HTML]{F9FFF9}\textbf{0.54 \textgr{\tiny (+0.11)}} \\
    \midrule
    \multirow{2}{*}{Harmon-0.5B} & \xmark & 0.28 & 0.39 & 0.41 & 0.31 & 0.38 & 0.22 & 0.33 \\
     & \cellcolor[HTML]{F9FFF9}\cmark & \cellcolor[HTML]{F9FFF9}\textbf{0.33 \textgr{\tiny (+0.05)}} & \cellcolor[HTML]{F9FFF9}\textbf{0.43 \textgr{\tiny (+0.04)}} & \cellcolor[HTML]{F9FFF9}\textbf{0.54 \textgr{\tiny (+0.13)}} & \cellcolor[HTML]{F9FFF9}\textbf{0.38 \textgr{\tiny (+0.07)}} & \cellcolor[HTML]{F9FFF9}\textbf{0.48 \textgr{\tiny (+0.10)}} & \cellcolor[HTML]{F9FFF9}\textbf{0.24 \textgr{\tiny (+0.02)}} & \cellcolor[HTML]{F9FFF9}\textbf{0.40 \textgr{\tiny (+0.07)}} \\
    \midrule
    \multirow{2}{*}{Harmon-1.5B} & \xmark & 0.38 & 0.48 & 0.52 & 0.37 & 0.44 & 0.29 & 0.41 \\
     & \cellcolor[HTML]{F9FFF9}\cmark & \cellcolor[HTML]{F9FFF9}\textbf{0.44 \textgr{\tiny (+0.06)}} & \cellcolor[HTML]{F9FFF9}\textbf{0.58 \textgr{\tiny (+0.10)}} & \cellcolor[HTML]{F9FFF9}\textbf{0.57 \textgr{\tiny (+0.05)}} & \cellcolor[HTML]{F9FFF9}\textbf{0.48 \textgr{\tiny (+0.11)}} & \cellcolor[HTML]{F9FFF9}\textbf{0.58 \textgr{\tiny (+0.14)}} & \cellcolor[HTML]{F9FFF9}\textbf{0.32 \textgr{\tiny (+0.03)}} & \cellcolor[HTML]{F9FFF9}\textbf{0.50 \textgr{\tiny (+0.09)}} \\
    \midrule
    \multirow{2}{*}{BAGEL} & \xmark & 0.42 & \textbf{0.53} & 0.64 & 0.42 & 0.57 & 0.43 & 0.50 \\
     & \cellcolor[HTML]{F9FFF9}\cmark & \cellcolor[HTML]{F9FFF9}\textbf{0.43 \textgr{\tiny (+0.01)}} & \cellcolor[HTML]{F9FFF9}0.51 \textre{\tiny (-0.02)} & \cellcolor[HTML]{F9FFF9}\textbf{0.67 \textgr{\tiny (+0.03)}} & \cellcolor[HTML]{F9FFF9}\textbf{0.46 \textgr{\tiny (+0.04)}} & \cellcolor[HTML]{F9FFF9}\textbf{0.59 \textgr{\tiny (+0.02)}} & \cellcolor[HTML]{F9FFF9}\textbf{0.46 \textgr{\tiny (+0.03)}} & \cellcolor[HTML]{F9FFF9}\textbf{0.52 \textgr{\tiny (+0.02)}} \\
    % \bottomrule[1pt
  \end{tabular}
  \vspace{-3mm}
\end{table*}

\subsection{{Additional Compositional Evaluation on T2I-CompBench}}

{To further substantiate the robustness of our evaluation framework and address concerns regarding the complexity of the compositional tests, we conducted an additional rigorous analysis utilizing the \textbf{T2I-CompBench}~\citep{huang2023t2i}. T2I-CompBench is a specialized dataset comprising 2,670 carefully curated test samples designed to systematically evaluate the compositional understanding capabilities of text-to-image models across \textbf{10 distinct compositional dimensions}. While GenEval primarily focuses on four fundamental attributes (object presence, count, color, and 2D position), T2I-CompBench systematically assesses a comprehensive range of high-level compositional attributes and relationships. Specifically, T2I-CompBench includes critical capability dimensions absent or underspecified in simpler benchmarks, such as:}

{\begin{itemize}[leftmargin=16pt]
    \item \textbf{Shape \& Texture:} Dedicated tests for understanding diverse geometric forms (e.g., ``a pentagonal stop sign'') and material properties (e.g., ``a plastic toy'').
    \item \textbf{Numeracy:} Precise control over object quantities (e.g., ``six airplanes'').
    \item \textbf{3D Spatial:} Crucially, the benchmark distinguishes between 2D spatial and 3D spatial reasoning, with the latter focusing specifically on depth perception and occlusion relationships (e.g., ``a chair hidden by a turtle'')---a key capability for evaluating complex scene understanding.
    \item \textbf{Complex Scenes:} High-level tasks like complex scenes and complex action/spatial scenarios evaluate models' generalization capability in situations involving multiple intertwined attributes, objects, and abstract relationships (e.g., ``The red hat was on top of the brown coat rack'' simultaneously testing color binding, spatial relationships, and object recognition).
\end{itemize}}

{Following SRUM~\citep{jin2025srum}, we employ QwenVL-2.5-32B~\citep{Qwen2.5-VL} as the designated multimodal evaluator. The results presented in Table~\ref{tab:t2icompbench_results_appendix} validate the effectiveness of \ours across all tested UMM architectures, consistently delivering substantial performance gains. Crucially, the results on T2I-CompBench serve as a powerful complement to our GenEval findings. While GenEval confirms alignment in fundamental attributes (Color) and 2D positioning, T2I-CompBench demonstrates that our method's improvements generalize to a much broader spectrum of visual semantics. 
\textbf{(I) Beyond Color:} The substantial gains in \textbf{Texture} and \textbf{Shape} (e.g., $+10.38$ on shape for OpenUni) prove that \ours refines the model's sensitivity to all fine-grained visual attributes, not just color binding. 
\textbf{(II) Beyond 2D Layout:} The remarkable improvements in \textbf{3D Spatial} reasoning (e.g., $+15.75$ for OpenUni, $+10.50$ for Harmon) indicate that \ours effectively teaches the model to comprehend depth, perspective, and occlusion, far exceeding simple up/down/left/right positional alignment. \textbf{(III) In Complex Scenario}, all models showed consistent gains (e.g., Harmon-1.5B increased by $+6.46$ points). The only minor fluctuation was observed in the Non-Spatial category for Show-o-512 ($-0.25$), which does not diminish the overwhelming positive trend across high-difficulty tasks.}

\begin{table*}[t!]
    \centering
    \caption{{\textbf{T2I-CompBench Performance Analysis.} 
    Comparison of performance across sub-categories before and after applying \ours.}}
    \vspace{-2mm}
    \label{tab:t2icompbench_results_appendix}
    \footnotesize
    \setlength{\tabcolsep}{4.0pt} % Extremely tight column spacing
    \renewcommand\arraystretch{1.3}
    
    % Updated structure: Category + 4 Models * 2 Columns
     % \arraystretch{1.2} % Increased row height slightly
    
    \begin{tabular}{c|cc|cc|cc|cc}
        \multirow{2}{*}{\textbf{Category}} & \multicolumn{2}{c|}{OpenUni-3.6B} & \multicolumn{2}{c|}{Harmon-1.5B} & \multicolumn{2}{c|}{Show-o-512} & \multicolumn{2}{c}{BAGEL} \\
        \cmidrule(lr){2-3} \cmidrule(lr){4-5} \cmidrule(lr){6-7} \cmidrule(lr){8-9}
         & Base & \ours & Base & \ours & Base & \ours & Base & \ours \\
        \midrule
        Color & 87.47 & \textbf{92.50} \textgr{\tiny \textbf{(+5.03)}} & 89.55 & \textbf{93.72} \textgr{\tiny \textbf{(+4.17)}} & 88.60 & \textbf{90.88} \textgr{\tiny \textbf{(+2.28)}} & 93.05 & \textbf{93.62} \textgr{\tiny \textbf{(+0.57)}} \\
        Shape & 75.22 & \textbf{85.60} \textgr{\tiny \textbf{(+10.38)}} & 79.09 & \textbf{84.32} \textgr{\tiny \textbf{(+5.23)}} & 77.68 & \textbf{82.22} \textgr{\tiny \textbf{(+4.54)}} & 84.25 & \textbf{85.72} \textgr{\tiny \textbf{(+1.47)}} \\
        Texture & 82.87 & \textbf{89.65} \textgr{\tiny \textbf{(+6.78)}} & 86.28 & \textbf{91.37} \textgr{\tiny \textbf{(+5.09)}} & 85.23 & \textbf{88.37} \textgr{\tiny \textbf{(+3.14)}} & 88.95 & \textbf{90.58} \textgr{\tiny \textbf{(+1.63)}} \\
        Spatial & 78.29 & \textbf{86.30} \textgr{\tiny \textbf{(+8.01)}} & 84.98 & \textbf{92.37} \textgr{\tiny \textbf{(+7.39)}} & 85.75 & \textbf{87.28} \textgr{\tiny \textbf{(+1.53)}} & 88.42 & \textbf{88.57} \textgr{\tiny \textbf{(+0.15)}} \\
        Non-Spatial & 83.10 & \textbf{87.18} \textgr{\tiny \textbf{(+4.08)}} & 82.77 & \textbf{87.12} \textgr{\tiny \textbf{(+4.35)}} & 85.38 & 85.13 \textre{\tiny \textbf{(-0.25)}} & 88.27 & \textbf{88.80} \textgr{\tiny \textbf{(+0.53)}} \\
        Numeracy & 64.44 & \textbf{75.97} \textgr{\tiny \textbf{(+11.53)}} & 69.19 & \textbf{79.78} \textgr{\tiny \textbf{(+10.59)}} & 71.39 & \textbf{74.42} \textgr{\tiny \textbf{(+3.03)}} & 79.43 & \textbf{81.88} \textgr{\tiny \textbf{(+2.45)}} \\
        3D Spatial & 68.60 & \textbf{84.35} \textgr{\tiny \textbf{(+15.75)}} & 75.58 & \textbf{86.08} \textgr{\tiny \textbf{(+10.50)}} & 75.38 & \textbf{81.42} \textgr{\tiny \textbf{(+6.04)}} & 81.47 & \textbf{83.50} \textgr{\tiny \textbf{(+2.03)}} \\
        Complex & 82.82 & \textbf{87.79} \textgr{\tiny \textbf{(+4.97)}} & 82.05 & \textbf{88.51} \textgr{\tiny \textbf{(+6.46)}} & 82.73 & \textbf{85.07} \textgr{\tiny \textbf{(+2.34)}} & 87.85 & \textbf{88.74} \textgr{\tiny \textbf{(+0.89)}} \\
        \midrule
        \textbf{Overall} & 78.84 & \textbf{86.49} \textgr{\tiny \textbf{(+7.65)}} & 81.36 & \textbf{88.03} \textgr{\tiny \textbf{(+6.67)}} & 82.12 & \textbf{84.47} \textgr{\tiny \textbf{(+2.35)}} & 86.74 & \textbf{87.89} \textgr{\tiny \textbf{(+1.15)}} \\
    \end{tabular}
    \vspace{-3mm}
\end{table*}

\subsection{{GenEval Subtask Dynamics}}

\label{sec:geneval_subtasks}

\begin{figure}[t]
    \centering
    \includegraphics[width=\textwidth]{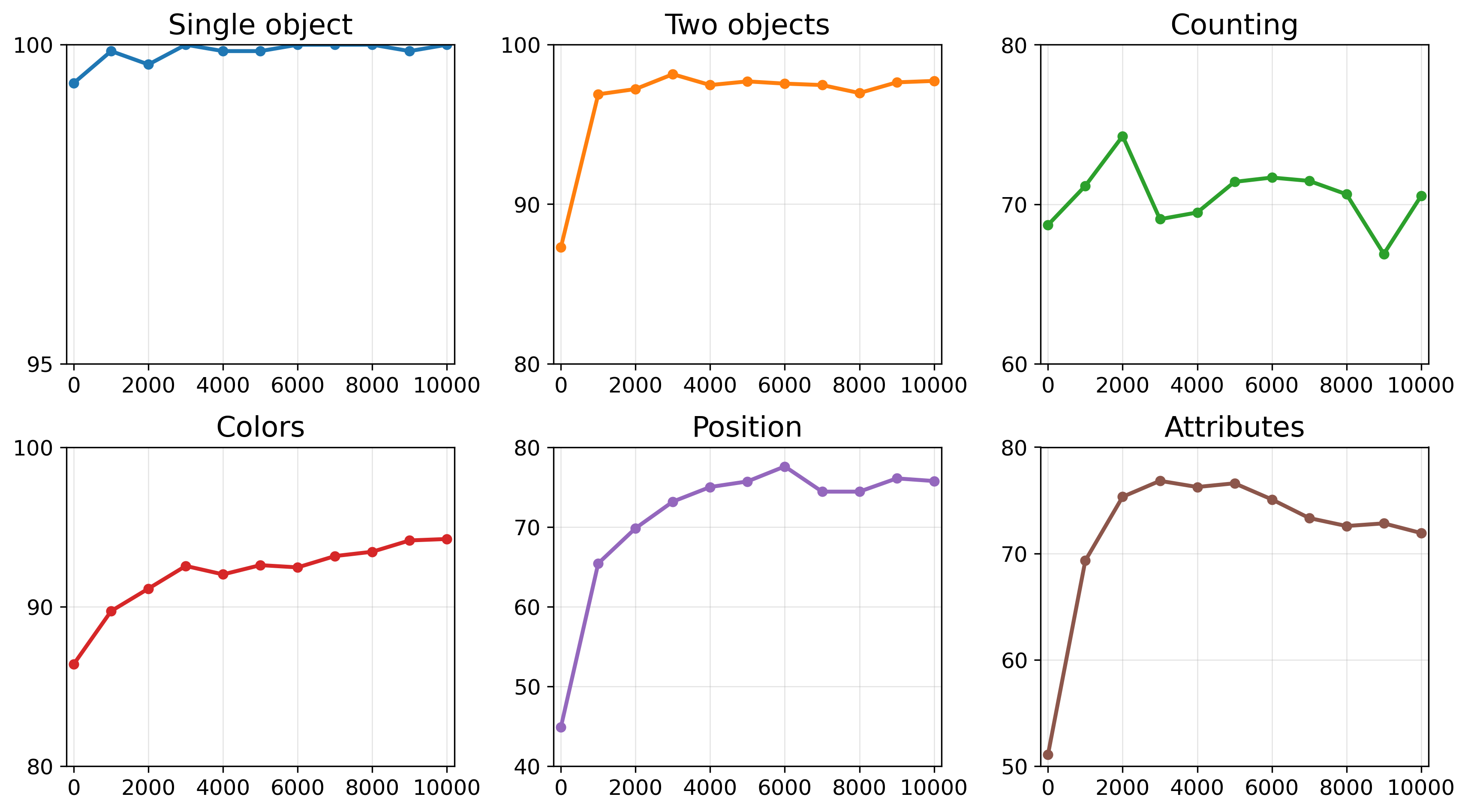}
    \vspace{-2mm}
    \caption{{\textbf{GenEval subtask performance dynamic during \ours training.} We plot the Harmon-1.5B scores for each GenEval component at different training steps.}}
    \label{fig:geneval_subtasks}
    \vspace{-3mm}
\end{figure}

{\textbf{How do individual GenEval subtasks evolve?} Figure~\ref{fig:geneval_subtasks} shows that \ours primarily boosts performance within the first $3\mathrm{k}\sim5\mathrm{k}$ steps before convergence. \emph{Single Object} and \emph{Two Objects} success rates are already saturated and quickly flatten once they touch the ceiling. \emph{Color} continues to climb through the entire run, indicating sustained improvements in attribute grounding. \emph{Position} rises sharply before $5\mathrm{k}$ steps and then stabilizes at its ceiling. \emph{Color Attributes} accuracy peaks around $5\mathrm{k}$ steps and then gently drops as we optimize for other objectives. \emph{Counting} exhibits only minor fluctuations without a clear upward trend, matching the limitation we discuss in Appendix~\ref{sec:limitations}. Understanding how to balance different subtasks for finer-grained control is an important direction for future work.}

\subsection{Further Discussion on Experiment Results}

\textbf{Why the improvements on Show-o is modest?} (I) The CLIP used in Show-o exhibits insufficient high-level semantic information~\citep{tong2024eyes}, which limits the upperbound of \ours. (II) Show-o's codebook has only 4096 tokens. fundamentally limits its representational capacity and generation performance. % (III) Show-o's original image reconstruction capabilities is minimal, producing reconstructions with significant noise artifacts. Consequently, \ours must simultaneously learn to mitigate noise generation while conditioning on CLIP features, presenting additional optimization challenges.

\textbf{Why the text-to-image generation improvements on BAGEL is modest?} (I) The data and computational resources prevent the full exploitation of the potential benefits of \ours. (II) BAGEL's pre-existing image editing training, which conditions on both SigLIP~\citep{siglip} and VAE~\citep{vae} features, has already endowed the model with good capabilities for semantic-level reconstruction. It constrains the improvement space in generation tasks, while correspondingly enabling substantial enhancements in image editing capabilities.

\subsection{{More Visualization of Reconstruction Results}}
\label{sec:reconstruction_visualization}

\begin{figure}[t]
    \centering
    \includegraphics[width=\textwidth]{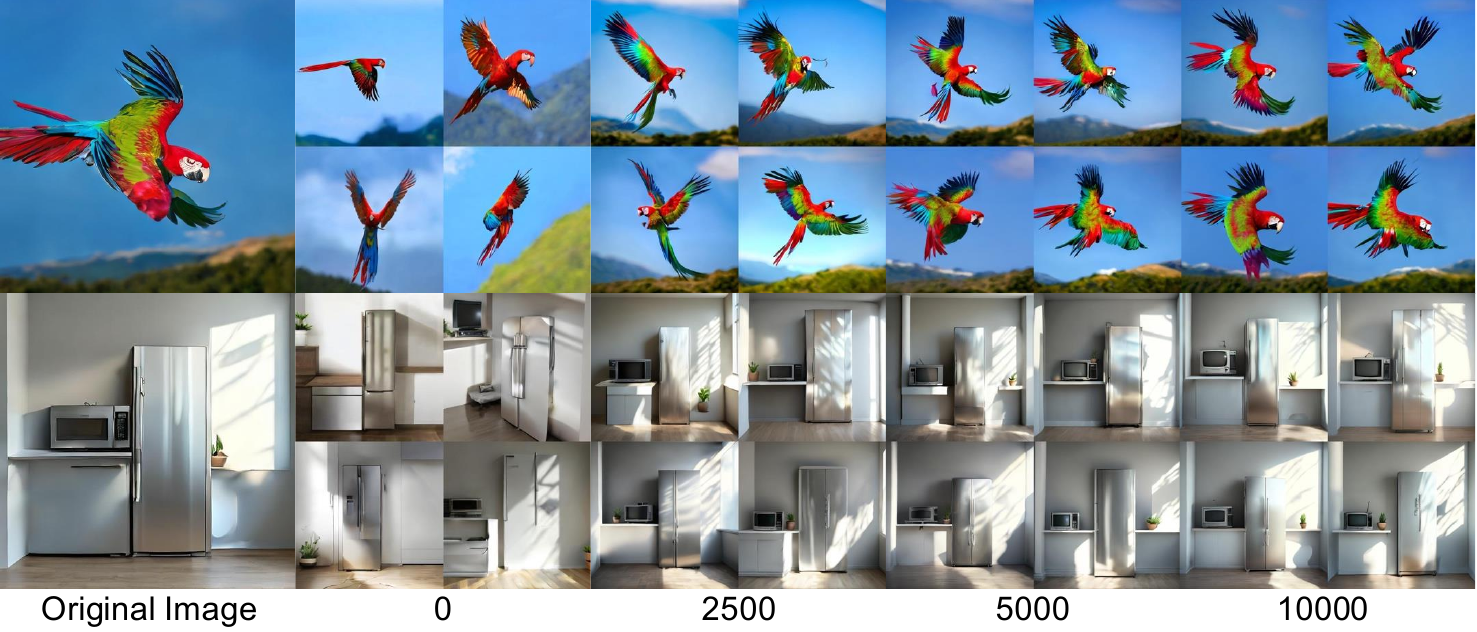}
    \caption{\textbf{Harmon reconstruction progression during \ours training.} (Left to Right) Early timesteps show blurry or incomplete reconstructions, while later stages achieve high-fidelity reconstruction with accurate semantic details, object structures, and color rendering.}
    \label{fig:recon_visualization_harmon}
\end{figure}

\begin{figure}[t]
    \centering
    \includegraphics[width=\textwidth]{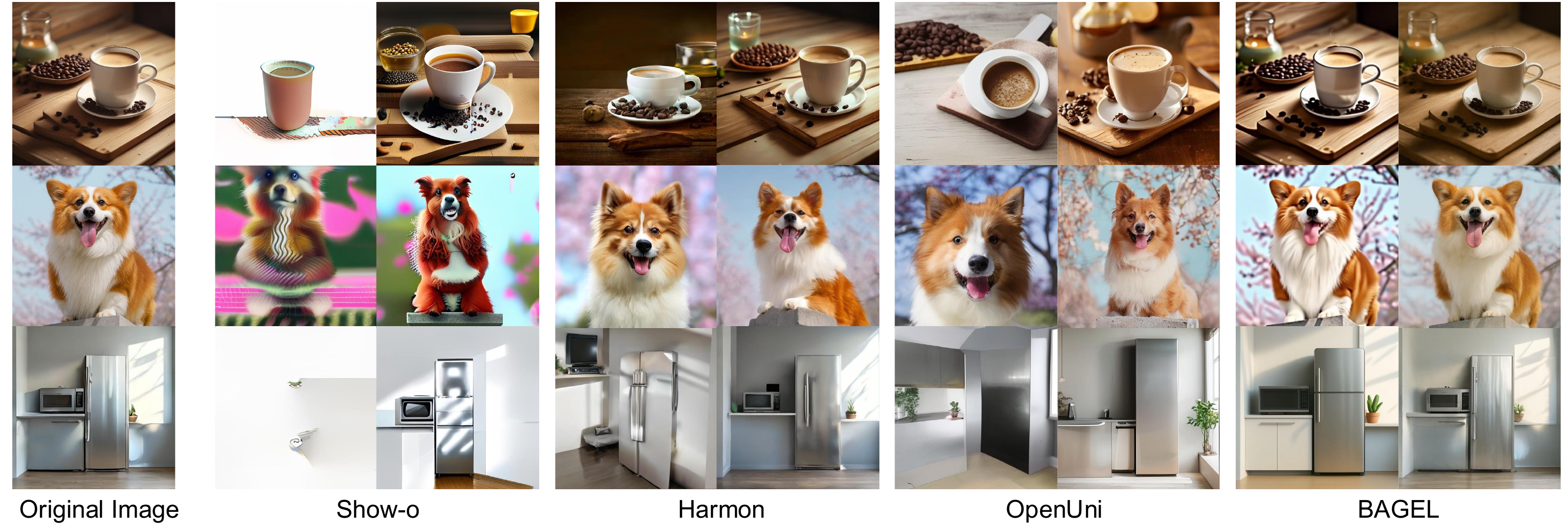}
    \caption{\textbf{Reconstruction comparison across UMM architectures.} For each model (Show-o, Harmon, OpenUni, BAGEL), we show reconstruction results before (left) and after (right) \ours post-training. \ours consistently improves reconstruction fidelity, especially in semantic details and color accuracy, which further benefits text-to-image generation.}
    \label{fig:recon_visualization_models}    
    \vspace{-4em}
\end{figure}

{To provide deeper insights into how \ours improves model capabilities, we visualize the reconstruction results. Figure~\ref{fig:recon_visualization_harmon} demonstrates how Harmon's reconstruction quality evolves across different training timesteps, while Figure~\ref{fig:recon_visualization_models} compares reconstruction quality before and after \ours post-training across all evaluated UMM architectures.}

{As shown in Figure~\ref{fig:recon_visualization_models}, \ours post-training yields substantial improvements in reconstruction quality across all evaluated architectures:}

{\begin{itemize}[leftmargin=16pt, noitemsep, topsep=2pt]
    \item \textbf{Show-o}: Reconstruction transitions from almost noisy outputs to clearer images with better preservation of semantic content and object structures.
    \item \textbf{OpenUni, Harmon}: Reconstruction quality improves significantly, with enhanced detail preservation and more accurate rendering of geometry, color and texture.
    \item \textbf{BAGEL}: Despite already having some reconstruction capability from pre-training, \ours further enhances reconstruction quality, particularly in preserving fine-grained semantic details and spatial relationships.
\end{itemize}}

{\textbf{Training progression analysis.} Figure~\ref{fig:recon_visualization_harmon} shows how Harmon’s reconstruction ability improves over the course of \ours training. Early outputs are blurry and lack clear structure, but as training progresses, the model learns to recover object shapes, colors, and layouts more accurately. By the end, the reconstructions are sharp and semantically faithful.}

\subsection{{More Ablation Studies}}
\label{sec:ablation_data}

{To evaluate the robustness of \ours, we conduct comprehensive ablation studies on the scale and source of the training data.}

{\textbf{Data Scale Scalability.} We first evaluate Harmon-1.5B with varying amounts of MidjourneyV6~\citep{midjourneyv6} data, ranging from 10,000 to 240,000. As shown in Table~\ref{tab:data_scale_analysis}, \ours demonstrates consistent performance gains as the data scale increases, highlighting its potential for further improvement with larger datasets.}

{\textbf{Data Source Robustness.} Crucially, we investigate whether \ours relies on high-quality synthetic data. We select three distinct datasets representing different domains and quality levels: \textbf{(I)} \textbf{COCO2014}~\citep{lin2014microsoft}, representing \textit{real-world} data with relatively \textit{lower} visual quality; \textbf{(II)} \textbf{JourneyDB}~\citep{journeydb}, representing synthetic data with \textit{medium} quality; \textbf{(III)} \textbf{BLIP3o-60k}~\citep{blip3} and \textbf{(IV)} \textbf{MidjourneyV6}~\citep{midjourneyv6}, representing \textit{high-quality} synthetic data. As shown in Table~\ref{tab:data_source_analysis}, \ours maintains stable and robust performance across all sources. Notably, even when trained on lower-quality real-world images (COCO), the method yields competitive improvements (GenEval 85.3), indicating that \ours is applicable to broad data distributions.}

\begin{table}[H]
    \centering
    % === Left Table: Data Scale ===
    \begin{minipage}[t]{0.48\linewidth}
        \centering
        \caption{{\textbf{Data scale analysis.} Performance with varying amounts of MidjourneyV6 data.}}
        \label{tab:data_scale_analysis}
        \vspace{-2mm} % Adjust space between caption and table
        \footnotesize % Use smaller font to fit
        \setlength{\tabcolsep}{4pt} % Tighten column spacing
        \begin{tabular}{lccc}
            \toprule
            Data & Scale & GenEval & DPG \\
            \midrule
            MidJourneyv6 & 10,000 & 84.6 & 86.82 \\
            MidJourneyv6 & 50,000 & 85.3 & 87.06 \\
            MidJourneyv6 & 240,000 & 85.7 & 87.21 \\
            \bottomrule
        \end{tabular}
    \end{minipage}
    \hfill % This pushes the two minipages apart
    % === Right Table: Data Source ===
    \begin{minipage}[t]{0.48\linewidth}
        \centering
        \caption{{\textbf{Data source robustness.} Performance across different data qualities.}}
        \label{tab:data_source_analysis}
        \vspace{-2mm}
        \footnotesize
        \setlength{\tabcolsep}{4pt}
        \begin{tabular}{lccc}
            \toprule
            Data & Quality & GenEval & DPG \\
            \midrule
            COCO2014 & Low & 85.3 & 86.82 \\
            JourneyDB & Medium & 85.2 & 86.71 \\
            BLIP3o & High & 85.2 & 86.50 \\
            MidJourneyv6 & High & 85.7 & 87.21 \\
            \bottomrule
        \end{tabular}
    \end{minipage}
    % Optional: Joint Label if you want to refer to both
    \label{tab:ablation_combined}
    \vspace{-3mm}
\end{table}

\section{Limitations and Future Work}
\label{sec:limitations}

Despite the effectiveness of \ours across various unified multimodal models (UMMs), our approach has several limitations that warrant discussion:

\textbf{The improvements on the counting task are minor.} As shown in Table~\ref{tab:model_improvement_appendix}, the model’s gains on counting are not substantial. We attribute this to object number being a mid-level visual feature that current MLLMs are not adept at extracting. BLINK~\citep{fu2024blinkmultimodallargelanguage} points out that the information required by high-level semantic tasks (e.g., distinguishing left from right, identifying color) can often be captured and leveraged in linguistic form, whereas counting resists language-only mediation in MLLMs. Consequently, \ours primarily enhances the model’s generative ability on tasks directly tied to high-level semantics, such as color attributes and positional relations. Improving counting may require the mixture training objective of high-quality counting datasets or the incorporation of reinforcement learning techniques.

\begin{figure}[ht]
    \centering
    \includegraphics[width=\textwidth]
    {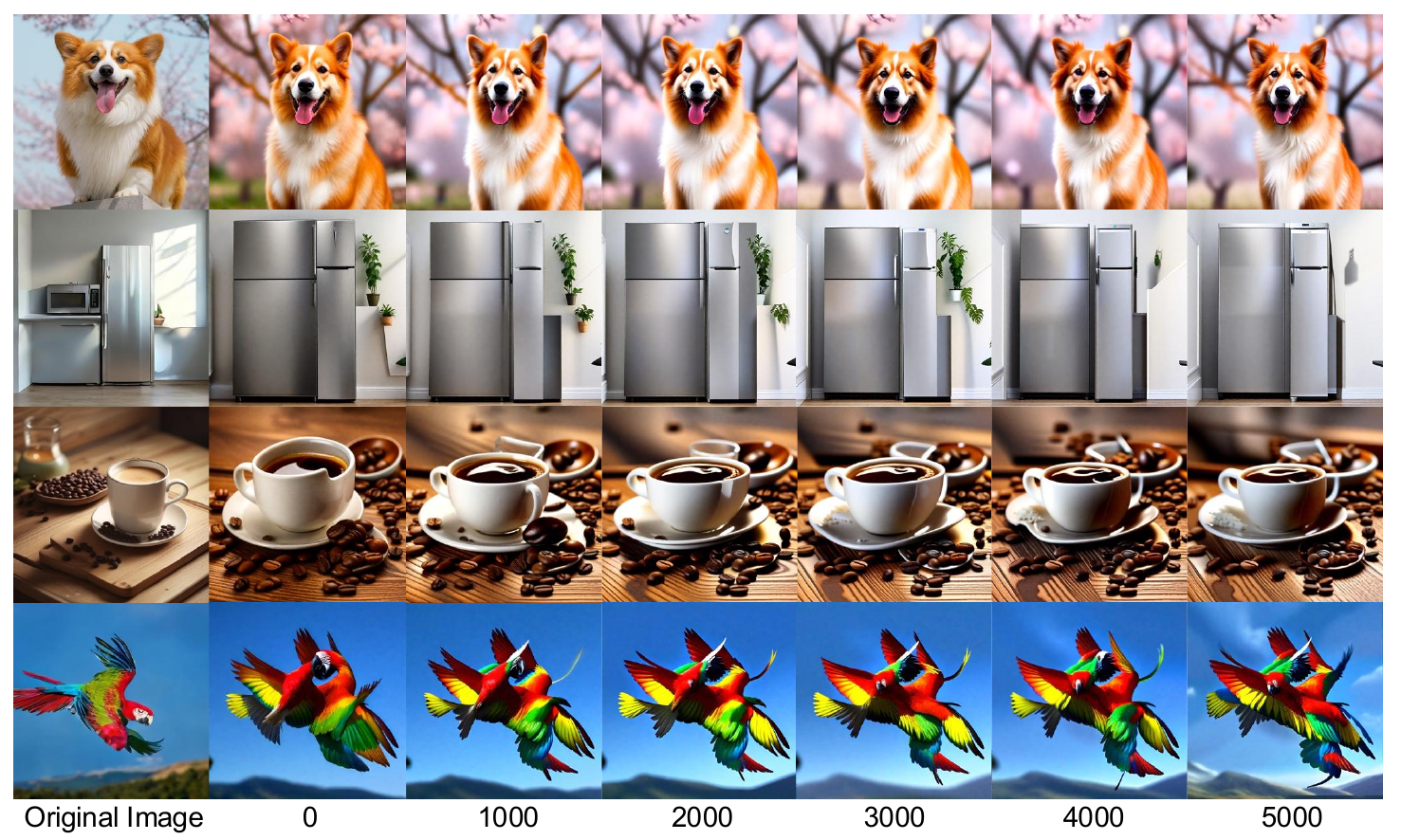}
    \caption{\textbf{BLIP-3o reconstruction results.} We observe that BLIP-3o possesses inherent image reconstruction capabilities. Following \ours application, reconstruction quality exhibits degradation rather than improvement, suggesting that BLIP-3o incorporates reconstruction objectives during pre-training.}
    \label{fig:blip3o_results}
\end{figure}

\textbf{Architecture-specific constraints.} \ours still exhibits limited effectiveness on BLIP-3o. As illustrated in Figure~\ref{fig:blip3o_results}, BLIP-3o possesses inherent image reconstruction capabilities. Application of \ours fails to improve reconstruction or text-to-image generation quality, likely because BLIP-3o's pre-training already incorporates strong reconstruction objectives, rendering additional reconstruction training counterproductive.

% \newpage

\section{Generated Captions}
\label{sec:generated_captions}

\subsection{Captions for Figure 2}

\textbf{10-word Caption}: Kitten and puppy cuddle warmly under a cozy yellow blanket.

\textbf{100-word Caption}: A heartwarming scene of pure companionship unfolds on a soft white bed where a tiny tabby kitten and a tricolor Corgi puppy nestle side by side. Wrapped gently under a fluffy mustard-yellow blanket, they gaze curiously at the viewer with wide, innocent eyes. The kitten's delicate stripes and the puppy’s soft fur blend harmoniously under the golden tones of the blanket, creating a tender atmosphere filled with comfort and warmth. Their small heads rest close, touching lightly, as if they’ve found solace in each other’s presence. The peaceful bedroom backdrop adds a serene touch to this moment of affection.

\textbf{500-word Caption}: In a warmly lit bedroom, the tranquil quiet of the morning is broken only by the soft breaths of two unlikely yet inseparable companions. Upon a pristine white bed, layered with thick cotton sheets and plush pillows, lies a tender tableau of comfort and innocence: a tabby kitten and a tricolor Corgi puppy, curled gently together beneath a soft, fleece mustard-yellow blanket. Their eyes are wide and bright, reflecting a mixture of curiosity and calm, as if they’ve just woken up from a shared dream. The kitten is small, delicate, with a soft grayish-brown tabby pattern running across its fur like brushstrokes. Its tiny paws are tucked beneath its chest, body pressed closely to the puppy’s for warmth and security. Its ears point upward slightly, twitching at the soft sounds of the morning breeze filtering through the room. Its eyes, a bright glimmering hazel, look out with an expression that is both alert and serene. The puppy, a young Pembroke Welsh Corgi, has a face full of character—expressive brown eyes, upright ears edged with tan, and a snout speckled with white. Its fur is a beautiful tricolor blend of rich black, creamy tan, and snowy white, each hue carefully painted by nature to highlight its playful features. Though still small, its body is slightly larger than the kitten’s, and its front paw is placed protectively near its feline friend. There is no tension between them—only the shared comfort of closeness, the unspoken language of trust. The blanket they share adds a golden warmth to the scene, its texture soft and inviting. It wraps around them like a cocoon, hinting at a chill in the air beyond the bed, but ensuring they remain snug in their haven. The folds of the blanket frame them like the petals of a flower, drawing the eye inward to their peaceful gaze and gentle postures. In the background, the bedroom is softly lit by morning light diffused through a window. The white and cream tones of the bedding and walls contribute to the serenity of the scene, allowing the vivid colors of the animals and their blanket to take center stage. Pillows are propped up neatly behind them, hinting that someone may have just stepped away, leaving behind these two nestled in a perfect moment of quiet affection. This image tells a silent story—not just of two animals coexisting, but of a genuine bond. It evokes themes of comfort, friendship, and the uncomplicated joy of presence. It reminds us of the purity in companionship, of how safety and peace can be found not just in familiar places, but in the closeness of those we trust. The kitten and the puppy, young and full of life, are captured here in a fleeting, gentle stillness that speaks to the heart.

\subsection{Captions for Figure 8}

\begin{figure}[t]
    \vspace{-0.5em}
    \centering
    \includegraphics[width=\textwidth]{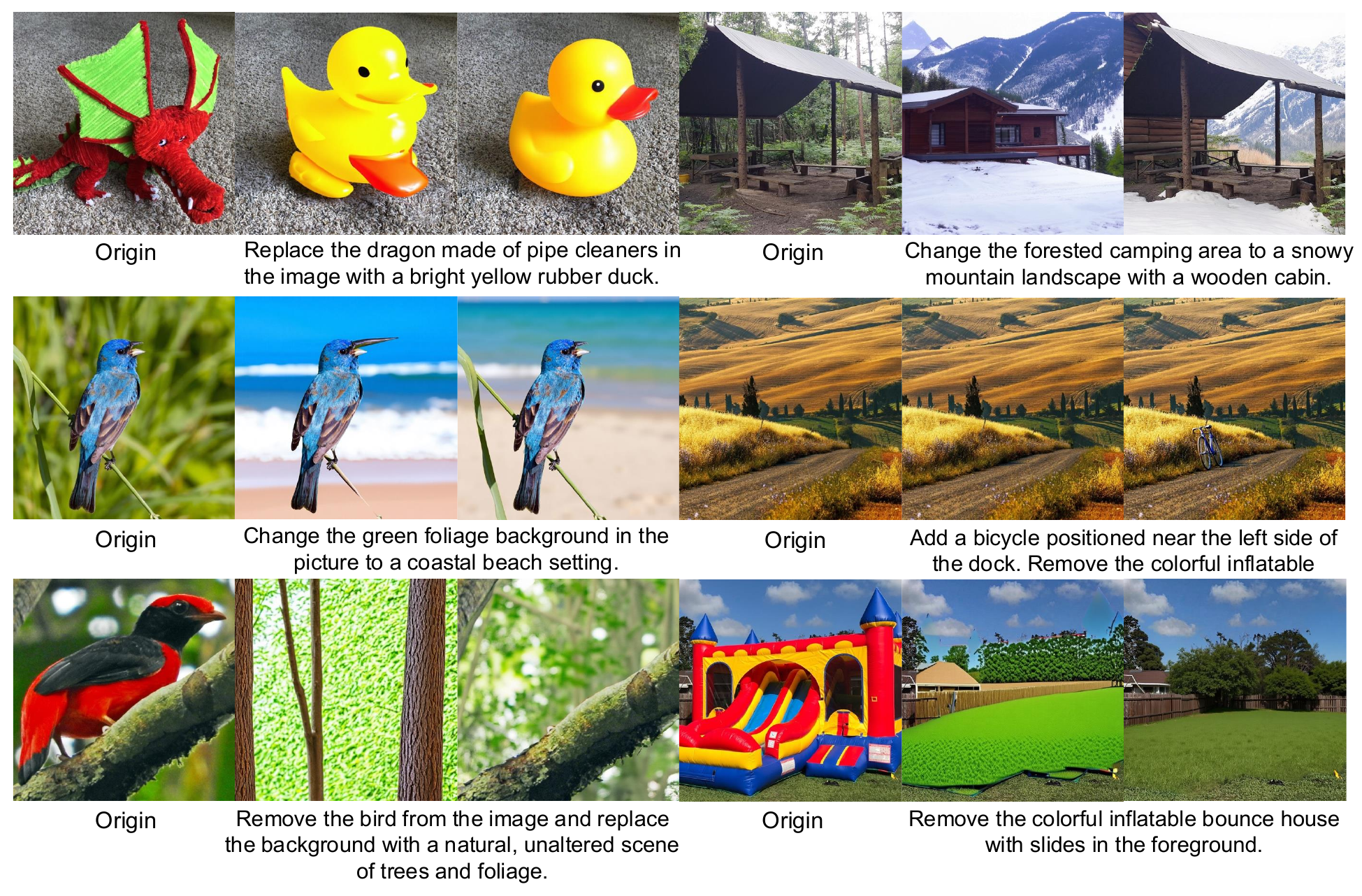}
    \vspace{-0.5em}
    \caption{\textbf{Additional ImgEdit benchmark results.} Qualitative results on image editing tasks. In each pair, the left image is generated by the baseline model and the right by the \ours post-trained model}
    \label{fig:more_on_imgedit}
    \vspace{-0.5em}
\end{figure}

$512\times 512$ images (left to right, top to bottom):

\begin{itemize}[leftmargin=*, noitemsep, topsep=0pt]
\item Two spiraling strands of rich, crimson-colored pasta rest elegantly on the surface of a polished dark wooden table, the grain of the wood accentuating their vibrant hue. This rustic Italian kitchen is bathed in the warm, golden light of the late afternoon sun, which highlights the intricate texture of the pasta. The table, set amidst traditional décor and terracotta pots filled with fresh herbs, offers a tranquil setting for this simple yet captivating culinary display.
\item A golden squirrel and a bright magenta elephant in bright sunlight.
\item During the warm glow of a dwindling summer evening, a particular fussy feline with distinctive calico markings is perched atop a garden table. The cat, seemingly indifferent to its surroundings, sports a pair of large, reflective aviator sunglasses that sit comically upon its small, furry face. Around the cat, there are scattered pots of blooming flowers, contributing to the charm of the scene, and in the background, hints of orange and pink skies are visible through the foliage.
\item a highly intricate and vibrant cityscape that reflects a fusion of Moebius's imaginative design and Makoto Shinkai's detailed animation style. The streets are aglow with neon signs in a kaleidoscope of colors, casting reflections on the glossy, rain-slicked pavements. Towering skyscrapers with glowing windows rise towards a starless night sky, as the artwork garners significant attention and praise on ArtStation.

\item A deep red rose with plush petals sits elegantly coiled atop an ivory, intricately patterned lace napkin. The napkin rests on a rustic wooden table that contributes to the charming garden setting. As the late evening sun casts a warm golden hue over the area, the shadows of surrounding foliage dance gently around the rose, enhancing the romantic ambiance. Nearby, the green leaves of the garden plants provide a fresh and verdant backdrop to the scene.

\item A close-up image capturing the intricate details of a maple leaf, which is composed entirely of clear, sparkling water droplets. The leaf is set against a smooth, dark background that accentuates its delicate water structure. The droplets glisten as they cling to the invisible veins of the leaf, creating a natural yet surreal piece of art.

\item A detailed photograph captures the image of a statue with the likeness of an ancient pharaoh, unexpectedly accessorized with a pair of bronze steampunk goggles resting atop its head. The statue is dressed in an anachronistic fashion, featuring a crisp white t-shirt and a fitted black leather jacket that contrasts with its traditional headdress. The background is a simple, solid color that accentuates the statue's unconventional attire and the intricate details of the steampunk eyewear.

\item On the soft, warm sand of the beach, a fluffy white rabbit with rounded ears is caught in a curious moment, gently placing its paw on the ribbed surface of a pink scallop shell. The scallop, slightly open, reveals its smooth interior contrasting with its coarse outer texture, while hues of pink and orange from the setting sun reflect off its surface. There's a tranquil ocean backdrop with the gentle ebbing of the tide, and the fading daylight casts a golden glow over the scene, highlighting the rabbit's soft fur and the shell's subtle color.

\end{itemize}

\begin{figure}[t]
    \vspace{-0.5em}
    \centering
    \includegraphics[width=\textwidth]{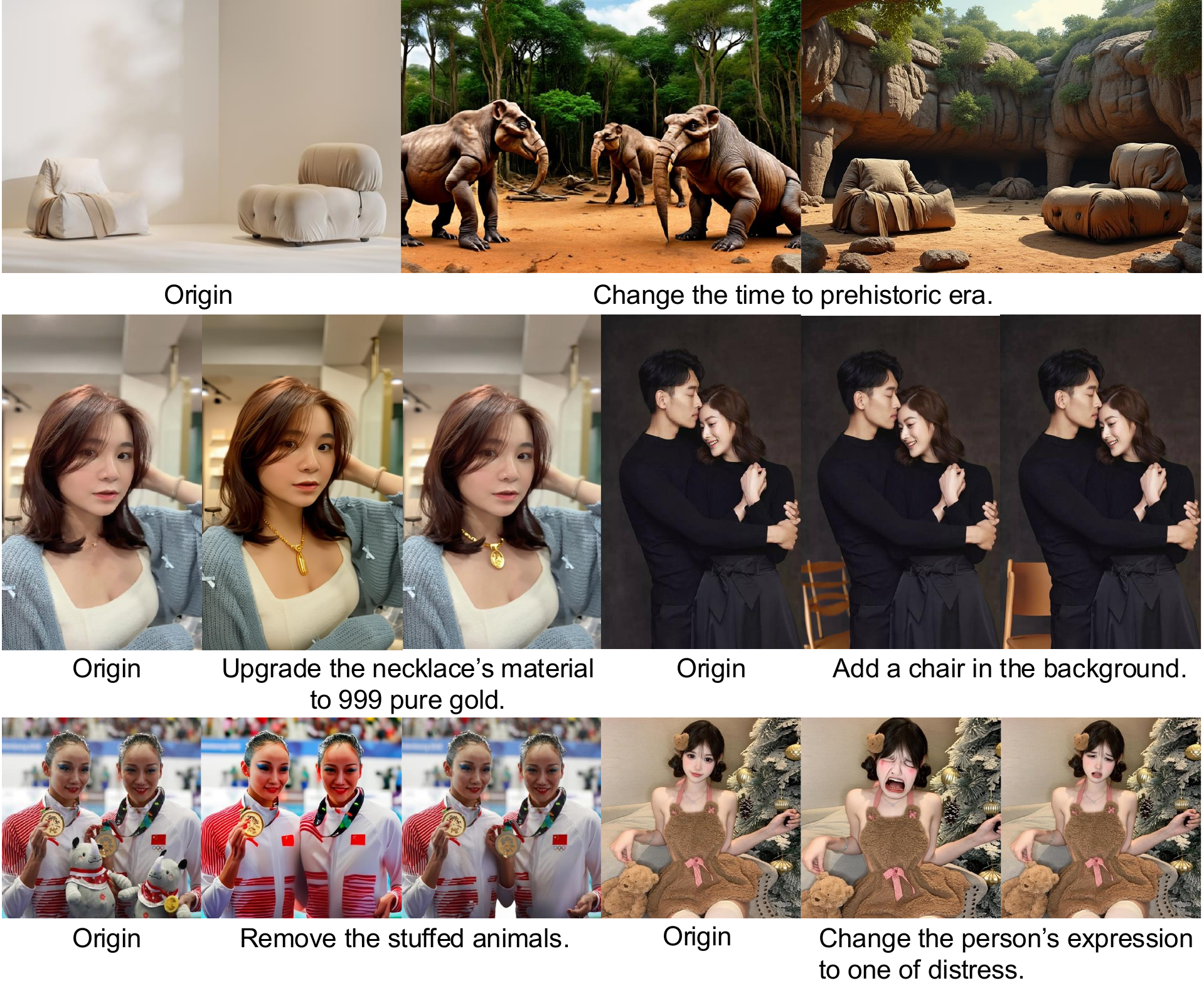}
    \vspace{-0.5em}
    \caption{\textbf{Additional GEdit-Bench-EN results.} Qualitative results on image editing tasks. In each pair, the left image is generated by the baseline BAGEL model and the right by the \ours post-trained model.}
    \label{fig:more_on_gedit}
    \vspace{-0.5em}
\end{figure}

$1024\times 1024$ images (left to right, top to bottom):

\begin{itemize}[leftmargin=*, noitemsep, topsep=0pt]
\item A transparent glass cube on an endless desert, with a burning candle inside casting shadows on the sand.
\item A drop of water containing a miniature forest with colorful tiny flower inside.
\item A vibrant and traditional depiction of Tteokguk, a Korean rice cake soup, served during the Chuseok festival. The image features a steaming bowl of clear broth filled with soft, chewy rice cakes, slices of zucchini, carrots, and green onions, garnished with a sprinkle of sesame seeds. Surrounding the bowl are traditional Korean elements, such as a woven basket with red dates, a small wooden spoon, and a wooden table with a warm, earthy tone. The atmosphere is cozy and festive, with soft, natural lighting and a slightly blurred background to emphasize the dish. The scene captures the essence of Korean culture and the warmth of the Chuseok celebration, with a focus on authenticity and detail.
\item Photorealistic closeup image of two pirate ships battling each other as they sail inside a cup of coffee.
\item A cute handmade felt doll of a little girl standing on a grassy patch. She wears an orange knitted hooded dress with blue buttons and matching boots. The girl is holding a smiling sun-shaped balloon made of felt. Fluffy white clouds with button details float in the sky, and soft green felt trees surround the scene. The style is whimsical, cozy, and playful, with pastel colors and a dreamy, handcrafted aesthetic.
\item The word 'RECA' is written on a street surface, with the word 'STARTS' written just below it, surrounded by colorful chalk drawings and playful doodles.
\item A Van Gogh style painting of a cyberpunk city at sunrise.
\item An anime-style portrait of a girl with big sparkling eyes, detailed hair highlights, soft gradient background, vibrant colors.
\item A dreamy composition of a young woman with butterflies emerging from her skin, wings glowing in soft golden hues, surreal and enchanting. She is wearing a wedding dress and has white angel wings, waving her hand.
\item A surreal split-face portrait: left side realistic woman with soft skin and a vivid blue eye, right side robotic cyborg with exposed steel plates, fluorescent blue circuits, tiny gears, and a blood-red mechanical eye, cinematic lighting, futuristic and striking.
\end{itemize}

\section{More Qualitative Results on Image Editing}
\label{sec:more_image_editing}

\begin{figure}[H]
    \vspace{-2em}
    \centering
    \includegraphics[width=\textwidth]{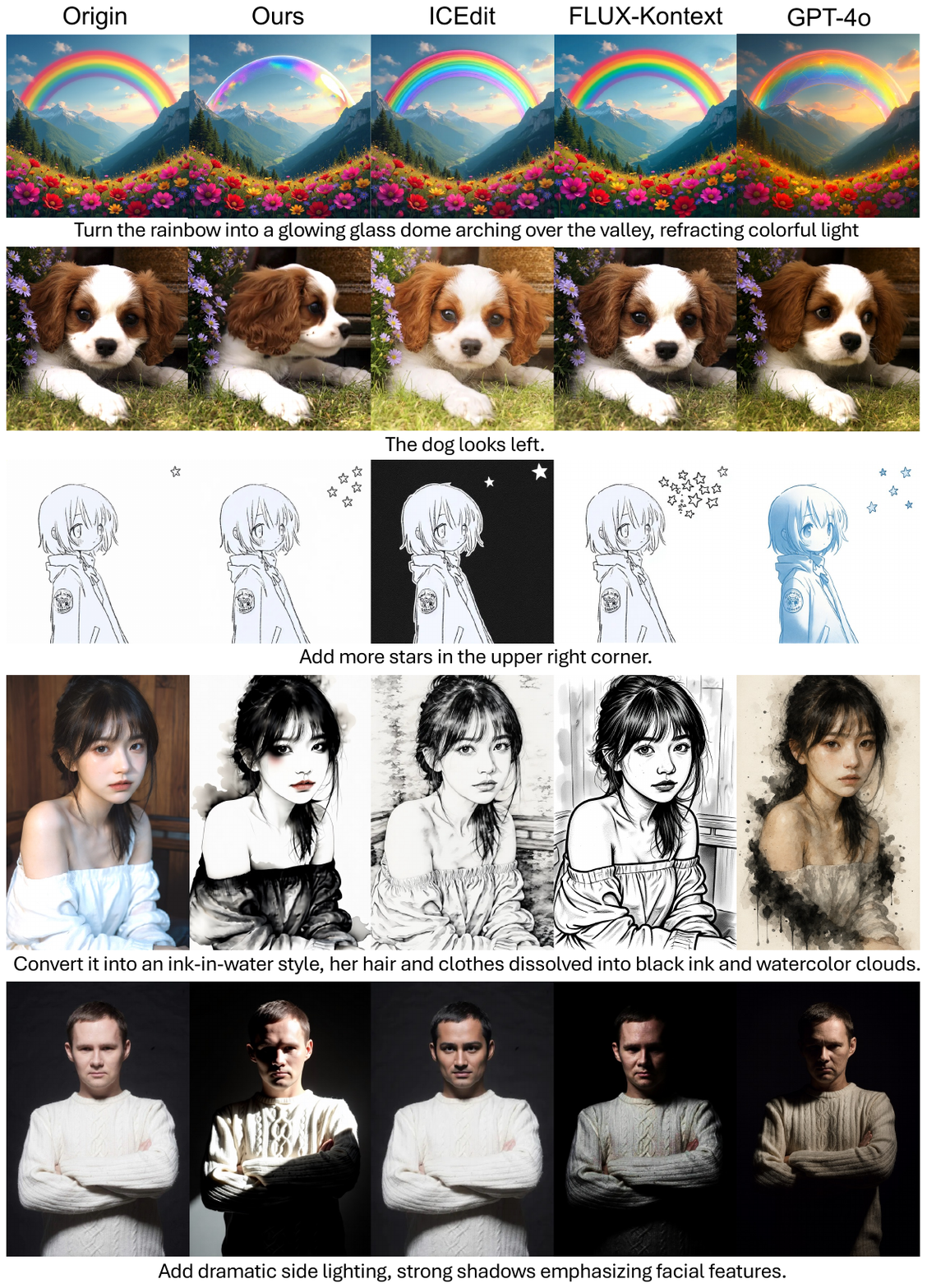}
    \vspace{-2em}
    \caption{\textbf{Qualitative comparison of editing results across different methods}. Our method achieves more faithful instruction following (e.g., rainbow dome, star addition), better identity preservation (e.g., pet and human faces), and stronger background consistency.}
    \label{fig:edit_comparison}
    \vspace{-0.5em}
\end{figure}

We provide additional qualitative comparisons in this section. Figure~\ref{fig:edit_comparison} presents a comparison between the \ours post-trained BAGEL and previous SOTA models, ICEdit~\citep{zhang2025context}, FLUX-Kontext~\citep{labs2025flux}, and GPT-4o-Image. Compared with ICEdit and FLUX-Kontext, our method demonstrates stronger instruction-following capability. In contrast to GPT-4o-Image, our method exhibits superior identity preservation and background fidelity. Figure~\ref{fig:more_on_imgedit} and~\ref{fig:more_on_gedit} showcase consistent improvements in semantic consistency, instruction following, and visual quality preservation across various editing tasks including background modification, style transfer, and object manipulation.

\newpage

\newpage

\section{More Qualitative Results on Text-to-Image Generation}
\label{sec:more_qualitative_results}

\begin{figure}[H]
    \vspace{-0.5em}
    \centering
    \includegraphics[width=0.76\textwidth,page=1]{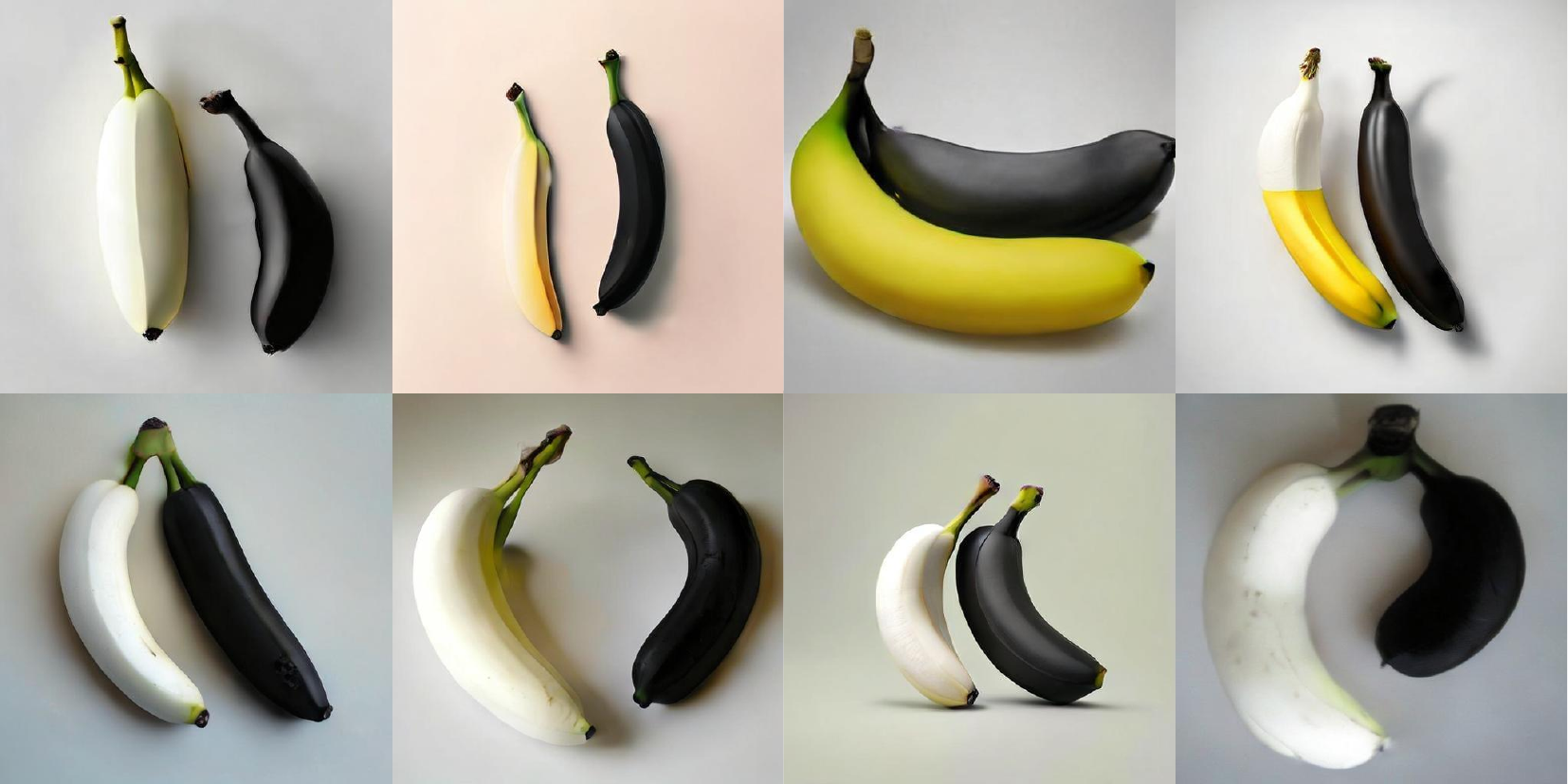}
    \vspace{-0.5em}
    \caption{\textbf{Uncurated generation results from Harmon-1.5B (top) and post-trained model (bottom)}. Prompt: A white banana and a black banana.}
    \label{fig:white_black_banana}
    \vspace{-0.5em}
\end{figure}

\begin{figure}[H]
    \vspace{-0.5em}
    \centering
    \includegraphics[width=0.76\textwidth,page=2]{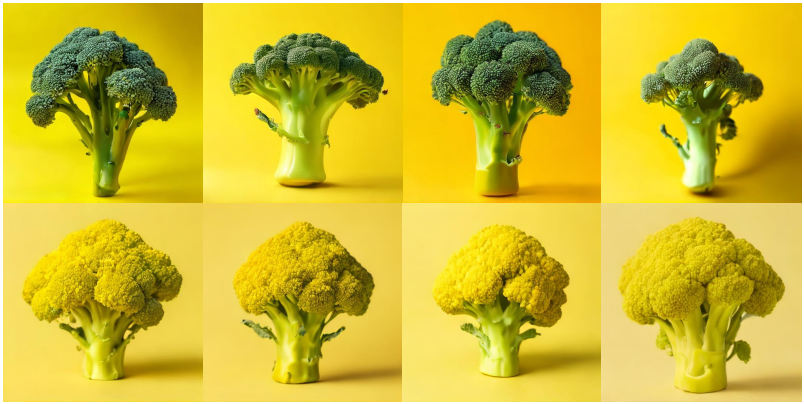}
    \vspace{-0.5em}
    \caption{\textbf{Uncurated generation results from Harmon-1.5B (top) and post-trained model (bottom)}. Prompt: A photo of a yellow broccoli.}
    \label{fig:yellow_broccoli}
    \vspace{-0.5em}
\end{figure}

\begin{figure}[H]
    \vspace{-0.5em}
    \centering
    \includegraphics[width=0.76\textwidth,page=3]
    {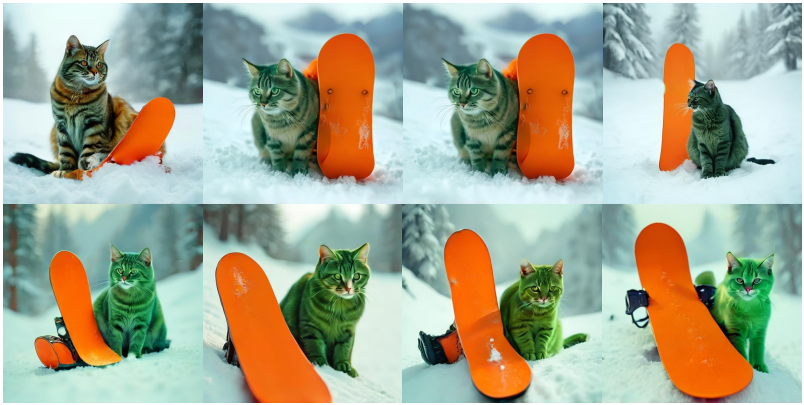}
    \vspace{-0.5em}
    \caption{\textbf{Uncurated generation results from Harmon-1.5B (top) and post-trained model (bottom)}. Prompt: A photo of an orange snowboard and a green cat.}
    \label{fig:orange_snowboard_cat}
    \vspace{-0.5em}
\end{figure}

\newpage

\begin{figure}[H]
    \vspace{-0.5em}
    \centering
    \includegraphics[width=0.76\textwidth,page=5]{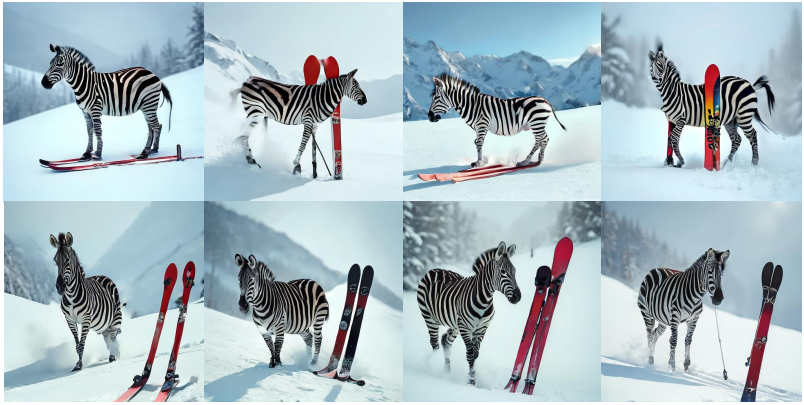}
    \vspace{-0.5em}
    \caption{\textbf{Uncurated generation results from Harmon-1.5B (top) and post-trained model (bottom)}. Prompt: A photo of a skis right of a zebra.}
    \label{fig:skis_zebra}
    \vspace{-0.5em}
\end{figure}

\begin{figure}[H]
    \vspace{-0.5em}
    \centering
    \includegraphics[width=0.76\textwidth,page=4]{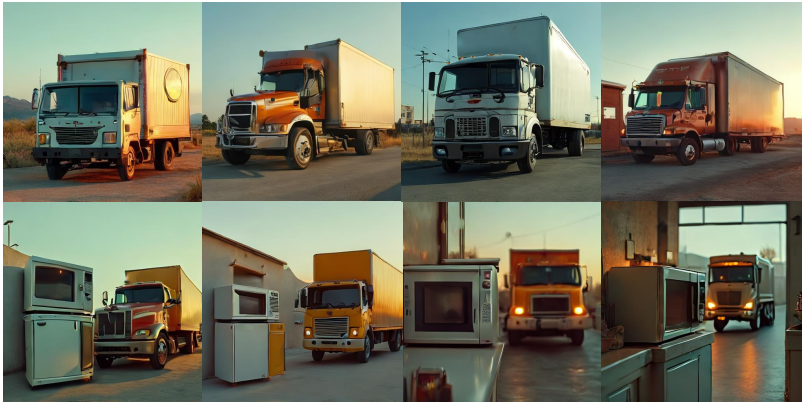}
    \vspace{-0.5em}
    \caption{\textbf{Uncurated generation results from Harmon-1.5B (top) and post-trained model (bottom)}. Prompt: A photo of a microwave and a truck.}
    \label{fig:microwave_truck}
    \vspace{-0.5em}
\end{figure}

\begin{figure}[H]
    \vspace{-1em}
    \centering
    \includegraphics[width=0.76\textwidth,page=10]{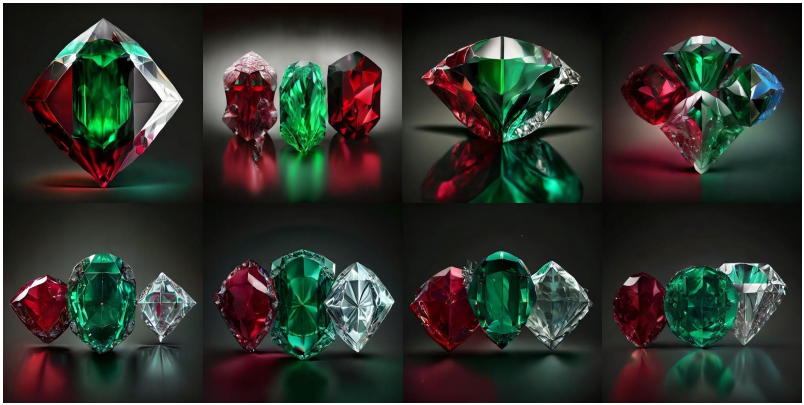}
    \vspace{-0.5em}
    \caption{\textbf{Uncurated generation results from Harmon-1.5B (top) and post-trained model (bottom)}. Prompt: A diamond on the right, an emerald in the middle, a ruby on the left.}
    \label{fig:diamond_emerald_ruby}
    \vspace{-1em}
\end{figure}

\begin{figure}[H]
    \vspace{-1em}
    \centering
    \includegraphics[width=0.76\textwidth,page=11]{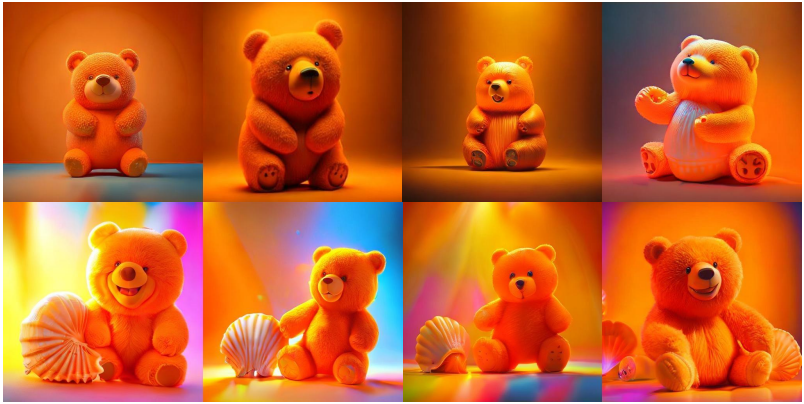}
    \vspace{-0.5em}
    \caption{\textbf{Uncurated generation results from Harmon-1.5B (top) and post-trained model (bottom)}. Prompt: Cheerful and bright, vibrant lighting. A shell and a bright orange bear in a bright setting.}
    \vspace{-1em}
\end{figure}

\begin{figure}[H]
    \vspace{-0.5em}
    \centering
    \includegraphics[width=0.76\textwidth,page=6]{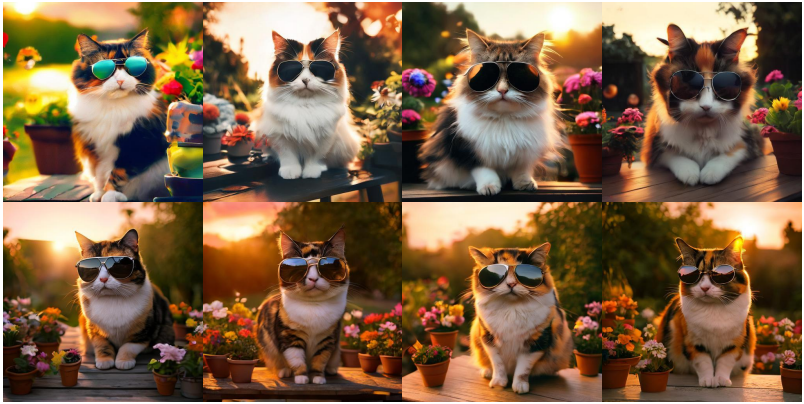}
    \vspace{-0.5em}
    \caption{\textbf{Uncurated generation results from Harmon-1.5B (top) and post-trained model (bottom)}. Prompt: During the warm glow of a dwindling summer evening, a particular fussy feline with distinctive calico markings is perched atop a garden table. The cat, seemingly indifferent to its surroundings, sports a pair of large, reflective aviator sunglasses that sit comically upon its small, furry face. Around the cat, there are scattered pots of blooming flowers, contributing to the charm of the scene, and in the background, hints of orange and pink skies are visible through the foliage.}
    \label{fig:cat_sunglasses}
    \vspace{-0.5em}
\end{figure}

\begin{figure}[H]
    \vspace{-0.5em}
    \centering
    \includegraphics[width=0.76\textwidth,page=7]{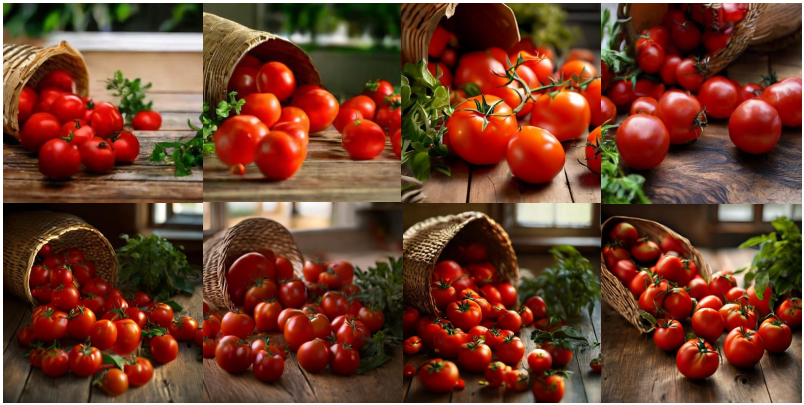}
    \vspace{-0.5em}
    \caption{\textbf{Uncurated generation results from Harmon-1.5B (top) and post-trained model (bottom)}. Prompt: A vivid scene unfolds where several deep red, perfectly round tomatoes spill from a woven brown basket onto a rustic wooden tabletop. The basket lies on its side as the plump tomatoes scatter across the surface, some touching the dark green leaves of a nearby herb plant. In the background, the blurred outline of an open kitchen window lets in soft, natural light, casting gentle shadows around the fallen produce.}
    \label{fig:tomatoes_basket}
    \vspace{-0.5em}
\end{figure}

\begin{figure}[H]
    \vspace{-0.5em}
    \centering
    \includegraphics[width=0.76\textwidth,page=8]{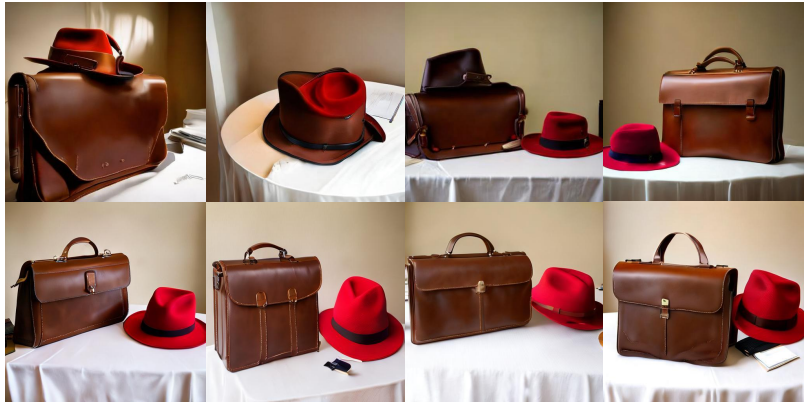}
    \vspace{-0.5em}
    \caption{\textbf{Uncurated generation results from Harmon-1.5B (top) and post-trained model (bottom)}. Prompt: A polished brown leather briefcase with visible stitching details rests on a white tablecloth, displaying a sense of organization amidst the surrounding environment. Beside the briefcase, a vibrant red fedora hat provides a striking contrast against the pristine table covering. The table, placed in a room with light beige walls, gives an impression of a professional setting with a touch of personal style.}
    \label{fig:briefcase_hat}
    \vspace{-0.5em}
\end{figure}

\begin{figure}[H]
    \vspace{-0.5em}
    \centering
    \includegraphics[width=0.76\textwidth,page=9]{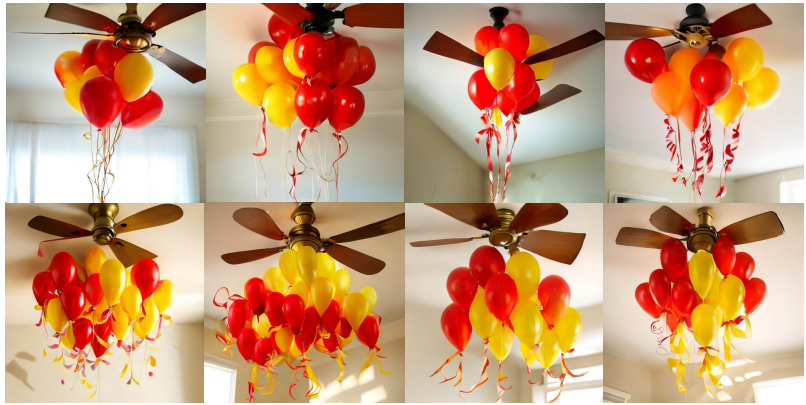}
    \vspace{-0.5em}
    \caption{\textbf{Uncurated generation results from Harmon-1.5B (top) and post-trained model (bottom)}. Prompt: a festive array of red and yellow balloons tied with curling ribbons, gently bobbing from the breeze of a spinning ceiling fan. The fan has wooden blades and a brass finish, which contrasts with the bright colors of the balloons. The balloons are clustered in a joyful bunch, casting soft shadows on the ceiling above.}
    \label{fig:balloons_fan}
    \vspace{-0.5em}
\end{figure}

\begin{figure}[H]
    \vspace{-0.5em}
    \centering
    \includegraphics[width=0.9\textwidth,page=1]{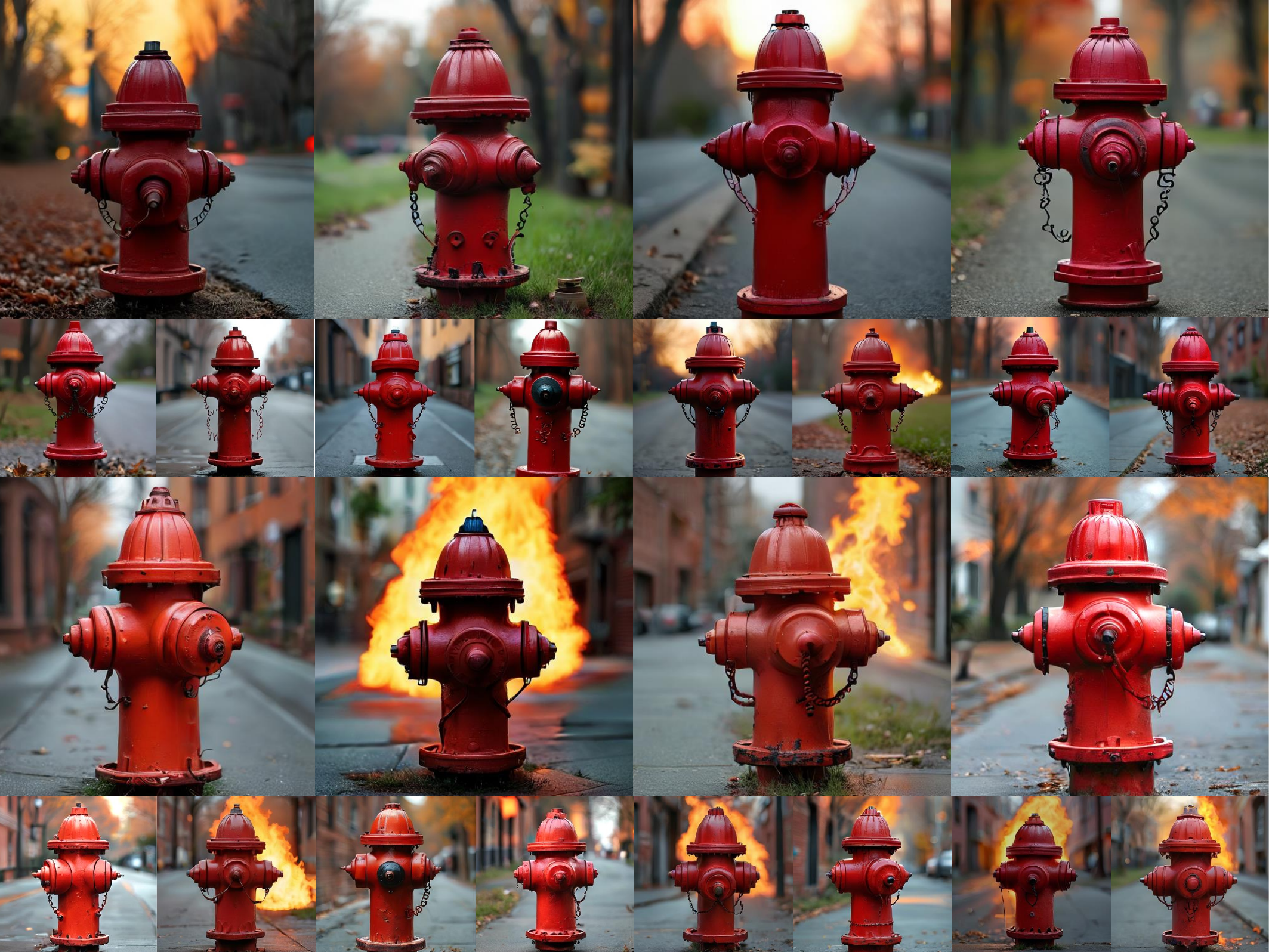}
    \vspace{-0.5em}
    \caption{{\textbf{Uncurated generation results from Harmon-1.5B (top) and post-trained model (bottom)}. Prompt: a photo of a fire hydrant.}}
    % \label{fig:balloons_fan}
    \vspace{-0.5em}
\end{figure}

\begin{figure}[H]
    \vspace{-0.5em}
    \centering
    \includegraphics[width=0.9\textwidth,page=2]{figures/diversity.pdf}
    \vspace{-0.5em}
    \caption{{\textbf{Uncurated generation results from BAGEL (top) and post-trained model (bottom)}. Prompt: a photo of a surfboard.}}
    % \label{fig:balloons_fan}
    \vspace{-0.5em}
\end{figure}

\begin{figure}[H]
    \vspace{-0.5em}
    \centering
    \includegraphics[width=0.9\textwidth,page=3]{figures/diversity.pdf}
    \vspace{-0.5em}
    \caption{{\textbf{Uncurated generation results from OpenUni (top) and post-trained model (bottom)}. Prompt: a photo of a suitcase.}}
    % \label{fig:balloons_fan}
    \vspace{-0.5em}
\end{figure}

\begin{figure}[H]
    \vspace{-0.5em}
    \centering
    \includegraphics[width=0.9\textwidth,page=4]{figures/diversity.pdf}
    \vspace{-0.5em}
    \caption{{\textbf{Uncurated generation results from Show-o-512x512 (top) and post-trained model (bottom)}. Prompt: a photo of a cup.}}
    % \label{fig:balloons_fan}
    \vspace{-0.5em}
\end{figure}

\section{The Use of Large Language Models (LLMs)}

\label{sec:llm_usage}

During the preparation of this paper, we used large language model (LLM) only as a writing assistant to check grammar, spelling, and to polish the clarity of expression. The LLM was not involved in the research design, experiments, or analysis, and the authors take full responsibility for all content of this work.

%% file: iclr2025_conference.bib
@article{openuni,
  title={OpenUni: A Simple Baseline for Unified Multimodal Understanding and Generation},
  author={Wu, Size and Wu, Zhonghua and Gong, Zerui and Tao, Qingyi and Jin, Sheng and Li, Qinyue and Li, Wei and Loy, Chen Change},
  journal={arXiv preprint arXiv:2505.23661},
  year={2025}
}

@article{zhu2025internvl3,
  title={Internvl3: Exploring advanced training and test-time recipes for open-source multimodal models},
  author={Zhu, Jinguo and Wang, Weiyun and Chen, Zhe and Liu, Zhaoyang and Ye, Shenglong and Gu, Lixin and Tian, Hao and Duan, Yuchen and Su, Weijie and Shao, Jie and others},
  journal={arXiv preprint arXiv:2504.10479},
  year={2025}
}

@article{wei2022chain,
  title={Chain-of-thought prompting elicits reasoning in large language models},
  author={Wei, Jason and Wang, Xuezhi and Schuurmans, Dale and Bosma, Maarten and Xia, Fei and Chi, Ed and Le, Quoc V and Zhou, Denny and others},
  journal={Advances in neural information processing systems},
  volume={35},
  pages={24824--24837},
  year={2022}
}

@article{rafailov2023direct,
  title={Direct preference optimization: Your language model is secretly a reward model},
  author={Rafailov, Rafael and Sharma, Archit and Mitchell, Eric and Manning, Christopher D and Ermon, Stefano and Finn, Chelsea},
  journal={Advances in neural information processing systems},
  volume={36},
  pages={53728--53741},
  year={2023}
}

@article{shao2024deepseekmath,
  title={Deepseekmath: Pushing the limits of mathematical reasoning in open language models},
  author={Shao, Zhihong and Wang, Peiyi and Zhu, Qihao and Xu, Runxin and Song, Junxiao and Bi, Xiao and Zhang, Haowei and Zhang, Mingchuan and Li, YK and Wu, Yang and others},
  journal={arXiv preprint arXiv:2402.03300},
  year={2024}
}

@article{wei2025skywork,
  title={Skywork UniPic 2.0: Building Kontext Model with Online RL for Unified Multimodal Model},
  author={Wei, Hongyang and Xu, Baixin and Liu, Hongbo and Wu, Cyrus and Liu, Jie and Peng, Yi and Wang, Peiyu and Liu, Zexiang and He, Jingwen and Xietian, Yidan and others},
  journal={arXiv preprint arXiv:2509.04548},
  year={2025}
}

@article{fan2025unified,
  title={Unified autoregressive visual generation and understanding with continuous tokens},
  author={Fan, Lijie and Tang, Luming and Qin, Siyang and Li, Tianhong and Yang, Xuan and Qiao, Siyuan and Steiner, Andreas and Sun, Chen and Li, Yuanzhen and Zhu, Tao and others},
  journal={arXiv preprint arXiv:2503.13436},
  year={2025}
}

@article{chen2025blip3,
  title={Blip3-o: A family of fully open unified multimodal models-architecture, training and dataset},
  author={Chen, Jiuhai and Xu, Zhiyang and Pan, Xichen and Hu, Yushi and Qin, Can and Goldstein, Tom and Huang, Lifu and Zhou, Tianyi and Xie, Saining and Savarese, Silvio and others},
  journal={arXiv preprint arXiv:2505.09568},
  year={2025}
}

@article{yan2025gpt,
  title={Gpt-imgeval: A comprehensive benchmark for diagnosing gpt4o in image generation},
  author={Yan, Zhiyuan and Ye, Junyan and Li, Weijia and Huang, Zilong and Yuan, Shenghai and He, Xiangyang and Lin, Kaiqing and He, Jun and He, Conghui and Yuan, Li},
  journal={arXiv preprint arXiv:2504.02782},
  year={2025}
}

@inproceedings{chang2022maskgit,
  title={Maskgit: Masked generative image transformer},
  author={Chang, Huiwen and Zhang, Han and Jiang, Lu and Liu, Ce and Freeman, William T},
  booktitle={CVPR},
  pages={11315--11325},
  year={2022}
}

@article{allakhverdov2025image,
  title={Image Reconstruction as a Tool for Feature Analysis},
  author={Allakhverdov, Eduard and Tarasov, Dmitrii and Goncharova, Elizaveta and Kuznetsov, Andrey},
  journal={arXiv preprint arXiv:2506.07803},
  year={2025}
}

@article{labs2025flux,
  title={FLUX. 1 Kontext: Flow Matching for In-Context Image Generation and Editing in Latent Space},
  author={Labs, Black Forest and Batifol, Stephen and Blattmann, Andreas and Boesel, Frederic and Consul, Saksham and Diagne, Cyril and Dockhorn, Tim and English, Jack and English, Zion and Esser, Patrick and others},
  journal={arXiv preprint arXiv:2506.15742},
  year={2025}
}

@article{vqvae,
  title={Neural discrete representation learning},
  author={Van Den Oord, Aaron and Vinyals, Oriol and others},
  journal={NeurIPS},
  volume={30},
  year={2017}
}

@inproceedings{vqgan,
  title={Taming transformers for high-resolution image synthesis},
  author={Esser, Patrick and Rombach, Robin and Ommer, Bjorn},
  booktitle={CVPR},
  pages={12873--12883},
  year={2021}
}

@article{sdxl,
  title={Sdxl: Improving latent diffusion models for high-resolution image synthesis},
  author={Podell, Dustin and English, Zion and Lacey, Kyle and Blattmann, Andreas and Dockhorn, Tim and M{\"u}ller, Jonas and Penna, Joe and Rombach, Robin},
  journal={arXiv preprint arXiv:2307.01952},
  year={2023}
}

@article{team2024chameleon,
  title={Chameleon: Mixed-modal early-fusion foundation models},
  author={Team, Chameleon},
  journal={arXiv preprint arXiv:2405.09818},
  year={2024}
}

@article{seed-x,
  title={Seed-x: Multimodal models with unified multi-granularity comprehension and generation},
  author={Ge, Yuying and Zhao, Sijie and Zhu, Jinguo and Ge, Yixiao and Yi, Kun and Song, Lin and Li, Chen and Ding, Xiaohan and Shan, Ying},
  journal={arXiv preprint arXiv:2404.14396},
  year={2024}
}

@article{llava,
  title={Visual instruction tuning},
  author={Liu, Haotian and Li, Chunyuan and Wu, Qingyang and Lee, Yong Jae},
  journal={NeurIPS},
  volume={36},
  year={2024}
}

@article{li2024autoregressive,
  title={Autoregressive image generation without vector quantization},
  author={Li, Tianhong and Tian, Yonglong and Li, He and Deng, Mingyang and He, Kaiming},
  journal={Advances in Neural Information Processing Systems},
  volume={37},
  pages={56424--56445},
  year={2024}
}

@article{yan2025unified,
  title={Unified Multimodal Model as Auto-Encoder},
  author={Yan, Zhiyuan and Lin, Kaiqing and Li, Zongjian and Ye, Junyan and Han, Hui and Wang, Zhendong and Liu, Hao and Lin, Bin and Li, Hao and Xu, Xue and others},
  journal={arXiv preprint arXiv:2509.09666},
  year={2025}
}

@article{an2025unictokens,
  title={UniCTokens: Boosting Personalized Understanding and Generation via Unified Concept Tokens},
  author={An, Ruichuan and Yang, Sihan and Zhang, Renrui and Shen, Zijun and Lu, Ming and Dai, Gaole and Liang, Hao and Guo, Ziyu and Yan, Shilin and Luo, Yulin and others},
  journal={arXiv preprint arXiv:2505.14671},
  year={2025}
}

@article{ye2025echo,
  title={Echo-4o: Harnessing the power of gpt-4o synthetic images for improved image generation},
  author={Ye, Junyan and Jiang, Dongzhi and Wang, Zihao and Zhu, Leqi and Hu, Zhenghao and Huang, Zilong and He, Jun and Yan, Zhiyuan and Yu, Jinghua and Li, Hongsheng and others},
  journal={arXiv preprint arXiv:2508.09987},
  year={2025}
}

@article{niu2025does,
  title={Does Understanding Inform Generation in Unified Multimodal Models? From Analysis to Path Forward},
  author={Niu, Yuwei and Jin, Weiyang and Liao, Jiaqi and Feng, Chaoran and Jin, Peng and Lin, Bin and Li, Zongjian and Zhu, Bin and Yu, Weihao and Yuan, Li},
  journal={arXiv preprint arXiv:2511.20561},
  year={2025}
}

@article{guo2025can,
  title={Can We Generate Images with CoT? Let's Verify and Reinforce Image Generation Step by Step},
  author={Guo, Ziyu and Zhang, Renrui and Tong, Chengzhuo and Zhao, Zhizheng and Gao, Peng and Li, Hongsheng and Heng, Pheng-Ann},
  journal={arXiv:2501.13926},
  year={2025}
}

@article{wang2025mint,
  title={MINT: Multi-modal Chain of Thought in Unified Generative Models for Enhanced Image Generation},
  author={Wang, Yi and Liu, Mushui and He, Wanggui and Zhang, Longxiang and Huang, Ziwei and Zhang, Guanghao and Shu, Fangxun and Tao, Zhong and She, Dong and Yu, Zhelun and others},
  journal={arXiv:2503.01298},
  year={2025}
}

@article{fang2025got,
  title={Got: Unleashing reasoning capability of multimodal large language model for visual generation and editing},
  author={Fang, Rongyao and Duan, Chengqi and Wang, Kun and Huang, Linjiang and Li, Hao and Yan, Shilin and Tian, Hao and Zeng, Xingyu and Zhao, Rui and Dai, Jifeng and others},
  journal={arXiv:2503.10639},
  year={2025}
}

@article{llama,
  author       = {Hugo Touvron and
                  Thibaut Lavril and
                  Gautier Izacard and
                  Xavier Martinet and
                  Marie{-}Anne Lachaux and
                  Timoth{\'{e}}e Lacroix and
                  Baptiste Rozi{\`{e}}re and
                  Naman Goyal and
                  Eric Hambro and
                  Faisal Azhar and
                  Aur{\'{e}}lien Rodriguez and
                  Armand Joulin and
                  Edouard Grave and
                  Guillaume Lample},
  title        = {LLaMA: Open and Efficient Foundation Language Models},
  journal      = {CoRR},
  volume       = {abs/2302.13971},
  year         = {2023}
}

@inproceedings{gpt3,
  author       = {Tom B. Brown and
                  Benjamin Mann and
                  Nick Ryder and
                  Melanie Subbiah and
                  Jared Kaplan and
                  Prafulla Dhariwal and
                  Arvind Neelakantan and
                  Pranav Shyam and
                  Girish Sastry and
                  Amanda Askell and
                  Sandhini Agarwal and
                  Ariel Herbert{-}Voss and
                  Gretchen Krueger and
                  Tom Henighan and
                  Rewon Child and
                  Aditya Ramesh and
                  Daniel M. Ziegler and
                  Jeffrey Wu and
                  Clemens Winter and
                  Christopher Hesse and
                  Mark Chen and
                  Eric Sigler and
                  Mateusz Litwin and
                  Scott Gray and
                  Benjamin Chess and
                  Jack Clark and
                  Christopher Berner and
                  Sam McCandlish and
                  Alec Radford and
                  Ilya Sutskever and
                  Dario Amodei},
  title        = {Language Models are Few-Shot Learners},
  booktitle    = {NeurIPS},
  year         = {2020}
}

@inproceedings{clip,
  author       = {Alec Radford and
                  Jong Wook Kim and
                  Chris Hallacy and
                  Aditya Ramesh and
                  Gabriel Goh and
                  Sandhini Agarwal and
                  Girish Sastry and
                  Amanda Askell and
                  Pamela Mishkin and
                  Jack Clark and
                  Gretchen Krueger and
                  Ilya Sutskever},
  title        = {Learning Transferable Visual Models From Natural Language Supervision},
  booktitle    = {{ICML}},
  pages        = {8748--8763},
  year         = {2021}
}

@inproceedings{geneval,
  author       = {Dhruba Ghosh and
                  Hannaneh Hajishirzi and
                  Ludwig Schmidt},
  title        = {GenEval: An object-focused framework for evaluating text-to-image
                  alignment},
  booktitle    = {NeurIPS},
  year         = {2023},
}

@inproceedings{sd3,
  title={Scaling rectified flow transformers for high-resolution image synthesis},
  author={Esser, Patrick and Kulal, Sumith and Blattmann, Andreas and Entezari, Rahim and M{\"u}ller, Jonas and Saini, Harry and Levi, Yam and Lorenz, Dominik and Sauer, Axel and Boesel, Frederic and others},
  booktitle={ICML},
  year={2024}
}

@inproceedings{sohl2015deep,
  title={Deep unsupervised learning using nonequilibrium thermodynamics},
  author={Sohl-Dickstein, Jascha and Weiss, Eric and Maheswaranathan, Niru and Ganguli, Surya},
  booktitle={ICML},
  pages={2256--2265},
  year={2015},
}

@article{caron2021emerging,
	title={Emerging properties in self-supervised vision transformers},
	author={Caron, Mathilde and Touvron, Hugo and Misra, Ishan and J{\'e}gou, Herv{\'e} and Mairal, Julien and Bojanowski, Piotr and Joulin, Armand},
	journal={Proceedings of the IEEE/CVF international conference on computer vision},
	pages={9650--9660},
	year={2021}
}

@article{qwen_vl,
  author       = {Jinze Bai and
                  Shuai Bai and
                  Shusheng Yang and
                  Shijie Wang and
                  Sinan Tan and
                  Peng Wang and
                  Junyang Lin and
                  Chang Zhou and
                  Jingren Zhou},
  title        = {Qwen-VL: {A} Frontier Large Vision-Language Model with Versatile Abilities},
  journal      = {CoRR},
  volume       = {abs/2308.12966},
  year         = {2023}
}

@article{vae,
  title={Auto-encoding variational bayes},
  author={Kingma, Diederik P and Welling, Max},
  journal={arXiv preprint arXiv:1312.6114},
  year={2013}
}

@article{ho2022classifier,
  title={Classifier-free diffusion guidance},
  author={Ho, Jonathan and Salimans, Tim},
  journal={arXiv preprint arXiv:2207.12598},
  year={2022}
}

@inproceedings{journeydb,
  author       = {Keqiang Sun and
                  Junting Pan and
                  Yuying Ge and
                  Hao Li and
                  Haodong Duan and
                  Xiaoshi Wu and
                  Renrui Zhang and
                  Aojun Zhou and
                  Zipeng Qin and
                  Yi Wang and
                  Jifeng Dai and
                  Yu Qiao and
                  Limin Wang and
                  Hongsheng Li},
  title        = {JourneyDB: {A} Benchmark for Generative Image Understanding},
  booktitle    = {NeurIPS},
  year         = {2023},
}

@article{jin2025srum,
  title={SRUM: Fine-Grained Self-Rewarding for Unified Multimodal Models},
  author={Jin, Weiyang and Niu, Yuwei and Liao, Jiaqi and Duan, Chengqi and Li, Aoxue and Gao, Shenghua and Liu, Xihui},
  journal={arXiv preprint arXiv:2510.12784},
  year={2025}
}

@article{huang2023t2i,
  title={T2i-compbench: A comprehensive benchmark for open-world compositional text-to-image generation},
  author={Huang, Kaiyi and Sun, Kaiyue and Xie, Enze and Li, Zhenguo and Liu, Xihui},
  journal={Advances in Neural Information Processing Systems},
  volume={36},
  pages={78723--78747},
  year={2023}
}

@inproceedings{ddpm,
    author      = {Ho, Jonathan and Jain, Ajay and Abbeel, Pieter},
    booktitle   = {NeurIPS},
    pages       = {6840--6851},
    title       = {Denoising Diffusion Probabilistic Models},
    year        = {2020}
}

@article{zhang2025unifiedmultimodalunderstandinggeneration,
  title={Unified multimodal understanding and generation models: Advances, challenges, and opportunities},
  author={Zhang, Xinjie and Guo, Jintao and Zhao, Shanshan and Fu, Minghao and Duan, Lunhao and Hu, Jiakui and Chng, Yong Xien and Wang, Guo-Hua and Chen, Qing-Guo and Xu, Zhao and others},
  journal={arXiv preprint arXiv:2505.02567},
  year={2025}
}

@article{tian2025unigen,
  title={UniGen: Enhanced Training \& Test-Time Strategies for Unified Multimodal Understanding and Generation},
  author={Tian, Rui and Gao, Mingfei and Xu, Mingze and Hu, Jiaming and Lu, Jiasen and Wu, Zuxuan and Yang, Yinfei and Dehghan, Afshin},
  journal={arXiv preprint arXiv:2505.14682},
  year={2025}
}

@inproceedings{pope,
author       = {Yifan Li and
              Yifan Du and
              Kun Zhou and
              Jinpeng Wang and
              Wayne Xin Zhao and
              Ji{-}Rong Wen},
title        = {Evaluating Object Hallucination in Large Vision-Language Models},
booktitle    = {{EMNLP}},
pages        = {292--305},
publisher    = {Association for Computational Linguistics},
year         = {2023}
}

@article{mme,
author       = {Chaoyou Fu and
              Peixian Chen and
              Yunhang Shen and
              Yulei Qin and
              Mengdan Zhang and
              Xu Lin and
              Zhenyu Qiu and
              Wei Lin and
              Jinrui Yang and
              Xiawu Zheng and
              Ke Li and
              Xing Sun and
              Rongrong Ji},
title        = {{MME:} {A} Comprehensive Evaluation Benchmark for Multimodal Large
              Language Models},
journal      = {CoRR},
volume       = {abs/2306.13394},
year         = {2023}
}

@inproceedings{mmmu,
author       = {Xiang Yue and
              Yuansheng Ni and
              Tianyu Zheng and
              Kai Zhang and
              Ruoqi Liu and
              Ge Zhang and
              Samuel Stevens and
              Dongfu Jiang and
              Weiming Ren and
              Yuxuan Sun and
              Cong Wei and
              Botao Yu and
              Ruibin Yuan and
              Renliang Sun and
              Ming Yin and
              Boyuan Zheng and
              Zhenzhu Yang and
              Yibo Liu and
              Wenhao Huang and
              Huan Sun and
              Yu Su and
              Wenhu Chen},
title        = {{MMMU:} {A} Massive Multi-Discipline Multimodal Understanding and
              Reasoning Benchmark for Expert {AGI}},
booktitle    = {{CVPR}},
pages        = {9556--9567},
publisher    = {{IEEE}},
year         = {2024}
}

@inproceedings{gqa,
author       = {Drew A. Hudson and
          Christopher D. Manning},
title        = {{GQA:} {A} New Dataset for Real-World Visual Reasoning and Compositional
          Question Answering},
booktitle    = {{CVPR}},
pages        = {6700--6709},
publisher    = {Computer Vision Foundation / {IEEE}},
year         = {2019}
}

@inproceedings{
showo,
title={Show-o: One Single Transformer to Unify Multimodal Understanding and Generation},
author={Jinheng Xie and Weijia Mao and Zechen Bai and David Junhao Zhang and Weihao Wang and Kevin Qinghong Lin and Yuchao Gu and Zhijie Chen and Zhenheng Yang and Mike Zheng Shou},
booktitle={ICLR},
year={2025},
}

@article{DiT,
  title={Scalable Diffusion Models with Transformers},
  author={William Peebles and Saining Xie},
  year={2022},
  journal={arXiv preprint arXiv:2212.09748},
}

@inproceedings{flow,
title={Flow Matching for Generative Modeling},
author={Yaron Lipman and Ricky T. Q. Chen and Heli Ben-Hamu and Maximilian Nickel and Matthew Le},
booktitle={The Eleventh International Conference on Learning Representations },
year={2023},
url={https://openreview.net/forum?id=PqvMRDCJT9t}
}

@inproceedings{
zhou2025transfusion,
title={Transfusion: Predict the Next Token and Diffuse Images with One Multi-Modal Model},
author={Chunting Zhou and LILI YU and Arun Babu and Kushal Tirumala and Michihiro Yasunaga and Leonid Shamis and Jacob Kahn and Xuezhe Ma and Luke Zettlemoyer and Omer Levy},
booktitle={ICLR},
year={2025},
}

@article{wang2024emu3,
  title={Emu3: Next-Token Prediction is All You Need},
  author={Wang, Xinlong and Zhang, Xiaosong and Luo, Zhengxiong and Sun, Quan and Cui, Yufeng and Wang, Jinsheng and Zhang, Fan and Wang, Yueze and Li, Zhen and Yu, Qiying and others},
  journal={arXiv preprint arXiv:2409.18869},
  year={2024}
}

@article{qwen2.5,
    title   = {Qwen2.5 Technical Report}, 
    author  = {An Yang and Baosong Yang and Beichen Zhang and Binyuan Hui and Bo Zheng and Bowen Yu and Chengyuan Li and Dayiheng Liu and Fei Huang and Haoran Wei and Huan Lin and Jian Yang and Jianhong Tu and Jianwei Zhang and Jianxin Yang and Jiaxi Yang and Jingren Zhou and Junyang Lin and Kai Dang and Keming Lu and Keqin Bao and Kexin Yang and Le Yu and Mei Li and Mingfeng Xue and Pei Zhang and Qin Zhu and Rui Men and Runji Lin and Tianhao Li and Tingyu Xia and Xingzhang Ren and Xuancheng Ren and Yang Fan and Yang Su and Yichang Zhang and Yu Wan and Yuqiong Liu and Zeyu Cui and Zhenru Zhang and Zihan Qiu},
    journal = {arXiv preprint arXiv:2412.15115},
    year    = {2024}
}

@article{Qwen2.5-VL,
  title={Qwen2.5-VL Technical Report},
  author={Bai, Shuai and Chen, Keqin and Liu, Xuejing and Wang, Jialin and Ge, Wenbin and Song, Sibo and Dang, Kai and Wang, Peng and Wang, Shijie and Tang, Jun and Zhong, Humen and Zhu, Yuanzhi and Yang, Mingkun and Li, Zhaohai and Wan, Jianqiang and Wang, Pengfei and Ding, Wei and Fu, Zheren and Xu, Yiheng and Ye, Jiabo and Zhang, Xi and Xie, Tianbao and Cheng, Zesen and Zhang, Hang and Yang, Zhibo and Xu, Haiyang and Lin, Junyang},
  journal={arXiv preprint arXiv:2502.13923},
  year={2025}
}

@article{internvl,
  title={Expanding Performance Boundaries of Open-Source Multimodal Models with Model, Data, and Test-Time Scaling},
  author={Chen, Zhe and Wang, Weiyun and Cao, Yue and Liu, Yangzhou and Gao, Zhangwei and Cui, Erfei and Zhu, Jinguo and Ye, Shenglong and Tian, Hao and Liu, Zhaoyang and others},
  journal={arXiv preprint arXiv:2412.05271},
  year={2024}
}

@article{vila-u,
  title={Vila-u: a unified foundation model integrating visual understanding and generation},
  author={Wu, Yecheng and Zhang, Zhuoyang and Chen, Junyu and Tang, Haotian and Li, Dacheng and Fang, Yunhao and Zhu, Ligeng and Xie, Enze and Yin, Hongxu and Yi, Li and others},
  journal={arXiv preprint arXiv:2409.04429},
  year={2024}
}

@article{tong2024metamorph,
  title={Metamorph: Multimodal understanding and generation via instruction tuning},
  author={Tong, Shengbang and Fan, David and Zhu, Jiachen and Xiong, Yunyang and Chen, Xinlei and Sinha, Koustuv and Rabbat, Michael and LeCun, Yann and Xie, Saining and Liu, Zhuang},
  journal={arXiv preprint arXiv:2412.14164},
  year={2024}
}

@article{qu2024tokenflow,
  title={Tokenflow: Unified image tokenizer for multimodal understanding and generation},
  author={Qu, Liao and Zhang, Huichao and Liu, Yiheng and Wang, Xu and Jiang, Yi and Gao, Yiming and Ye, Hu and Du, Daniel K and Yuan, Zehuan and Wu, Xinglong},
  journal={arXiv preprint arXiv:2412.03069},
  year={2024}
}

@article{chen2025janus,
  title={Janus-Pro: Unified Multimodal Understanding and Generation with Data and Model Scaling},
  author={Chen, Xiaokang and Wu, Zhiyu and Liu, Xingchao and Pan, Zizheng and Liu, Wen and Xie, Zhenda and Yu, Xingkai and Ruan, Chong},
  journal={arXiv preprint arXiv:2501.17811},
  year={2025}
}

@misc{ma2024janusflow,
      title={JanusFlow: Harmonizing Autoregression and Rectified Flow for Unified Multimodal Understanding and Generation}, 
      author={Yiyang Ma and Xingchao Liu and Xiaokang Chen and Wen Liu and Chengyue Wu and Zhiyu Wu and Zizheng Pan and Zhenda Xie and Haowei Zhang and Xingkai yu and Liang Zhao and Yisong Wang and Jiaying Liu and Chong Ruan},
      journal={arXiv preprint arXiv:2411.07975},
      year={2024}
}

@article{unifluid,
  title={Unified autoregressive visual generation and understanding with continuous tokens},
  author={Fan, Lijie and Tang, Luming and Qin, Siyang and Li, Tianhong and Yang, Xuan and Qiao, Siyuan and Steiner, Andreas and Sun, Chen and Li, Yuanzhen and Zhu, Tao and others},
  journal={arXiv preprint arXiv:2503.13436},
  year={2025}
}

@article{metaqueries,
  title={Transfer between Modalities with MetaQueries},
  author={Pan, Xichen and Shukla, Satya Narayan and Singh, Aashu and Zhao, Zhuokai and Mishra, Shlok Kumar and Wang, Jialiang and Xu, Zhiyang and Chen, Jiuhai and Li, Kunpeng and Juefei-Xu, Felix and Hou, Ji and Xie, Saining},
  journal={arXiv preprint arXiv:2504.06256},
  year={2025}
}

@article{janus,
  title={Janus-Pro: Unified Multimodal Understanding and Generation with Data and Model Scaling},
  author={Chen, Xiaokang and Wu, Zhiyu and Liu, Xingchao and Pan, Zizheng and Liu, Wen and Xie, Zhenda and Yu, Xingkai and Ruan, Chong},
  journal={arXiv preprint arXiv:2501.17811},
  year={2025}
}

@article{MMAR,
  author       = {Jian Yang and
                  Dacheng Yin and
                  Yizhou Zhou and
                  Fengyun Rao and
                  Wei Zhai and
                  Yang Cao and
                  Zheng{-}Jun Zha},
  title        = {{MMAR:} Towards Lossless Multi-Modal Auto-Regressive Probabilistic
                  Modeling},
journal={arXiv preprint arXiv:2410.10798},
  year         = {2024},
}

@article{lmfusion,
  author       = {Weijia Shi and
                  Xiaochuang Han and
                  Chunting Zhou and
                  Weixin Liang and
                  Xi Victoria Lin and
                  Luke Zettlemoyer and
                  Lili Yu},
  title        = {LMFusion: Adapting Pretrained Language Models for Multimodal Generation},
journal = {arXiv preprint arXiv: 2412.15188},
  year         = {2024}
}

@article{liquid,
  title={Liquid: Language models are scalable multi-modal generators},
  author={Wu, Junfeng and Jiang, Yi and Ma, Chuofan and Liu, Yuliang and Zhao, Hengshuang and Yuan, Zehuan and Bai, Song and Bai, Xiang},
  journal={arXiv preprint arXiv:2412.04332},
  year={2024}
}

@article{wu2025harmon,
  title={Harmonizing visual representations for unified multimodal understanding and generation},
  author={Wu, Size and Zhang, Wenwei and Xu, Lumin and Jin, Sheng and Wu, Zhonghua and Tao, Qingyi and Liu, Wentao and Li, Wei and Loy, Chen Change},
  journal={arXiv preprint arXiv:2503.21979},
  year={2025}
}

@article{Nexus-Gen,
      title={Nexus-Gen: A Unified Model for Image Understanding, Generation, and Editing}, 
      author={Hong Zhang and Zhongjie Duan and Xingjun Wang and Yingda Chen and Yuze Zhao and Yu Zhang},
      journal={arXiv preprint arXiv:2504.21356},
      year={2025}
}

@article{seedbench,
  title={Seed-bench: Benchmarking multimodal llms with generative comprehension},
  author={Li, Bohao and Wang, Rui and Wang, Guangzhi and Ge, Yuying and Ge, Yixiao and Shan, Ying},
  journal={arXiv preprint arXiv:2307.16125},
  year={2023}
}

@misc{siglip,
      title={Sigmoid Loss for Language Image Pre-Training}, 
      author={Xiaohua Zhai and Basil Mustafa and Alexander Kolesnikov and Lucas Beyer},
      year={2023},
      eprint={2303.15343},
      archivePrefix={arXiv},
      primaryClass={cs.CV}
}

@misc{dpgbench,
      title={ELLA: Equip Diffusion Models with LLM for Enhanced Semantic Alignment}, 
      author={Xiwei Hu and Rui Wang and Yixiao Fang and Bin Fu and Pei Cheng and Gang Yu},
      year={2024},
      eprint={2403.05135},
      archivePrefix={arXiv},
      primaryClass={cs.CV}
}

@inproceedings{tong2024eyes,
  title={Eyes wide shut? exploring the visual shortcomings of multimodal llms},
  author={Tong, Shengbang and Liu, Zhuang and Zhai, Yuexiang and Ma, Yi and LeCun, Yann and Xie, Saining},
  booktitle={Proceedings of the IEEE/CVF Conference on Computer Vision and Pattern Recognition},
  pages={9568--9578},
  year={2024}
}

@article{wisebench,
  title={WISE: A World Knowledge-Informed Semantic Evaluation for Text-to-Image Generation},
  author={Niu, Yuwei and Ning, Munan and Zheng, Mengren and Lin, Bin and Jin, Peng and Liao, Jiaqi and Ning, Kunpeng and Zhu, Bin and Yuan, Li},
  journal={arXiv preprint arXiv:2503.07265},
  year={2025}
}

@article{dehghani2023patch,
  title={Patch n’pack: Navit, a vision transformer for any aspect ratio and resolution},
  author={Dehghani, Mostafa and Mustafa, Basil and Djolonga, Josip and Heek, Jonathan and Minderer, Matthias and Caron, Mathilde and Steiner, Andreas and Puigcerver, Joan and Geirhos, Robert and Alabdulmohsin, Ibrahim M and others},
  journal={Advances in Neural Information Processing Systems},
  volume={36},
  pages={2252--2274},
  year={2023}
}

@misc{dalle3,
  title={Improving Image Generation with Better Captions},
  author={James Betker and Gabriel Goh and Li Jing and † TimBrooks and Jianfeng Wang and Linjie Li and † LongOuyang and † JuntangZhuang and † JoyceLee and † YufeiGuo and † WesamManassra and † PrafullaDhariwal and † CaseyChu and † YunxinJiao and Aditya Ramesh},
  year={2023},
}

@article{bagel,
  title={Emerging properties in unified multimodal pretraining},
  author={Deng, Chaorui and Zhu, Deyao and Li, Kunchang and Gou, Chenhui and Li, Feng and Wang, Zeyu and Zhong, Shu and Yu, Weihao and Nie, Xiaonan and Song, Ziang and others},
  journal={arXiv preprint arXiv:2505.14683},
  year={2025}
}

@article{kuprashevich2025nohumansrequiredautonomoushighqualityimage,
  title={NoHumansRequired: Autonomous High-Quality Image Editing Triplet Mining},
  author={Kuprashevich, Maksim and Alekseenko, Grigorii and Tolstykh, Irina and Fedorov, Georgii and Suleimanov, Bulat and Dokholyan, Vladimir and Gordeev, Aleksandr},
  journal={Available at SSRN 5381374},
  year={2025}
}

@article{wang2025ovisu1technicalreport,
  title={Ovis-U1 Technical Report},
  author={Wang, Guo-Hua and Zhao, Shanshan and Zhang, Xinjie and Cao, Liangfu and Zhan, Pengxin and Duan, Lunhao and Lu, Shiyin and Fu, Minghao and Chen, Xiaohao and Zhao, Jianshan and others},
  journal={arXiv preprint arXiv:2506.23044},
  year={2025}
}

@article{zhang2025context,
  title={In-context edit: Enabling instructional image editing with in-context generation in large scale diffusion transformer},
  author={Zhang, Zechuan and Xie, Ji and Lu, Yu and Yang, Zongxin and Yang, Yi},
  journal={arXiv preprint arXiv:2504.20690},
  year={2025}
}

@article{luo2024deem,
  title={Deem: Diffusion models serve as the eyes of large language models for image perception},
  author={Luo, Run and Li, Yunshui and Chen, Longze and He, Wanwei and Lin, Ting-En and Liu, Ziqiang and Zhang, Lei and Song, Zikai and Xia, Xiaobo and Liu, Tongliang and others},
  journal={arXiv preprint arXiv:2405.15232},
  year={2024}
}

@article{li2025onecatdecoderonlyautoregressivemodel,
  title={OneCAT: Decoder-Only Auto-Regressive Model for Unified Understanding and Generation},
  author={Li, Han and Peng, Xinyu and Wang, Yaoming and Peng, Zelin and Chen, Xin and Weng, Rongxiang and Wang, Jingang and Cai, Xunliang and Dai, Wenrui and Xiong, Hongkai},
  journal={arXiv preprint arXiv:2509.03498},
  year={2025}
}

@article{wu2025qwen,
  title={Qwen-image technical report},
  author={Wu, Chenfei and Li, Jiahao and Zhou, Jingren and Lin, Junyang and Gao, Kaiyuan and Yan, Kun and Yin, Sheng-ming and Bai, Shuai and Xu, Xiao and Chen, Yilei and others},
  journal={arXiv preprint arXiv:2508.02324},
  year={2025}
}

@article{wang2025skyworkunipicunifiedautoregressive,
  title={Skywork UniPic: Unified Autoregressive Modeling for Visual Understanding and Generation},
  author={Wang, Peiyu and Peng, Yi and Gan, Yimeng and Hu, Liang and Xie, Tianyidan and Wang, Xiaokun and Wei, Yichen and Tang, Chuanxin and Zhu, Bo and Li, Changshi and others},
  journal={arXiv preprint arXiv:2508.03320},
  year={2025}
}

@inproceedings{ma2025learning,
  title={Learning visual generative priors without text},
  author={Ma, Shuailei and Zheng, Kecheng and Wei, Ying and Wu, Wei and Lu, Fan and Zhang, Yifei and Xie, Chen-Wei and Gong, Biao and Zhu, Jiapeng and Shen, Yujun},
  booktitle={Proceedings of the Computer Vision and Pattern Recognition Conference},
  pages={8051--8061},
  year={2025}
}

@article{wang2025ross3d,
  title={Ross3d: Reconstructive visual instruction tuning with 3d-awareness},
  author={Wang, Haochen and Zhao, Yucheng and Wang, Tiancai and Fan, Haoqiang and Zhang, Xiangyu and Zhang, Zhaoxiang},
  journal={arXiv preprint arXiv:2504.01901},
  year={2025}
}

@article{yu2024representation,
  title={Representation alignment for generation: Training diffusion transformers is easier than you think},
  author={Yu, Sihyun and Kwak, Sangkyung and Jang, Huiwon and Jeong, Jongheon and Huang, Jonathan and Shin, Jinwoo and Xie, Saining},
  journal={arXiv preprint arXiv:2410.06940},
  year={2024}
}

@article{wang2025gptimageedit15mmillionscalegptgeneratedimage,
  title={Gpt-image-edit-1.5 m: A million-scale, gpt-generated image dataset},
  author={Wang, Yuhan and Yang, Siwei and Zhao, Bingchen and Zhang, Letian and Liu, Qing and Zhou, Yuyin and Xie, Cihang},
  journal={arXiv preprint arXiv:2507.21033},
  year={2025}
}

@article{blip3,
  title={Blip3-o: A family of fully open unified multimodal models-architecture, training and dataset},
  author={Chen, Jiuhai and Xu, Zhiyang and Pan, Xichen and Hu, Yushi and Qin, Can and Goldstein, Tom and Huang, Lifu and Zhou, Tianyi and Xie, Saining and Savarese, Silvio and others},
  journal={arXiv preprint arXiv:2505.09568},
  year={2025}
}

@inproceedings{fu2024blinkmultimodallargelanguage,
  title={Blink: Multimodal large language models can see but not perceive},
  author={Fu, Xingyu and Hu, Yushi and Li, Bangzheng and Feng, Yu and Wang, Haoyu and Lin, Xudong and Roth, Dan and Smith, Noah A and Ma, Wei-Chiu and Krishna, Ranjay},
  booktitle={European Conference on Computer Vision},
  pages={148--166},
  year={2024},
  organization={Springer}
}

@misc{texttoimage2M,
    title={jackyhate/text-to-image-2M},
    author={jackyhate},
    url={https://huggingface.co/datasets/jackyhate/text-to-image-2M},
    year={2024},
}

@misc{midjourneyv6,
    title={CortexLM/midjourney-v6},
    author={CortexLM},
    url={https://huggingface.co/datasets/CortexLM/midjourney-v6},
    year={2024},
}

@misc{midjourneyv6llava,
    title={brivangl/midjourney-v6-llava},
    author={brivangl},
    url={https://huggingface.co/datasets/brivangl/midjourney-v6-llava},
    year={2024},
}

@misc{openai2024introducing4o,
  title={Introducing GPT-4o with image generation capabilities},
  author={OpenAI},
  url={https://openai.com/index/introducing-4o-image-generation},
  year={2024},
  note={Accessed: 2025-07-04}
}

@article{liu2024playgroundv3improvingtexttoimage,
  title={Playground v3: Improving text-to-image alignment with deep-fusion large language models},
  author={Liu, Bingchen and Akhgari, Ehsan and Visheratin, Alexander and Kamko, Aleks and Xu, Linmiao and Shrirao, Shivam and Lambert, Chase and Souza, Joao and Doshi, Suhail and Li, Daiqing},
  journal={arXiv preprint arXiv:2409.10695},
  year={2024}
}

@misc{flux,
    title = {Black Forest Labs; Frontier AI Lab},
    url = {https://blackforestlabs.ai/},
    author = {BlackForest},
    year = {2024}
}

@inproceedings{vilex,
  title={Visual lexicon: Rich image features in language space},
  author={Wang, XuDong and Zhou, Xingyi and Fathi, Alireza and Darrell, Trevor and Schmid, Cordelia},
  booktitle={Proceedings of the Computer Vision and Pattern Recognition Conference},
  pages={19736--19747},
  year={2025}
}

@article{wu2025omnigen2explorationadvancedmultimodal,
  title={OmniGen2: Exploration to Advanced Multimodal Generation},
  author={Chenyuan Wu and Pengfei Zheng and Ruiran Yan and Shitao Xiao and Xin Luo and Yueze Wang and Wanli Li and Xiyan Jiang and Yexin Liu and Junjie Zhou and Ze Liu and Ziyi Xia and Chaofan Li and Haoge Deng and Jiahao Wang and Kun Luo and Bo Zhang and Defu Lian and Xinlong Wang and Zhongyuan Wang and Tiejun Huang and Zheng Liu},
  journal={arXiv preprint arXiv:2506.18871},
  year={2025}
}

@article{lin2025uniworld,
  title={Uniworld: High-resolution semantic encoders for unified visual understanding and generation},
  author={Lin, Bin and Li, Zongjian and Cheng, Xinhua and Niu, Yuwei and Ye, Yang and He, Xianyi and Yuan, Shenghai and Yu, Wangbo and Wang, Shaodong and Ge, Yunyang and others},
  journal={arXiv preprint arXiv:2506.03147},
  year={2025}
}

@article{jiang2025co,
  title={Co-Reinforcement Learning for Unified Multimodal Understanding and Generation},
  author={Jiang, Jingjing and Si, Chongjie and Luo, Jun and Zhang, Hanwang and Ma, Chao},
  journal={arXiv preprint arXiv:2505.17534},
  year={2025}
}

@article{mao2025unirl,
  title={UniRL: Self-Improving Unified Multimodal Models via Supervised and Reinforcement Learning},
  author={Mao, Weijia and Yang, Zhenheng and Shou, Mike Zheng},
  journal={arXiv preprint arXiv:2505.23380},
  year={2025}
}

@article{chen2025sharegpt,
  title={ShareGPT-4o-Image: Aligning Multimodal Models with GPT-4o-Level Image Generation},
  author={Chen, Junying and Cai, Zhenyang and Chen, Pengcheng and Chen, Shunian and Ji, Ke and Wang, Xidong and Yang, Yunjin and Wang, Benyou},
  journal={arXiv preprint arXiv:2506.18095},
  year={2025}
}

@article{ye2025imgedit,
  title={Imgedit: A unified image editing dataset and benchmark},
  author={Ye, Yang and He, Xianyi and Li, Zongjian and Lin, Bin and Yuan, Shenghai and Yan, Zhiyuan and Hou, Bohan and Yuan, Li},
  journal={arXiv preprint arXiv:2505.20275},
  year={2025}
}

@article{xie2025show,
  title={Show-o2: Improved Native Unified Multimodal Models},
  author={Xie, Jinheng and Yang, Zhenheng and Shou, Mike Zheng},
  journal={arXiv preprint arXiv:2506.15564},
  year={2025}
}

@article{geng2025xomnireinforcementlearningmakes,
  title={X-omni: Reinforcement learning makes discrete autoregressive image generative models great again},
  author={Geng, Zigang and Wang, Yibing and Ma, Yeyao and Li, Chen and Rao, Yongming and Gu, Shuyang and Zhong, Zhao and Lu, Qinglin and Hu, Han and Zhang, Xiaosong and others},
  journal={arXiv preprint arXiv:2507.22058},
  year={2025}
}

@article{liu2025step1x,
  title={Step1x-edit: A practical framework for general image editing},
  author={Liu, Shiyu and Han, Yucheng and Xing, Peng and Yin, Fukun and Wang, Rui and Cheng, Wei and Liao, Jiaqi and Wang, Yingming and Fu, Honghao and Han, Chunrui and others},
  journal={arXiv preprint arXiv:2504.17761},
  year={2025}
}

@article{ma2025genhancer,
  title={GenHancer: Imperfect generative models are secretly strong vision-centric enhancers},
  author={Ma, Shijie and Ge, Yuying and Wang, Teng and Guo, Yuxin and Ge, Yixiao and Shan, Ying},
  journal={arXiv preprint arXiv:2503.19480},
  year={2025}
}

@article{wang2024diffusion,
  title={Diffusion feedback helps clip see better},
  author={Wang, Wenxuan and Sun, Quan and Zhang, Fan and Tang, Yepeng and Liu, Jing and Wang, Xinlong},
  journal={arXiv preprint arXiv:2407.20171},
  year={2024}
}

@inproceedings{pan2025generative,
  title={Generative Multimodal Pretraining with Discrete Diffusion Timestep Tokens},
  author={Pan, Kaihang and Lin, Wang and Yue, Zhongqi and Ao, Tenglong and Jia, Liyu and Zhao, Wei and Li, Juncheng and Tang, Siliang and Zhang, Hanwang},
  booktitle={Proceedings of the Computer Vision and Pattern Recognition Conference},
  pages={26136--26146},
  year={2025}
}

@article{wang2024reconstructive,   title={Reconstructive visual instruction tuning},   author={Wang, Haochen and Zheng, Anlin and Zhao, Yucheng and Wang, Tiancai and Ge, Zheng and Zhang, Xiangyu and Zhang, Zhaoxiang},   journal={arXiv preprint arXiv:2410.09575},   year={2024} }

@inproceedings{ma2024sit,
  title={Sit: Exploring flow and diffusion-based generative models with scalable interpolant transformers},
  author={Ma, Nanye and Goldstein, Mark and Albergo, Michael S and Boffi, Nicholas M and Vanden-Eijnden, Eric and Xie, Saining},
  booktitle={European Conference on Computer Vision},
  pages={23--40},
  year={2024},
  organization={Springer}
}

@inproceedings{chen2024spatialvlm,
  title={Spatialvlm: Endowing vision-language models with spatial reasoning capabilities},
  author={Chen, Boyuan and Xu, Zhuo and Kirmani, Sean and Ichter, Brain and Sadigh, Dorsa and Guibas, Leonidas and Xia, Fei},
  booktitle={Proceedings of the IEEE/CVF Conference on Computer Vision and Pattern Recognition},
  pages={14455--14465},
  year={2024}
}

@article{huang2025vision,
  title={Vision-r1: Incentivizing reasoning capability in multimodal large language models},
  author={Huang, Wenxuan and Jia, Bohan and Zhai, Zijie and Cao, Shaosheng and Ye, Zheyu and Zhao, Fei and Xu, Zhe and Hu, Yao and Lin, Shaohui},
  journal={arXiv preprint arXiv:2503.06749},
  year={2025}
}

@inproceedings{yao2025reconstruction,
  title={Reconstruction vs. generation: Taming optimization dilemma in latent diffusion models},
  author={Yao, Jingfeng and Yang, Bin and Wang, Xinggang},
  booktitle={Proceedings of the Computer Vision and Pattern Recognition Conference},
  pages={15703--15712},
  year={2025}
}

@article{xie2025sana,
  title={Sana 1.5: Efficient scaling of training-time and inference-time compute in linear diffusion transformer},
  author={Xie, Enze and Chen, Junsong and Zhao, Yuyang and Yu, Jincheng and Zhu, Ligeng and Wu, Chengyue and Lin, Yujun and Zhang, Zhekai and Li, Muyang and Chen, Junyu and others},
  journal={arXiv preprint arXiv:2501.18427},
  year={2025}
}

@article{han2025self,
  title={Self-Contradiction as Self-Improvement: Mitigating the Generation-Understanding Gap in MLLMs},
  author={Han, Yujin and Chen, Hao and Han, Andi and Wang, Zhiheng and Lin, Xinyu and Zhang, Yingya and Zhang, Shiwei and Zou, Difan},
  journal={arXiv preprint arXiv:2507.16663},
  year={2025}
}

@inproceedings{lin2014microsoft,
  title={Microsoft coco: Common objects in context},
  author={Lin, Tsung-Yi and Maire, Michael and Belongie, Serge and Hays, James and Perona, Pietro and Ramanan, Deva and Doll{\'a}r, Piotr and Zitnick, C Lawrence},
  booktitle={European conference on computer vision},
  pages={740--755},
  year={2014},
  organization={Springer}
}

@article{abdi2010principal,
  title={Principal component analysis},
  author={Abdi, Herv{\'e} and Williams, Lynne J},
  journal={Wiley interdisciplinary reviews: computational statistics},
  volume={2},
  number={4},
  pages={433--459},
  year={2010},
  publisher={Wiley Online Library}
}

@article{liu2022flow,
  title={Flow straight and fast: Learning to generate and transfer data with rectified flow},
  author={Liu, Xingchao and Gong, Chengyue and Liu, Qiang},
  journal={arXiv preprint arXiv:2209.03003},
  year={2022}
}
